\documentclass{article}

\usepackage{microtype}
\usepackage{graphicx}
\usepackage{subfigure}
\usepackage{booktabs} 

\usepackage{hyperref}

\usepackage[accepted]{icml2023}

\usepackage{amsmath}
\usepackage{amssymb}
\usepackage{mathtools}
\usepackage{amsthm}
\usepackage{xcolor}

\usepackage[capitalize,noabbrev]{cleveref}

\theoremstyle{plain}
\newtheorem{theorem}{Theorem}[section]

\newtheorem{lemma}[theorem]{Lemma}

\theoremstyle{definition}

\newtheorem{assumption}{Assumption}
\theoremstyle{remark}

\usepackage[textsize=tiny]{todonotes}
\usepackage{soul}

\icmltitlerunning{Improved Algorithms for Multi-period Packing Problems with Bandit Feedback}

\makeatother

\begin{document}
\global\long\def\Real{\mathbb{R}}%
\global\long\def\Natural{\mathbb{N}}%
\global\long\def\Expectation{\mathbb{E}}%
\global\long\def\Probability{\mathbb{P}}%
\global\long\def\Var{\mathbb{V}}%
\global\long\def\CE#1#2{\Expectation\left[\left.#1\right|#2\right]}%
\global\long\def\CP#1#2{\Probability\left(\left.#1\right|#2\right)}%
\global\long\def\abs#1{\left|#1\right|}%
\global\long\def\norm#1{\left\Vert #1\right\Vert }%
\global\long\def\Indicator#1{\mathbb{I}\left(#1\right)}%

\global\long\def\Parameter#1{\theta_{\star}^{(#1)}}%
\global\long\def\RParameter#1{W_{\star}^{(#1)}}%
\global\long\def\Setofcontexts#1{\mathcal{X}_{#1}}%
\global\long\def\Classdistribution#1{\mathbb{F}_{#1}}%
\global\long\def\Conentry#1#2#3#4{b_{#2,#3}^{(#1)}(#4)}%
\global\long\def\Contilde#1#2#3{\tilde{\mathbf{b}}_{#2,#3}^{(#1)}}%
\global\long\def\Consumption#1#2#3{\mathbf{b}_{#2,#3}^{(#1)}}%
\global\long\def\Optcon#1#2{\mathbf{b}_{#2}^{\star(#1)}}%
\global\long\def\Conhat#1#2#3{\widehat{\mathbf{b}}_{#2,#3}^{(#1)}}%
\global\long\def\Reward#1#2#3{r_{#2,#3}^{(#1)}}%
\global\long\def\Context#1#2#3{\mathbf{x}_{#2,#3}^{(#1)}}%
\global\long\def\Error#1#2{\eta_{#1,#2}}%
\global\long\def\RError#1#2{\mathbf{n}_{#1,#2}}%
\global\long\def\Policy#1#2#3{\pi_{#2,#3}^{(#1)}}%
\global\long\def\OptPolicy#1#2{\pi_{#2}^{\star(#1)}}%
\global\long\def\TildePolicy#1#2#3{\tilde{\pi}_{#2,#3}^{(#1)}}%
\global\long\def\HatPolicy#1#2#3{\widehat{\pi}_{#2,#3}^{(#1)}}%
\global\long\def\Estimator#1#2{\widehat{\theta}_{#2}^{(#1)}}%
\global\long\def\OptUtility#1#2{u_{#2}^{\star(#1)}}%
\global\long\def\TildeUtility#1#2#3{\tilde{u}{}_{#2,#3}^{(#1)}}%
\global\long\def\HatUtility#1#2#3{\widehat{u}{}_{#2,#3}^{(#1)}}%
\global\long\def\StackedEstimator#1{\widehat{\Theta}_{#1}}%
\global\long\def\DREstimator#1{\widehat{\Theta}_{#1}^{DR}}%
\global\long\def\IPW#1{\widehat{\Theta}_{#1}^{IPW}}%
\global\long\def\DRY#1#2#3{r_{#2,#3}^{DR(#1)}}%
\global\long\def\TildeR#1#2{\tilde{r}_{#1,#2}}%
\global\long\def\TildeB#1#2{\tilde{\mathbf{b}}_{#1,#2}}%
\global\long\def\TildeError#1#2#3{\tilde{\eta}_{#2,#3}^{(#1)}}%
\global\long\def\NewProb#1#2{\phi_{#1,#2}}%
\global\long\def\StackedContext#1#2{\tilde{X}_{#1,#2}}%
\global\long\def\Selfbound#1#2{\beta_{#1}(#2)}%
\global\long\def\Predbound#1#2#3{\gamma_{#1,#2}(#3)}%

\global\long\def\Class#1{j_{#1}}%
\global\long\def\Action#1{a_{#1}}%
\global\long\def\PseudoAction#1{h_{#1}}%
\global\long\def\Filtration#1{\mathcal{F}_{#1}}%
\global\long\def\History#1{\mathcal{H}_{#1}}%
\global\long\def\XX#1{\boldsymbol{X}_{#1}}%
\global\long\def\OptAction#1{a_{#1}^{\star}}%
\global\long\def\Impute#1{\check{\beta}_{#1}}%
\global\long\def\Maxeigen#1{\lambda_{\max}\!\left(#1\right)}%
\global\long\def\Mineigen#1{\lambda_{\min}\!\left(#1\right)}%
\global\long\def\Trace#1{\text{Tr}\left(#1\right)}%
\global\long\def\SetofContexts#1{\mathcal{X}_{#1}}%
\global\long\def\Ridgebeta#1{\widehat{\beta}_{#1}^{ridge}}%
\global\long\def\Order#1{\langle#1\rangle}%
\global\long\def\AdmitRound#1{\tau(#1)}%

\twocolumn[
\icmltitle{Improved Algorithms for Multi-period Multi-class Packing Problems with~Bandit~Feedback}

\icmlsetsymbol{equal}{*}

\begin{icmlauthorlist}
\icmlauthor{Wonyoung Kim}{CU}
\icmlauthor{Garud Iyengar}{CU}
\icmlauthor{Assaf Zeevi}{CU}
\end{icmlauthorlist}

\icmlaffiliation{CU}{Columbia University, New York, NY, USA}

\icmlcorrespondingauthor{Wonyoung Kim}{wk2389@columbia.edu}

\icmlkeywords{Resource constraints}
\vskip 0.3in
]

\printAffiliationsAndNotice{}  
\begin{abstract}

We consider the linear contextual multi-class multi-period packing problem~(LMMP) where the goal is to pack items such that the total vector of consumption is below a  given budget vector and the total value is as large as possible.
We consider the setting where the reward and the consumption vector associated with each action is a class-dependent linear function of the context, and the decision-maker receives bandit feedback.  
LMMP includes linear contextual bandits with knapsacks and online revenue management as special cases.
We establish a new estimator which guarantees a faster convergence rate, and consequently, a lower regret in LMMP. 
We propose a bandit policy that is a closed-form function of said estimated parameters. 
When the contexts are non-degenerate, the regret of the proposed policy is sublinear in the context dimension, the number of classes, and the time horizon~$T$ when the budget grows at least as $\sqrt{T}$.
We also resolve an open problem posed in~\citet{agrawal2016linear}, and extend the result to a multi-class setting. 
Our numerical experiments clearly demonstrate that the performance of our policy is superior to other benchmarks in the literature.
\end{abstract}


\section{Introduction}
In the multi-period packing problem~(MPP) the decision-maker ``packs'' the arrivals so that the total consumption across a set of resources is below a given budget vector and the reward is maximized. 
A variant of the packing problem, where items consume multiple resources and the decisions must be made sequentially with bandit feedback for a fixed time horizon, is known as bandits with knapsacks
\citep{agrawal2014bandits,badanidiyuru2018bandits,immorlica2019adversarial}.
MPPs also arise in online revenue management~\citep{besbes2012blind,ferreira2018online}.  
MPPs in the literature assume that all arrivals belong to a single class. 
However, in several application domains (e.g., operations, healthcare, and e-commerce), the arrivals are heterogeneous, and personalizing decisions to each distinct population or class is of paramount importance. 
In this paper, we consider a class of linear \emph{multi-class} multi-period packing problems~(LMMP).
At each round, there is a single arrival that belongs to one of $J$ classes, and the decision-maker observes the $d$-dimensional context and the cost for $K$ different available actions. 
The outcome of selecting an action is a random sample of the reward and a consumption vector for $m$ resources with an expected value that is a class-dependent linear function of the $d$-dimensional contexts.
The goal of the problem is to minimize the cumulative regret over a time horizon $T$ while ensuring that the total resource consumed is at most $B$.

The LMMP problem is a generalization of several~problems including linear contextual bandits with knapsacks (LinCBwK) introduced by~\citet{agrawal2016linear}.
They proposed an online mirror descent-based algorithm that achieves $\tilde{O}(OPT/B\cdot d\sqrt{T})$ regret when the budget $B$ for each of the $m$ resources is $\Omega(\sqrt{d}T^{3/4})$, where $OPT$ is the reward obtained by the oracle policy. 
Although the regret bound is meaningful for $B=\Omega(d\sqrt{T})$, establishing the regret bound for smaller budget values was left as an open problem.
\citet{chu2011contextual} established a regret bound sublinear in $d$ for the linear contextual bandit setting, which is a special case of LinCBwK with no budget constraints. 
Thus, the following question remained open: 
\textit{``Is there an algorithm for  LinCBwK  that achieves sublinear
dependence on $d$ with budget $B=\Omega(\sqrt{T})$?'' }

We propose a novel algorithm and an improved estimation strategy that settles this open problem and generalizes the result to the more general class of LMMP.
The proposed algorithm achieves $\tilde{O}(OPT/B \sqrt{JdT})$ regret with budget $B=\Omega(\sqrt{JdT})$ under non-degenerate contexts.
While regret of the existing algorithms grows linearly in the number of classes $J$, our estimator is able to pool data from different classes and avoids linear dependence on $J$. 
To reiterate, the improved regret bound results from the novel estimator which has faster convergence rates. 

Our main contributions are summarized as follows: 
\begin{itemize}
    \item We propose a new problem class -- linear multi-class multi-period packing problems (LMMP). 
    This problem generalizes a variety of problems including LinCBwK and online revenue management problems to the multi-class setting.
    \item We propose a novel estimator that uses contexts for \emph{all} actions (including the contexts in skipped rounds) and yields $O(\sqrt{Jd/n})$ convergence rate for $J$ classes, context dimension~$d$, and $n$ admitted arrivals (Theorem \ref{thm:u_convergence}).
    \item We propose a novel \texttt{AMF} (Allocate to the Maximum First) algorithm which achieves $\tilde{O}(OPT/B\sqrt{JdT})$ regret with budget $B=\Omega(\sqrt{JdT})$ where $OPT$ is the reward obtained by oracle policy (Theorem \ref{thm:regret_bound}). 
    For the single class setting with $J=1$, we improve the existing bound by $\sqrt{d}$ and show that the bound is valid when $B=\Omega(\sqrt{dT})$, thus resolving an open problem posed in \citet{agrawal2016linear} regarding LinCBwK.
    \item We evaluate our proposed algorithm on a suite of synthetic experiments and demonstrate its superior performances.
\end{itemize}
All proofs omitted from the front matter can be found in the Appendix.


\section{Related Works}
There are two streams of work that are relevant for LMMP. 
In online revenue management literature, \citet{gallego1994optimal} introduced the dynamic pricing problem where the demand is a known function of price (action). 
\citet{besbes2009dynamic} and \citet{besbes2012blind} extended the problem under unknown demands with multiple resource constraints.
\citet{ferreira2018online} proposed a Thompson sampling-based algorithm and extended it to contextual bandits with knapsacks.  
When the expected demand is a linear function of the price vector, the dynamic pricing problem is a special case of linear contextual bandits with knapsack~(LinCBwK) proposed by~\citet{agrawal2016linear}.

The LinCBwk is a common generalization of bandits with knapsacks~\citep{badanidiyuru2018bandits,immorlica2019adversarial,li21symmetry} and online stochastic packing problems~\citep{feldman2010online,agrawal2014fast,devanur2011near}.  
Recently, \citet{sankararaman2021bandits} proved a logarithmic regret
bound for LinCBwK when there exists a problem-dependent gap between the reward of the optimal action and the other actions.  
Instead of the gap assumption, we require non-degeneracy of the stochastic contexts (see Assumption~\ref{assum:positive_definiteness} for a precise definition) to obtain a regret bound sublinear in $d$ and extends to the case when the contexts are generated from $J$ different class.

\citet{amani2019linear} proposed a variant of LinCBwK where the selected action must satisfy a single constraint with high probability in all rounds, i.e.,  LinCBwK with anytime constraints.
\citet{moradipari21safethompson} and \citet{pacchiano21linear} proposed a Thompson sampling-based algorithm and an upper confidence bound-based algorithm, respectively, for LinCBwK with a single anytime constraint. 
\citet{liu2021efficient} highlighted the difference between global and anytime constraints and proposed a pessimistic-optimistic algorithm for the anytime constraints.
We focus on the global constraints; however, we note that the extension to the anytime constraints is straightforward with minor modifications.

\subsection{Notation}
Let $\Real_{+}$ denote the set of positive real numbers.
For two real numbers $a,b\in\Real$, we write $a\wedge b:=\min\{a,b\}$ and $a\vee b:=\max\{a,b\}$. 
For a natural number $N \in\mathbb{N}$, let $[N] := \{1,\ldots, N\}$.

\section{Linear Multi-period Packing Problem}

Let $[J]$ denote the set of classes with arrival probabilities $p=\{p_j\}_{j \in [J]}$, where $p_{\min}:=\min_{j\in[J]}p_{j}>0$. 
For simplicity, we assume that the class arrival probabilities $p$ are known while the same theoretical results can be obtained when the probabilities are unknown to the decision-maker (See Section~\ref{subsec:unknown_p} for details).
In each round $t\in[T]$, the covariates $\{\Context jkt\in[0,1]^{d}:k\in[K]\}$ are drawn from an unknown class-specific distribution $\Classdistribution j$ and the decision-maker observes an arrival of the form $(\Class t,\{\Context{\Class t}kt :k\in[K]\})$, where $\Class t\in[J]$ is the arrived class.
Upon observing the arrival, the decision-maker can either take one of $K$ different actions or skip the arrival.
When the arrival is skipped, the decision-maker does not obtain any rewards or consume any resources. 
When the decision-maker chooses an action $\Action t\in[K]$, the reward and consumption of the resource are given by
\begin{align*}
  \CE{\Reward{\Class t}{\Action t}t}{\History t}
  & =\left\{ \Parameter{\Class t}\right\} ^{\top}\Context{\Class t}{\Action t}t\in[-1,1],\\
  \CE{\Consumption{\Class t}{\Action t}t}{\History t}
  & =\left\{
    \RParameter{\Class
    t}\right\}
    ^{\top}\Context{\Class
    t}{\Action
    t}t\in[0,1]^{m}, 
\end{align*}
for some unknown class-specific parameters $\Parameter j\in[0,1]^{d}$ and $\RParameter j\in[0,1]^{d\times m}$.
The sigma algebra $\History t$ is generated by the class-specific variables $\{\Class s,\Context{\Class s}ks:s\in[t],k\in[K]\}$, actions $\{\Action s:s\in\mathcal{A}_t\}$, consumption vectors $\{\Consumption{\Class s}{\Action s}s:s\in\mathcal{A}_{t-1}\}$ and rewards $\{\Reward{\Class s}{\Action s}t:s\in\mathcal{A}_{t-1}\}$, where $\mathcal{A}_{t}$ is the rounds admitted by the decision-maker until round $t$.
The process terminates at the horizon $T$ or runs out of budget $B\in\Real_{+}^{m}$ for some resources $r\in[m]$.
The problem reduces to LinCBwK when the number of class is $J=1$.

LMMP allows each class to have a different set of contexts and parameters, which is required in many applications such as e-commerce, clinical trials, and dynamic pricing.
For example, consider an e-commerce setting with $J$ classes of customers with $J$ different preferences.  
At each decision point $t$, the decision-maker must make one of $K$ different $d$-dimensional offers: the $k$-th offer will result in a random consumption of $m$ resources with mean consumption vector $(W^{(j_t)}_{\star})^{\top}\mathbf{x}_{k,t}^{(j_t)}$ and results in random reward with mean $(\theta^{(j_t)}_{\star})^\top \mathbf{x}_{k,t}^{(j_t)}$. 
Note that the context $\mathbf{x}_{k,t}^{(j_t)}$ can include
the price charged to the class $j_t$ customers as one of the components.
This feature of LMMP is novel to the literature and allows for a personalized decision for each class.

Let $\rho \in \Real_{+}^{m}$ denote per-period budget vector for $m$ resources.
Without loss of generality, one can assume that $\rho=(B/T) \mathbf{1}_m$, by rescaling $\RParameter{j}$.
We assume that $\rho$ is known to the decision-maker, which is essential to target the correct optimal policy.
Unlike the unconstrained finite horizon bandit problems, the optimal policy of LMMP depends on $\rho$.
Without knowledge of $\rho$, the bandit policy cannot converge to the correct optimal policy, which leads to a cumulative regret linear in $T$.
As a result, many related problems, such as LinCBwk \citep{agrawal2016linear} and online revenue management \citep{ferreira2018online}, commonly assume knowledge of both $B$ and $T$.

In our work, we assume that $B$ is possibly unknown at first but known at the end of the round.
Specifically, the decision-maker only observes the initial inventory $B_1$, which will increase to $B$ before the end of the horizon $T$.
This scenario is relevant to online inventory management, where a product's inventory is supplied at different time points. 
This assumption is more practical than in \citet{agrawal2016linear} where $B$ and $OPT$ must be known to the decision-maker.
When $OPT$ is unknown, \citet{agrawal2016linear} proposed to estimate $OPT$ with $\sqrt{T}$ number of rounds, which requires the knowledge of $T$ and budget $B=\Omega(\sqrt{d} T^{\frac{3}{4}} )$.
Instead of estimating $OPT$, we use $\rho$ to avoid the required budget $B=\Omega(\sqrt{d}T^{\frac{3}{4}})$.


We benchmark the performance of the decision-maker's policy relative to that of an oracle who knows the distributions $\{\Classdistribution j:j\in[J]\}$ and the parameters $\{\Parameter j,\RParameter j:j\in[J]\}$, but does not know the arrivals $\{(j_{t},\Context jkt):t\in[T]\}$ a-priori.  
In this case, the optimal static policy for the oracle $\{\OptPolicy
jk:j\in[J],k\in[K]\}$ is the solution to the following optimization problem: 
\begin{equation}
  \begin{split}
    \max_{\pi_{k}^{(j)}}
    & \sum_{j=1}^{J}\sum_{k=1}^{K}p_{j}\pi_{k}^{(j)}\Expectation_{\mathbf{x}_{k}\sim
    \Classdistribution 
    j}\left[\left\{ \Parameter j\right\}
    ^{\top}\mathbf{x}_{k}\right]\\ 
  \text{s.t.}  &\sum_{j=1}^{j}\sum_{k=1}^{K}p_{j}\pi_{k}^{(j)}\Expectation_{\mathbf{x}_{k}\sim
    \Classdistribution 
    j}\!\left[\left\{ \RParameter j\!\right\}
    ^{\top}\!\!\mathbf{x}_{k}\right]\!\le\rho,\\ 
  & \sum_{k=1}^{K}\pi_{k}^{(j)}\le1,\:\forall j\in[J],\\  
  & \pi_{k}^{(j)}\ge0,\:\forall j\in[J],\forall k\in[K],
\end{split}
\label{eq:oracle_problem}
\end{equation}
Note that the oracle policy depends on $\rho$ and thus both $B$ and $T$.
Then the expected reward obtained by the oracle is
\begin{equation*}
    OPT:= T\sum_{j=1}^{J}\sum_{k=1}^{K}p_{j}\pi_{k}^{\star(j)}
  \Expectation_{\mathbf{x}_{k}
  \sim\Classdistribution
  j}\left[\left\{ \Parameter j\right\}
  ^{\top}\mathbf{x}_{k}\right].
\end{equation*}
Let $\pi:=\{\Policy jkt:j\in[J],k\in[K],t\in[T]\}$ denote the adapted
(randomized) control policy of the decision-maker, i.e. she chooses action $k\in[K]$ when the arrival at time $t \in [T]$ belongs to 
class $j\in[J]$.
Note that $\sum_{k=1}^{K}\Policy jkt \le 1$ in order to allow the decision-maker to skip an arrival and save the inventory for later use.
Our goal is to compute a policy that minimizes the cumulative regret $\mathcal{R}_{T}^{\pi}$ defined as 
\[
\mathcal{R}_{T}^{\pi}:=OPT-\Expectation\left[\sum_{t=1}^{T}R_{t}^{\pi}\right], 
\]
where $R_{t}^{\pi}:=\sum_{k=1}^{K}\pi_{k,t}^{(\Class
   t)}\Expectation\left[\left\{ \Parameter{\Class t}\right\}
   ^{\top}\Context{\Class t}kt\right]$ 
is the expected reward obtained by policy $\pi$ at time $t$. 

For the LMMP problem, we assume the following regularity conditions
on the stochastic processes.

\begin{assumption}\label{assum:error} (Sub-Gaussian errors) 
For each $t\in[T]$, the error of the reward $\Error kt=\Reward{\Class t}kt-\left\{ \Parameter{\Class t}\right\} ^{\top}\Context{\Class t}kt$
is conditionally zero-mean $\sigma_{r}$-sub-Gaussian for a fixed
constant $\sigma_{r}\ge0$.
In other words, $\CE{\exp\left(v\Error kt\right)}{\History t}\le\exp\left(\frac{v^{2}\sigma_{r}^{2}}{2}\right)$
for all $v\in\Real$. 
For the consumption vectors, $\CE{\mathbf{v}^{\top}\{ \Consumption{\Class t}kt-(W_{\star}^{(\Class t)})^{\top}\Context{\Class t}kt\} }{\History t}\le\exp(\frac{\norm{\mathbf{v}}_{2}^{2}\sigma_{b}^{2}}{2})$ for all $\mathbf{v}\in\Real^{m}$.
\end{assumption}

\begin{assumption}\label{assum:independent_contexts} (Independently distributed contexts) The set of contexts $\{\Context jkt:k\in[K]\}$ are generated independently over $t\in[T]$.
The contexts and cost in the same round and class can be correlated with each other.
\end{assumption}

\begin{assumption}\label{assum:positive_definiteness} (Positive
definiteness of average covariances) For each $t\in[T]$ and $j\in[J]$,
there exists $\alpha>0$, such that
\[
\Mineigen{\Expectation\text{\ensuremath{\left[\frac{1}{K}\sum_{k=1}^{K}\Context jkt\left\{  \Context jkt\right\}  ^{\top}\right]}}}\ge\alpha.
\]
\end{assumption}

Assumptions~\ref{assum:error} and~\ref{assum:independent_contexts} are standard in stochastic contextual bandits with knapsacks literature\citep{agrawal2016linear,sankararaman2021bandits,sivakumar2022smoothed}.
In the multi-class scenario, Assumption \ref{assum:independent_contexts} implies that all the
contexts are drawn independently over time steps, but their distribution may vary depending on the class.
The independence of contexts is supported by \citet{bastani2020online} and \citet{kim2021doubly}, as this assumption is practical in real-world applications such as clinical trials where patient health covariates are independent of those of other patients.
Assumption \ref{assum:positive_definiteness} implies that the density of the covariate distribution is non-degenerate.
This assumption is necessary to estimate all entries of the parameters through linear regression in the statistics literature.
In our work, we use the estimator $\check{\Theta}_n$ for $\Theta_{\star}$ in pseudo-rewards as defined in equation (5).
Therefore, the convergence of $\|\check{\Theta}_n-\Theta_{\star}\|_{2}$ and the accuracy of the pseudo-rewards are heavily dependent on Assumption~\ref{assum:positive_definiteness}.
Recent literature on contextual bandits (without constraints) has utilized Assumption~\ref{assum:positive_definiteness} to improve the dependency of $d$ on the regret bound \citep{bastani2020online,kim2021doubly,bastani2021mostly,oh2021sparsity}.
As noted by \citet{kannan2018smoothed} and \citet{sivakumar2020structured,sivakumar2022smoothed}, contexts with measurement errors provide enough variability to satisfy Assumption~\ref{assum:positive_definiteness}.

\section{Proposed Method}

In this section, we present our proposed estimator for the parameters
$\{\Parameter j,\RParameter j:j\in[J]\}$ and the proposed closed form bandit policy.

\subsection{Proposed Estimator\label{subsec:estimator}}

In sequential decision-making problems with contexts, the decision-maker observes the contexts for all actions, \emph{but} the reward for only selected actions, i.e. the rewards for unselected actions remain missing.
A statistical missing data technique called the doubly robust (DR) method is employed to handle the missing rewards for linear contextual bandit problem \citep{kim2019doubly,dimakopoulou2019balanced,kim2021doubly,kim2022squeeze,kim2022double}.
However, extensions to LinCBwK or LMMP problems have not been explored yet.

To apply the DR method to the LMMP problem, we modify the randomization technique proposed by~\citet{kim2022squeeze}.
For each $n\in\Natural$, let $\AdmitRound n$ be the round when the $n$-th admission happens (recall that the bandit policy allows for skipping some arrivals). 
Clearly, $n\le\AdmitRound n<\AdmitRound{n+1}$ holds. 
Let 
\[
\Theta_{\star}:=\begin{pmatrix}
\Parameter 1\\
\vdots\\
\Parameter J
\end{pmatrix},
\mathbf{W}_{\star}:=\begin{pmatrix}\RParameter 1\\
\vdots\\
\RParameter J
\end{pmatrix},
\tilde{X}_{k,n}:=\begin{pmatrix}\mathbf{0}_{d}\\
\vdots\\
\Context{\Class{\AdmitRound n}}k{\AdmitRound n}\\
\mathbf{0}_{d}
\end{pmatrix}
\]
denote the stacked parameter vectors, and zero padded contexts where
$\Context{\Class{\AdmitRound n}}k{\AdmitRound n}$ 
is located after the $j_{\tau(n)}-1$ of $\mathbf{0}_{d}$ vectors. 
Then the score for the ridge estimator for $\Theta^{\star}$ at round $\AdmitRound n$ is: 
\begin{align*}
 & \sum_{\nu=1}^{n}\left(\Reward{\Class{\AdmitRound{\nu}}}{\Action{\AdmitRound{\nu}}}{\AdmitRound{\nu}}-\Theta^{\top}\tilde{X}_{\Action{\AdmitRound{\nu}},\nu}\right)\tilde{X}_{\Action{\AdmitRound{\nu}},\nu}\\
 & =\sum_{\nu=1}^{n}\sum_{k=1}^{K}\Indicator{\Action{\AdmitRound{\nu}}=k}\left(\Reward{j_{\AdmitRound{\nu}}}k{\AdmitRound{\nu}}-\Theta^{\top}\tilde{X}_{k,\nu}\right)\tilde{X}_{k,\nu},
\end{align*}
where $\Theta\in\Real^{J\cdot d}$. 
Dividing the score by the probability $\Policy{\Class{\AdmitRound{\nu}}}k{\AdmitRound{\nu}}$ gives the
inverse probability weighted (IPW) score, 
\[
\sum_{\nu=1}^{n}\sum_{k=1}^{K}\frac{\Indicator{\Action{\AdmitRound{\nu}}=k}}{\Policy{\Class{\AdmitRound{\nu}}}k{\AdmitRound{\nu}}}\left(\Reward{j_{\AdmitRound{\nu}}}k{\AdmitRound{\nu}}-\Theta^{\top}\tilde{X}_{k,\nu}\right)\tilde{X}_{k,\nu}.
\]
To obtain the DR score, \citet{bang2005doubly,kim2021doubly} proposed
to subtract the nuisance tangent space generated by an imputed estimator
$\check{\Theta}$:
\[
\sum_{\nu=1}^{n}\sum_{k=1}^{K}\frac{\Indicator{\Action{\AdmitRound{\nu}}=k}}{\Policy{\Class{\AdmitRound{\nu}}}k{\AdmitRound{\nu}}}\left(\tilde{X}_{k,\nu}^{\top}\check{\Theta}-\tilde{X}_{k,\nu}^{\top}\Theta\right)\tilde{X}_{k,\nu},
\]
from the IPW score. Then the following DR score
\begin{equation}
\sum_{\nu=1}^{n}\sum_{k=1}^{K}\left\{ \DRY{\check{\Theta}}k{\AdmitRound{\nu}}-\tilde{X}_{k,\nu}^{\top}\Theta\right\} \tilde{X}_{k,\nu},\label{eq:DR_score}
\end{equation}
is obtained where 
\begin{equation}
\DRY{\check{\Theta}}k{\nu}\!\!:=\!\!\frac{\Indicator{\Action{\AdmitRound{\nu}}\!\!=\!k}}{\Policy{\Class{\AdmitRound{\nu}}}k{\AdmitRound{\nu}}}\Reward{j_{\AdmitRound{\nu}}}k{\AdmitRound{\nu}}+\!\left\{ \!1\!-\!\!\frac{\Indicator{\Action{\AdmitRound{\nu}}\!\!=\!k}}{\Policy{\Class{\AdmitRound{\nu}}}k{\AdmitRound{\nu}}}\!\right\} \!\!\tilde{X}_{k,\nu}^{\top}\check{\Theta}.
\label{eq:DRY}
\end{equation}
The score \eqref{eq:DR_score} has a similar form with the score equation for the ridge estimator. 
The difference with the ridge estimator is that it uses contexts for all actions $k\in[K]$ with the pseudo-reward $\DRY{\check{\Theta}}k{\nu}$ which is unbiased, i.e., $\Expectation[\DRY{\check{\Theta}}k{\nu}]=\Expectation[\Reward{\Class{\AdmitRound{\nu}}}k{\AdmitRound{\nu}}]$,
for any given $\check{\Theta}\in\Real^{J\cdot d}$. 
Adding the $\ell_2$ regularization norm and solving \eqref{eq:DR_score} leads to the DR estimator:
\[
\left(\sum_{\nu=1}^{n}\sum_{k=1}^{K}\!\tilde{X}_{k,\nu}\tilde{X}_{k,\nu}^{\top}\!\!+\! I_{J\cdot d}\!\right)^{\!\!-\!1}\!\!\!\!\left(\sum_{\nu=1}^{n}\sum_{k=1}^{K}\!\tilde{X}_{k,\nu}\DRY{\check{\Theta}}k{\AdmitRound{\nu}}\!\!\right).
\]
The main advantage of the DR estimator is that it uses contexts from \emph{all} $K$ actions. However, in our policy, some
$\Policy{\Class{\AdmitRound{\nu}}}k{\AdmitRound{\nu}}$ 
can be zero, and therefore, the pseudo-reward \eqref{eq:DRY} is not defined.
To handle this problem, we propose to introduce a random variable.  
After taking an action at round $\AdmitRound{\nu}$ and observing the selected action $\Action{\AdmitRound{\nu}}$, the decision-maker samples $\PseudoAction{\nu}$ from the distribution:
\begin{equation}
\begin{split}
\NewProb k{\nu}\!\!:=&\CP{\PseudoAction{\nu}=k}{\History{\AdmitRound n}}\!\!\\&=\!\!\begin{cases}
1\!-\!\frac{16(K-1)\log\left(\frac{dJ}{\delta}\right)}{\Mineigen{F_{\nu}}}\!\! & k\!=\!\Action{\AdmitRound{\nu}}\\
\frac{16\log\left(\frac{dJ}{\delta}\right)}{\Mineigen{F_{\nu}}} & k\!\neq\!\Action{\AdmitRound{\nu}}
\end{cases}
\end{split}
\label{eq:pseudo_action_probability}
\end{equation}
where $F_{\nu}\!:=\!\sum_{i,k=1}^{\nu,K}\tilde{X}_{k,i}\tilde{X}_{k,i}^{\top}\!+\!16d(K\!-\!1)\log\left(\frac{dJ}{\delta}\right)I_{J\cdot d}$ is the Gram matrix of contexts from $\nu$ admitted rounds and $\delta\in(0,1)$ is the confidence level. 
We would like to emphasize that $\PseudoAction{\nu}$ is sampled after observing the actions $\Action{\AdmitRound{\nu}}$ and does not affect the policies until round $\AdmitRound{\nu}$.

Sampling the random variables $\PseudoAction{\nu}$ after choosing actions is motivated by~\citet{kim2022squeeze} which uses bootstrap methods~\citep{efron1994introduction} and resampling methods~\citep{good2006resampling}.
To obtain the unbiased pseudo-rewards similar to~\eqref{eq:DRY}, we resample the action from another distribution with non-zero probabilities. 
The probabilities $\{\NewProb k{\nu}:k\in[K]\}$ are designed to control the level of exploration and exploitation for future rounds based on the ratio of confidence level to the number of admitted rounds. 
When the minimum eigenvalue of $F_{\nu}$ is small compared to $\log(1/\delta)$, the distribution of $\PseudoAction{\nu}$ is less concentrated on $\Action{\AdmitRound{\nu}}$ and tends to explore other actions.
As $\nu$ increases, the probabilities $\{\NewProb k{\nu}:k\in[K]\}$  concentrates on $\Action{\AdmitRound{\nu}}$, and the decision-maker tends to exploit.

Since we obtain non-zero probabilities $\{\NewProb k{\nu}:k\in[K],\nu\in[n]\}$, we define novel unbiased pseudo-rewards:
\begin{equation}
\TildeR k{\nu}\!:=\!\!\frac{\Indicator{\PseudoAction{\nu}\!=\!k}}{\NewProb k{\nu}}\Reward{\Class{\AdmitRound{\nu}}}k{\AdmitRound{\nu}}\!+\left\{ 1\!\!-\!\frac{\Indicator{\PseudoAction{\nu}\!=\!k}}{\NewProb k{\nu}}\!\right\} \!\tilde{X}_{k,\nu}^{\top}\!\check{\Theta}_{n},\label{eq:pseudo_rewards}
\end{equation}
where the imputation estimator $\check{\Theta}_{n}$ is an IPW estimator with new probabilities:
\begin{align*}
\check{\Theta}_{t}:= & A_{n}^{-1}\Biggl\{\sum_{\nu\in\Psi_{n}}\sum_{k=1}^{K}\frac{\Indicator{\PseudoAction{\nu}\!=\!k}}{\NewProb k{\nu}}\tilde{X}_{k,\nu}\Reward{\Class{\AdmitRound{\nu}}}k{\AdmitRound{\nu}}\\
 & \qquad+\sum_{\nu\notin\Psi_{n}}\tilde{X}_{\Action{\AdmitRound{\nu}},\nu}\Reward{\Class{\AdmitRound{\nu}}}{\Action{\AdmitRound{\nu}}}{\AdmitRound{\nu}}\Biggr\},\\
A_{n}:= & \sum_{\nu\in\Psi_{n}}\sum_{k=1}^{K}\frac{\Indicator{\PseudoAction{\nu}\!=\!k}}{\NewProb k{\nu}}\tilde{X}_{k,\nu}\tilde{X}_{k,\nu}^{\top}\\
 & +\sum_{\nu\notin\Psi_{n}}\tilde{X}_{\Action{\AdmitRound{\nu}},\nu}\tilde{X}_{\Action{\AdmitRound{\nu}},\nu}^{\top}+I_{J\cdot d},\\
\Psi_{n}:= & \left\{ \nu\in[n]:\PseudoAction{\nu}=\Action{\AdmitRound{\nu}}\right\} .
\end{align*}
The set $\Psi_{n}$ is introduced because we cannot observe $\frac{\Indicator{\PseudoAction{\nu}=k}}{\NewProb k{\nu}}\Reward{\Class{\AdmitRound{\nu}}}k{\AdmitRound{\nu}}$ in case of $\PseudoAction{\nu}\neq\Action{\AdmitRound{\nu}}$. 
In other words, we use the pseudo-rewards in \eqref{eq:pseudo_rewards} only at the rounds that satisfy $\PseudoAction{\nu}=\Action{\AdmitRound{\nu}}$.
Then our estimator with $n$ admitted samples is defined as
\begin{equation}
\begin{split}\widehat{\Theta}_{n}\!:= & V_{n}^{-1}\!\!\left\{
    \!\sum_{\nu\in\Psi_{n}}\sum_{k=1}^{K}\tilde{X}_{k,\nu}\TildeR k{\nu}\!+\!\!\sum_{\nu\notin\Psi_{n}}\!\tilde{X}_{\Action{\AdmitRound{\nu}},\nu}\Reward{\Class{\AdmitRound{\nu}}\!}{\Action{\AdmitRound{\nu}}}{\AdmitRound{\nu}}\!\right\}
  \\ 
V_{n}:= & \!\sum_{\nu\in\Psi_{n}}\sum_{k=1}^{K}\tilde{X}_{k,\nu}\tilde{X}_{k,\nu}^{\top}\!+\!\!\sum_{\nu\notin\Psi_{n}}\tilde{X}_{\Action{\AdmitRound{\nu}},\nu}\tilde{X}_{\Action{\AdmitRound{\nu}},\nu}^{\top}\!+\!I_{J\cdot d}.
\end{split}
\label{eq:r_estimator}
\end{equation}
\vskip -0.1in

Analogous to the construction of \eqref{eq:r_estimator}, we can also define the estimator for the resource consumption parameters $\{\RParameter j:j\in[J]\}$,
\begin{equation}
\widehat{\mathbf{W}}_{n}\!:=\!V_{n}^{-1}\!\Bigg[\!\sum_{\nu\in\Psi_{n}}\sum_{k=1}^{K}\tilde{X}_{k,\nu}\TildeB k{\nu}^{\top}+\!\!\sum_{\nu\notin\Psi_{n}}\!\!\tilde{X}_{\Action{\AdmitRound{\nu}},\nu}\mathbf{b}_{\Action{\AdmitRound{\nu}}\AdmitRound{\nu}}^{(\Class{\AdmitRound{\nu}})\top}\!\Bigg],
\label{eq:w_estimator}
\end{equation}
where the pseudo-consumption vectors and the imputation estimator are
\begin{align*}
\TildeB k{\nu}\!&:=\!\frac{\Indicator{\PseudoAction{\nu}\!=\!k}}{\NewProb k{\nu}}\Consumption{\Class{\AdmitRound{\nu}}}{\Action{\AdmitRound{\nu}}}{\AdmitRound{\nu}}\!+\!\!\left\{1\!-\!\frac{\Indicator{\PseudoAction{\nu}\!=\!k}}{\NewProb k{\nu}}\!\right\} \!\check{\mathbf{W}}_{n}^{\top}\tilde{X}_{k,\nu},\\
\check{\mathbf{W}}_{n}&:=A_{n}^{-1}\Bigg[\sum_{\nu\in\Psi_{n}}\sum_{k=1}^{K}\frac{\Indicator{\PseudoAction{\nu}\!=\!k}}{\NewProb k{\nu}}\tilde{X}_{k,\nu}\left\{\Consumption{\Class{\AdmitRound{\nu}}}k{\nu}\right\}^{\top}\\&\qquad+\sum_{\nu\notin\Psi_{n}}\tilde{X}_{\Action{\AdmitRound{\nu}},\nu}\left\{ 
  \Consumption{\Class{\AdmitRound{\nu}}}{\Action{\AdmitRound{\nu}}}{\AdmitRound{\nu}}\right\}
  ^{\top}\Bigg]. 
\end{align*}
The two estimators use the novel Gram matrix $V_{n}$ defined in \eqref{eq:r_estimator} consisting of contexts from \emph{all} $K$ actions. 
Now, we present estimation error bounds normalized by the novel Gram matrix $V_{n}$.
\begin{theorem}
\label{thm:self} (Self-normalized bound for the estimator) Suppose
Assumptions~\ref{assum:error} and~\ref{assum:independent_contexts} hold. 
For each $t\in[T]$, let $n_{t}$ denote the number of admitted arrivals until round $t$ and $\Psi_{n_{t}}:=\{\nu\in[n_{t}]:\PseudoAction{\nu}=\Action{\AdmitRound{\nu}}\}$,
where $\PseudoAction{\nu}$ is defined in \eqref{eq:pseudo_action_probability}.
Suppose $F_{n_{t}}:=\sum_{\nu=1}^{n_{t}}\sum_{k=1}^{K}\tilde{X}_{k,\nu}\tilde{X}_{k,\nu}^{\top}+16d(K-1)\log\frac{Jd}{\delta}I_{J\cdot d}$
satisfies
\begin{equation}
\begin{split}
&\Mineigen{F_{n_{t}}}
\\
&\ge4Kd\left\{ \sum_{\nu=1}^{n_{t}}\frac{144\left(K-1\right)\log\left(\frac{Jd}{\delta}\right)}{\Mineigen{F_{\nu}}}+35\log\frac{Jd}{\delta}\right\}, 
\end{split}
\label{eq:self_condition}
\end{equation}
for $\delta\in(0,1)$. 
For each $r\in[m]$ , let $\widehat{\mathbf{W}}_{n_{t},r}$
and $\mathbf{W}_{\star,r}$ be the $r$-th column of  $\widehat{\mathbf{W}}_{n_{t}}$
and $\mathbf{W}_{\star}$, respectively. 
Denote $\Selfbound{\sigma}{\delta}:=8\sqrt{Jd}+96\sigma\sqrt{Jd\log\frac{4}{\delta}}$.
Then with probability at least $1-4(m+1)\delta$,
\begin{equation}
\begin{split}\norm{\StackedEstimator{n_{t}}-\Theta^{*}}_{V_{n_{t}}}\le & \Selfbound{\sigma_{r}}{\delta},\\
\max_{r\in[m]}\norm{\widehat{\mathbf{W}}_{n_{t},r}-\mathbf{W}_{\star,r}}_{V_{n_{t}}}\le & \Selfbound{\sigma_{b}}{\delta}.
\end{split}
\label{eq:self_bound}
\end{equation}
\end{theorem}

Compared to the self-normalized bound in~\citet{abbasi2011improved} uses the Gram matrix consisting of selected contexts only, our bounds are normalized by $V_{n_{t}}$ 
This change in the Gram matrix enables us to develop a fast convergence rate. 
The condition~\eqref{eq:self_condition} is required for the eigenvalues of the Gram matrix $F_{n_{t}}$ to be large so that the probability $\NewProb{\Action{\AdmitRound{\nu}}}{\nu}$ is large and the estimators use the pseudo rewards and pseudo consumption
vectors for most of the rounds.  
We show in Lemma~\ref{lem:exploration_bound} that the
condition~\eqref{eq:self_condition} requires at most
rounds logarithmic in $T$, and does not affect the main order of the regret bound.

Using the novel estimators, we define the estimates for utility and resource consumption. 
Denote $\mathcal{C}_{t}^{(j)}:=\{s\in[t]:\Class{s}=j\}$ and \begin{equation}
\begin{split}
\widehat{u}_{k,t}^{(j)}\!:=\! &
\abs{\mathcal{C}_{t}^{(j)}}^{-1}\sum_{s\in\mathcal{C}_{t}^{(j)}}\left\{
      \Estimator j{t-1}\right\} ^{\top}\Context jks,\\ 
\widehat{\mathbf{b}}_{k,t}^{(j)}:=\! &
\abs{\mathcal{C}_{t}^{(j)}}^{-1}\sum_{s\in\mathcal{C}_{t}^{(j)}}\left\{
  \widehat{W}_{t-1}^{(j)}\right\} ^{\top}\Context jks. 
\end{split}
\label{eq:u_b_hat}
\end{equation}
The estimates \eqref{eq:u_b_hat} use the average of contexts in the same class to estimate the expected value over the context distribution. 
In this way, the decision-maker effectively uses previous contexts in all rounds including the \emph{skipped rounds}. 
Next, we establish a convergence rate for the estimators $\widehat{u}_{k,t}^{(j)}$ and $\widehat{\mathbf{b}}_{k,t}^{(j)}$. 

\begin{theorem}
\label{thm:u_convergence} (Convergence rate for the estimates) Suppose
Assumptions \ref{assum:error}-\ref{assum:positive_definiteness}
hold. 
Denote the expected utility $\OptUtility
jk:=\Expectation_{\mathbf{x}_{k}\sim\Classdistribution
  j}\left[\left\{ \Parameter j\right\} ^{\top}\mathbf{x}_{k}\right]$
and consumption 
$\mathbf{b}_{k}^{\star(j)}:=\Expectation_{\mathbf{x}_{k}\sim\Classdistribution
  j}\left[\left\{ \RParameter j\right\} ^{\top}\mathbf{x}_{k}\right]$. 
Set $\Predbound
t{\sigma}{\delta}:=\frac{16\sqrt{J\log\left(JKT\right)}}{\sqrt{t}}\!+\!\frac{6\Selfbound{\sigma}{\delta}}{\sqrt{n_{t}}}$, 
where $n_{t}$ is the number of admitted arrivals until round $t$
and $\Selfbound{\sigma}{\delta}$ is defined in Theorem \ref{thm:self}.
Suppose $t\ge8d\alpha^{-1} p_{\min}^{-1}\log JT$,
$\delta \in (0,T^{-1})$ and $F_{n_{t}}$ satisfies \eqref{eq:self_condition}.
Then with probability at least $1-4(m+1)\delta-7T^{-1}$,
\begin{equation}
\begin{split} & \sqrt{\sum_{j=1}^{J}p_{j}\max_{k\in[K]}\abs{\OptUtility
      jk-\widehat{u}_{k,t+1}^{(j)}}^{2}}\le\Predbound
  t{\sigma_{r}}{\delta},
  \\ 
&
\sqrt{\sum_{j=1}^{J}p_{j}\max_{k\in[K]}\norm{\mathbf{b}_{k}^{\star(j)}-\widehat{\mathbf{b}}_{k,t+1}^{(j)}}_{\infty}^{2}}\le\Predbound 
 t{\sigma_{b}}{\delta}. 
\end{split}
\label{eq:u_bound}
\end{equation}
\end{theorem}
The convergence rate of the estimates is $\tilde{O}(\sqrt{Jd}n_{t}^{-1/2})$.
In deriving the fast rate, the novel Gram matrix $V_{n_t}$ plays a significant role.
To prove Theorem~\ref{thm:u_convergence}, we bound the sum of squared maximum prediction error as follows:
\begin{align*}
&\frac{1}{n_{t}}\sum_{s\in\Psi_{n_{t}}}\max_{k\in[K]}\left\{ \!\left(\Parameter j\!-\!\Estimator jt\right)^{\top}\!\Context jks\!\right\} ^{2}\\
&=\frac{1}{n_{t}}\sum_{s\in\Psi_{n_{t}}}\max_{k\in[K]}\left(\!\Parameter j\!-\!\Estimator jt\!\right)\left(\Context jks\Context jks\right)^{\top}\!\left(\Parameter j\!-\!\Estimator jt\right)\\
&\le\frac{1}{n_{t}}\sum_{s\in\Psi_{n_{t}}}\left(\!\Parameter j\!-\!\Estimator jt\!\right)\left(\sum_{k=1}^{K}\Context jks\Context jks\right)^{\top}\!\left(\Parameter j\!-\!\Estimator jt\right)\\
&\le\frac{1}{n_{t}}\norm{\Parameter j\!-\!\Estimator jt}_{V_{n_{t}}}^{2}\!\!.
\end{align*}
Such a bound is not available if the Gram matrix is constructed using only contexts corresponding to selected actions. 
In this way, we obtain a faster convergence rate for the estimates for utility and consumption vectors.

\subsection{Proposed Algorithm}

Let $(K+1)$-th action denote skipping the arrival and $\Policy j{K+1}t:=\CP{\text{Skip the round }t}{\History t}$ denote the probability of skipping the arrival.
Since the decision-maker must choose an action or skip the round, we have $\sum_{k=1}^{K+1}\Policy jkt=1$. 
When the decision-maker skips round $t$, we set $\Context j{K+1}t:=0$ and $\Consumption j{K+1}t:=0$.
In round $t$, the randomized bandit policy is given by the optimal solution of the following optimization problem:
\begin{equation}
\begin{split}
\max_{\pi_{k,t}^{(\Class t)}} &
  \sum_{k=1}^{K+1}\pi_{k,t}^{(\Class t)}\left(\widehat{u}_{k,t}^{(\Class
      t)}+\frac{\Predbound{t-1}{\sigma_{r}}{\delta}}{\sqrt{p_{\Class
          t}}}\Indicator{k\in[K]}\right),\\ 
\text{s.t.} & \sum_{k=1}^{K+1}\pi_{k,t}^{(\Class
  t)}\left(\!\widehat{\mathbf{b}}_{k,t}^{(\Class
    t)}-\frac{\Predbound{t-1}{\sigma_{b}}{\delta}}{\sqrt{p_{\Class
        t}}}\mathbf{1}_{m}\!\right)\le\rho_{t}\vee0,\\ 
 & \sum_{k=1}^{K+1}\Policy{j_{t}}kt=1,\\
 & \Policy{j_{t}}kt\ge0,\quad\forall k\in[K+1],
\end{split}
\label{eq:bandit_problem}
\end{equation}
where $\rho_{t}:= \!t\rho\!-\!\sum_{s=1}^{t-1}\Consumption{\Class s}{\Action s}s$ is the difference between the used resources and planned budget until round $t$. 
The algorithm is optimistic in that it uses upper confidence bound (UCB) in rewards and lower confidence bound (LCB) in consumption while it regulates the consumption to be less than $t\rho$ with $\rho_{t}$.
In this way, the problem \eqref{eq:bandit_problem} balances between admitting the arrivals and saving the resources for later use. 
Next, we show that the optimal solution \eqref{eq:bandit_problem} is available in a closed form.

\begin{algorithm}[t]
\caption{Allocate to the Maximum First algorithm (\texttt{AMF})}
\label{alg:AORA}
\begin{algorithmic}

\STATE \textbf{INPUT}: confidence lengths $\gamma_{\theta},\gamma_{b}>0$,
confidence level $\delta\in(0,1)$. 
\STATE Initialize $F_{0}:=16d(K-1)\log\frac{Jd}{\delta}I_{J\cdot d}$,
$\rho_{1}:=\rho,$ $\widehat{\Theta}_{0}:=\mathbf{0}_{J\cdot d}$,
$\widehat{\mathbf{W}}_{0}:=\mathbf{0}_{J\cdot d\times m}$
\FOR{$t=1$ \textbf{to }$T$}
\STATE Observe arrival $(\Class t,\{\Context{\Class t}kt\}_{k\in[K]})$.
\IF{$F_{t-1}$ does not satisfy \eqref{eq:self_condition}}
\STATE Take action $\Action t=\arg\max_{k\in[K]}\rho\|\widehat{\mathbf{b}}_{k,t}^{(\Class t)}\|_{\infty}^{-1}$.
\ELSE
\STATE Compute $\widehat{u}_{k,t}^{(\Class t)}$ and $\widehat{\mathbf{b}}_{k,t}^{(\Class t)}$
with $\Estimator{\Class t}{t-1}$ and $\widehat{\mathbf{W}}_{t-1}^{(\Class t)}$.
\STATE Compute $\tilde{u}_{k,t}^{(\Class t)}:=\widehat{u}_{k,t}^{(\Class t)}+\frac{\gamma_{\theta}}{\sqrt{p_{j_t} n_{t-1}}}$
and $\tilde{\mathbf{b}}_{k,t}^{(\Class t)}:=\widehat{\mathbf{b}}_{k,t}^{(\Class t)}-\frac{\gamma_{b}}{\sqrt{p_{j_t} n_{t-1}}}\mathbf{1}_{m}$.
\STATE Take action $\Action t$ with the policy $\HatPolicy{j_{t}}1t,\ldots,\HatPolicy{j_{t}}{K+1}t$
defined~in~\eqref{eq:bandit_optimal_policy}.
\ENDIF
\IF{$\Action{t}\in[K]$}
\STATE Observe $\Reward{\Class t}{\Action t}t$ and $\mathbf{b}_{\Action t,t}^{(\Class t)}$,
then estimate $\widehat{\Theta}_{t}$ and $\widehat{\mathbf{W}}_{t}$
as in \eqref{eq:r_estimator} and \eqref{eq:w_estimator}, respectively.
\STATE Update $F_{t}=F_{t-1}+\sum_{k=1}^{K}\StackedContext kt\StackedContext kt^{\top}$.
\ENDIF
\STATE Update available resource $\rho_{t+1}=\rho_{t}+\rho-\Consumption{\Class t}{\Action t}t$.
\IF{$\sum_{s=1}^{t}\Consumption{\Class s}{\Action s}s\ge T\rho$}
\STATE Exit
\ENDIF
\ENDFOR
\end{algorithmic}
\end{algorithm}

\begin{lemma}
\label{lem:bandit_policy} (Optimal policy for bandit) Let
$\tilde{u}_{k,t}^{(\Class t)}:=\HatUtility{\Class t}kt+p_{\Class
  t}^{-1/2}\Predbound{t-1}{\sigma_{r}}{\delta}\Indicator{k\in[K]}$ 
and $\tilde{b}_{k,t}^{(\Class
  t)}(r):=\widehat{b}_{k,t}^{(\Class t)}(r)-p_{\Class
  t}^{-1/2}\Predbound{t-1}{\sigma_{b}}{\delta}$, for $r\in[m]$.  
For $i\in[K+1]$, let $\TildeUtility{\Class t}{k\Order i}t$ be a sequence of ordered variables of $\tilde{u}_{k,t}^{(\Class t)}$ in decreasing order, i.e. $\tilde{u}_{k\Order 1,t}^{(\Class t)}\ge\tilde{u}_{k\Order
  2,t}^{(\Class t)}\ge\cdots\ge\tilde{u}_{k\Order{K+1}t}^{(\Class t)}$. 
When there is a tie between $\TildeUtility{\Class t}{k\Order i}t$ and $\TildeUtility{\Class t}{k\Order{i+1}}t$, the index $k\Order i$ with the higher value for  
\[
\left(\min_{r\in[m]}\frac{\rho_{t}(r)\vee0-\sum_{h=1}^{i-1}\HatPolicy{\Class t}{k\Order h}t\tilde{b}_{k\Order h,t}^{(\Class t)}(r)}{\tilde{b}_{k\Order h,t}^{(\Class t)}(r)}\right)
\]
goes first. 
Then the policy defined as,
\begin{equation}
\begin{split}
\HatPolicy{\Class t}{k\Order 1}{t} &
  \!=\!\!\left(\min_{r\in[m]}\frac{\rho_{t}(r)\!\vee\!0}{\tilde{b}_{k\Order 1,t}^{(\Class t)}(r)}\right)\wedge1,\\ 
\HatPolicy{\Class t}{k\Order i}t &
\!=\!\!\left(\!\min_{r\in[m]}\!\frac{\rho_{t}(r)\!\vee\!0\!-\!\sum_{h=1}^{i-1}\!\HatPolicy{\Class t}{k\Order h}t\tilde{b}_{k\Order h,t}^{(\Class t)}(r)}{\tilde{b}_{k\Order i,t}^{(\Class t)}(r)}\!\right)\\ 
 & \quad\wedge\left(1-\sum_{h=i}^{i-1}\HatPolicy{\Class t}{k\Order
     h}t\right),\forall i\in[2,K+1], 
\end{split}
\label{eq:bandit_optimal_policy}
\end{equation}
is the optimal solution to~\eqref{eq:bandit_problem}.
\end{lemma}

Since the objective function of \eqref{eq:bandit_problem} is linear, we can obtain the maximum value by permuting the objective coefficients in decreasing order and allocating the greatest possible probability value in decreasing order of the objective coefficients. 
Note that $\HatPolicy{\Class t}kt$ is automatically set to zero when the utility is negative. 
This is because of the probability of skipping the arrival, $\HatPolicy{\Class t}{K+1}t = 1-\sum_{h=1}^{l-1}\HatPolicy{\Class t}{k\Order h}t$, when $\TildeUtility{\Class t}{K+1}t$ is the $l$-th largest weighted utility function and all the remaining probability is allocated to $\HatPolicy{\Class t}{K+1}t$. 
Therefor, the probabilities for actions $k$ with $\TildeUtility{\Class t}kt < \TildeUtility{\Class t}{K+1}t:=0$ are all zero.

Our proposed algorithm, Allocate to the Maximum First (\texttt{AMF}) is presented in Algorithm \ref{alg:AORA}. 
The algorithm first explores with the least consumption action until the eigenvalue condition for the estimator \eqref{eq:self_condition} holds. 
In each round of exploration, the Gram matrix of all actions is added to $F_{n_{t}}$, and any choice of action increases the eigenvalue of $F_{n_{t}}$. 
Once the condition \eqref{eq:self_condition} holds, the algorithm solves the problem \eqref{eq:bandit_problem} by computing the closed-form policy \eqref{eq:bandit_optimal_policy}.
The computational complexity of our algorithm is $\tilde{O}(d^{3}mKT+Jd^{3}T)$ where the main order occurs from updating the estimators and computing the eigenvalues of $J$ symmetric positive-definite matrix $F_{n_{t}}$.
Note that computing estimators does not depend on $J$ because the algorithm updates only $\Class t$-th variables for each $t\in[T]$.


\section{Regret Analysis}

In this section, we present our regret bound and regret analysis for the proposed \texttt{AMF} algorithm. 
\begin{theorem}
\label{thm:regret_bound} (Regret bound of \texttt{AMF}) Suppose Assumptions
\ref{assum:error}-\ref{assum:positive_definiteness} hold. 
Let $M_{\alpha,p,T}:=2304\alpha^{-2}p_{\min}^{-2}\log T+280\alpha^{-1}p_{\min}^{-1}$
and $C_{\sigma}(\delta):=96+1152\sigma\sqrt{\log\frac{4}{\delta}}$. 
Suppose $T$ and $\rho$ satisfies
$T\ge 8d\alpha^{-1}p_{\min}^{-1} \log JdT$, and $\rho\ge \sqrt{Jd/T}$. 
Setting $\gamma_\theta = 16\sqrt{J\log JKT} + 6\Selfbound{\sigma_r}{\delta}$ and $\gamma_b = 16\sqrt{J\log JKT} +6\Selfbound{\sigma_b}{\delta}$, the regret bound of \texttt{AMF} is 
\begin{align*}
\mathcal{R}_{T}^{\widehat{\pi}}\!&\le\!\!\left(\!2\!+\!\frac{OPT}{\rho T}\!\right)\!\!\Bigg\{\frac{4d\log JdT}{\alpha p_{\min}}\!+\!2dM_{\alpha,p,T}\log\frac{Jd}{\delta}\!+\!15\\
&\hspace{-0.5cm}+\!\!\left(96\!\sqrt{\log JKT}\!+\!3C_{\sigma_{r}\vee\sigma_{b}}(\delta)\!\right)\!\!\sqrt{JdT\log T}\!+\!10mT^{3}\!\delta\Bigg\}.
\end{align*}
For $\delta \in (0, m^{-1}T^{-3}$), the regret bound is
\begin{equation}
\mathcal{R}_{T}^{\widehat{\pi}}=O\left(\frac{OPT}{T\rho}\sqrt{JdT\log mJKT}\log T\right).
\label{eq:regret_bound}
\end{equation}
\end{theorem}
The regret bound~\eqref{eq:regret_bound} holds when the hyperparameter $\delta=m^{-1}T^{-3}$, which requires the knowledge of $T$.
However, in practice, selecting another value of $\delta$ does not affect the performance of the algorithm.
We provide the discussion on the sensitivity to the hyperparameter choice in Section~\ref{subsec:main_sensitivity}.

Setting $B=T\rho$, the main term of the regret bound
is $\tilde{O}(OPT/B \sqrt{JdT})$ for $B=\Omega(\sqrt{JdT})$.
The sublinear dependence of the regret bound on $J$, $d$, and $T$ is a direct consequence of the improved $\tilde{O}(\sqrt{Jd/n_{t}})$ convergence rate for the parameter estimates.
\citet{agrawal2016linear} establish a regret bound $\mathcal{R}_{T}^{\pi}=\tilde{O}(OPT/B \cdot d\sqrt{T})$ for the LinCBwK when $B=\Omega(\sqrt{d}T^{3/4})$. 
Our bound for LMMP (which subsumes LinCBwK as a special case) is improved by a $\sqrt{d}$ factor and is valid under budget constraints that relaxed from  $\Omega(\sqrt{d}T^{\frac{3}{4}})$ to $\Omega(\sqrt{d}T^{\frac{1}{2}})$.

For the proof of the regret bound, we first present the lower bound of the reward obtained by our algorithm.
\begin{lemma}
\label{lem:reward_lower_bound} Let $\tilde{u}_{k,t}^{(j)}$ and $\widehat{\mathbf{b}}_{k,t}^{(j)}$
be the estimates defined in \eqref{eq:u_b_hat}. 
Denote $\widehat{\pi}$ the policy of \texttt{AMF}.
Define the good events,
\begin{equation}
\begin{split}\mathcal{E}_{t} & :=\left\{ \tilde{u}_{k,t}^{(j)}\text{ and }\widehat{\mathbf{b}}_{k,t}^{(j)}\text{ satisfies \eqref{eq:u_bound}.}\right\} ,\\
\mathcal{M}_{t} & :=\left\{ F_{n_{t}}\text{ satisfies \eqref{eq:self_condition}.}\right\},
\end{split}
\label{eq:good_events}
\end{equation}
and $\mathcal{G}_{t}:=\mathcal{E}_t \cap \mathcal{M}_{t-1}$.
Let $\tau$ be the stopping time for the algorithm and $\xi:=\inf_{t\in[T]}\left\{ \mathcal{M}_{t-1}\cap\left\{ \rho_{t}>0\right\} \right\} $
be the starting time after the exploration for condition \eqref{eq:self_condition}.
Then, the total reward 
\begin{align*}
&\Expectation\left[\sum_{t=1}^{T}R_{t}^{\widehat{\pi}}\right]\ge\!\frac{OPT}{T}\Expectation\left[\tau-\xi\right]\!-\!\left(\!2\!+\!\frac{OPT}{\rho T}\right)\!\sum_{t=1}^{T}\Probability\!\left(\mathcal{G}_{t}^{c}\right)\\&\!\!-\!2\!\left(1\!+\!\frac{OPT}{\rho T}\right)\!\!\sqrt{T\Expectation\left[\sum_{t=1}^{T}\Predbound{t-1}{\sigma_{r}\vee\sigma_{b}}{\delta}^{2}\Indicator{\Action t\in[K]}\right]}.
\end{align*}
\end{lemma}
The lower bound consists of three main terms.
The first term $\frac{OPT}{T}\Expectation\left[\tau-\xi\right]$ relates to the period for which the algorithm uses the optimal policy \eqref{eq:bandit_optimal_policy}.
The second term $(2\!+\!\frac{OPT}{\rho
  T})\sum_{t=1}^{T}\!\Probability\left(\mathcal{G}_{t}^{c}\right)$ 
is the sum of the probability of bad events $\mathcal{M}_{t-1}^{c}$ over which the minimum eigenvalue of the Gram matrix $F_{n_{t}}$ is not large enough for the fast convergence rate, and the event $\mathcal{E}_{t}^{c}$ over which the estimator goes out of the confidence interval. And, the third term consists of the sum of confidence
lengths for the reward and consumption.

The following result bounds $\tau$, $\xi$ and the sum of bad events $\{\mathcal{M}_{t}^{c}:t\in[T]\}$. 

\begin{lemma}
\label{lem:exploration_bound} Suppose Assumptions \ref{assum:error}-\ref{assum:positive_definiteness} holds and $\rho > \sqrt{Jd/T}$.
Let $M_{\alpha,p,T}$ and $\Predbound{t}{\sigma}{\delta }$ denote the variables defined in Theorem~\ref{thm:regret_bound} and Theorem~\ref{thm:u_convergence}, respectively.
Then, for any $\delta\in(0,1/T^{2})$, the starting time $\xi:=\inf_{t\in[T]}\{\mathcal{M}_{t-1}\cap\{\rho_{t}>0\}\}$ and the stopping time $\tau$ of the \texttt{AMF} algorithm is bounded as
\begin{align*}
\Expectation\left[\xi\right]&\le\frac{1\!+\!dM_{\alpha,p,T}\log\left(\frac{Jd}{\delta}\right)\!+\!T^{2}\delta}{\rho}+1,\\
\Expectation\!\left[T\!-\!\tau\right]\!&\le\!\frac{4(m+1)T\delta+7+2\Predbound 1{\sigma_{b}}{\delta}}{\rho},
\end{align*}
and for $\mathcal{M}_{t}$ defined as in \eqref{eq:good_events}, 
\[
\sum_{t=1}^{T}\Probability\left(\mathcal{M}_{t-1}^{c}\right)\le T^{2}\delta+dM_{\alpha,p,T}\log\left(\frac{Jd}{\delta}\right).
\]
\end{lemma}

The regret bound follows from bounding the probability of $\mathcal{E}_{t}^{c}$ with Theorem~\ref{thm:u_convergence} and showing that the sum of square of $\Predbound{t}{\sigma}{\delta}$ is $O(Jd \log T)$.
The bound holds because the summation of $\Predbound{t}{\sigma}{\delta}^2=\tilde{O}(\frac{Jd}{n_t})$ over the rounds that $\Action{t} \in [K]$ happens is $\sum_{n=1}^{n_T} O(Jd/n) =O( Jd \log T)$.

\begin{figure}[t]
\vskip 0.2in
\centering
\includegraphics[width=0.48\textwidth]{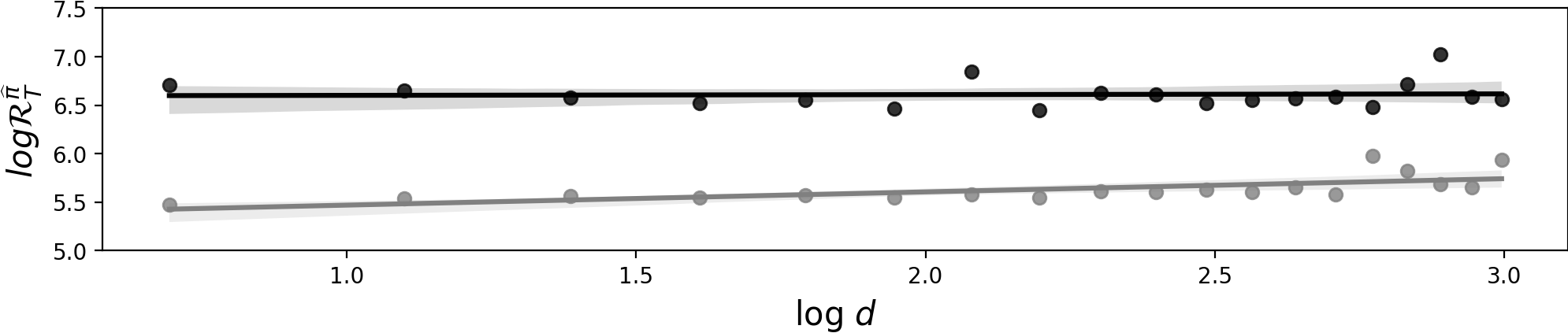}
\vskip -0.1in
\caption{\label{fig:proof} Logarithm of cumulative regret of the proposed \texttt{AMF} algorithm on various dimension $d$
when the per-period budget is $\rho=\sqrt{d/T}$. 
The gray (resp. black) line is the best fit line on the points when $T=5000$ (resp. $T=20000$).
}
\vskip -0.2in
\end{figure}

\begin{figure}[t]
\centering
\vskip 0.2in
\subfigure[Regret comparison with budget $B=\sqrt{d}T^{3/4}$]{{\label{fig:compB075}\includegraphics[width=0.235\textwidth]{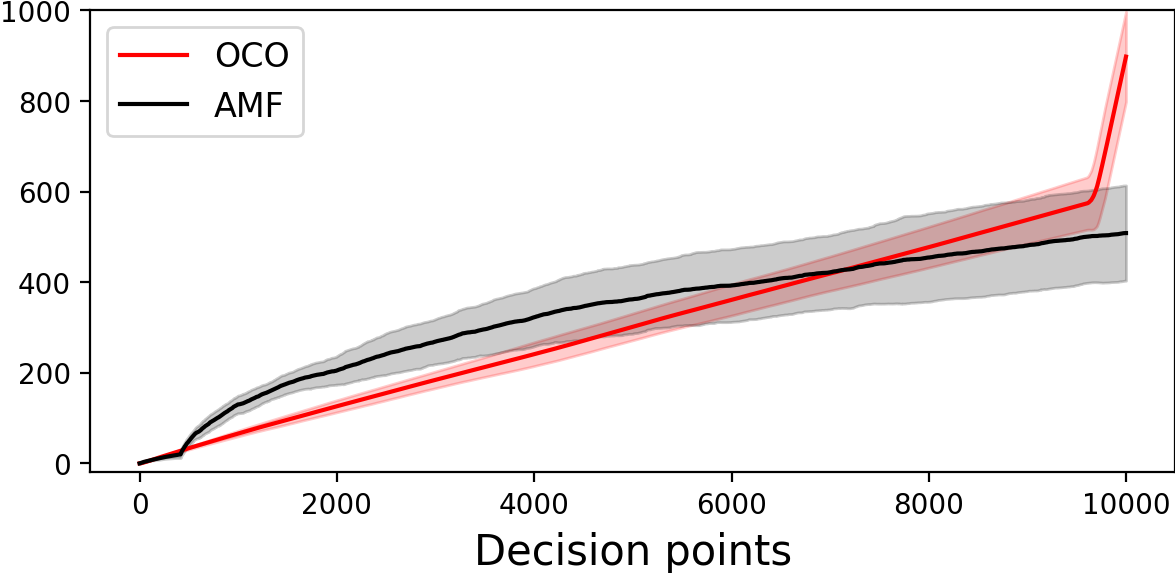}
\includegraphics[width=0.235\textwidth]{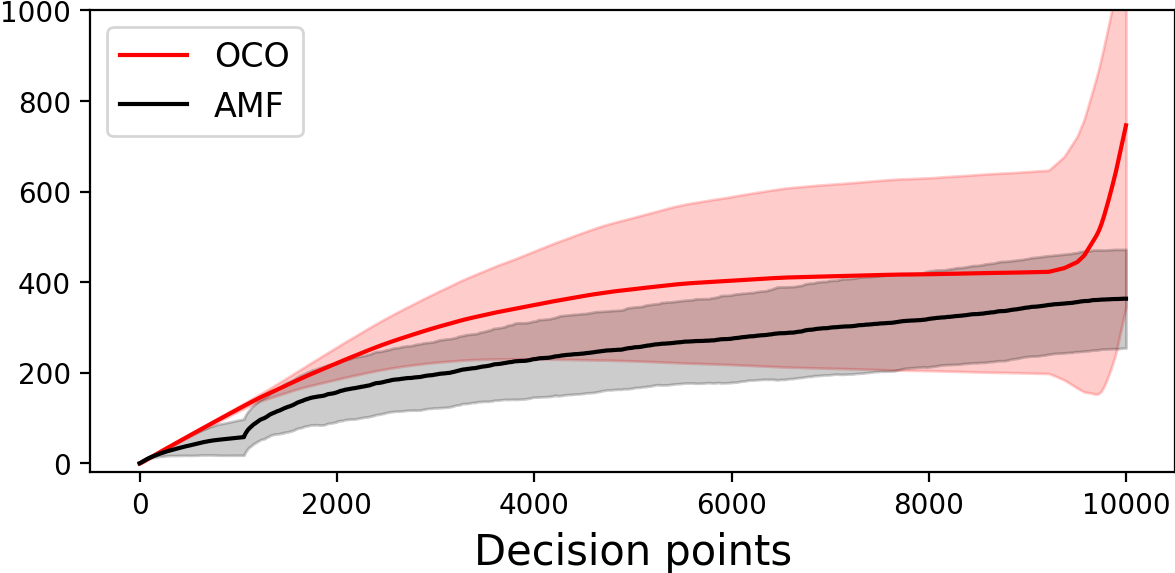}
}}
\subfigure[Regret comparison with budget $B=\sqrt{dT}$]{{\label{fig:compB05}\includegraphics[width=0.235\textwidth]{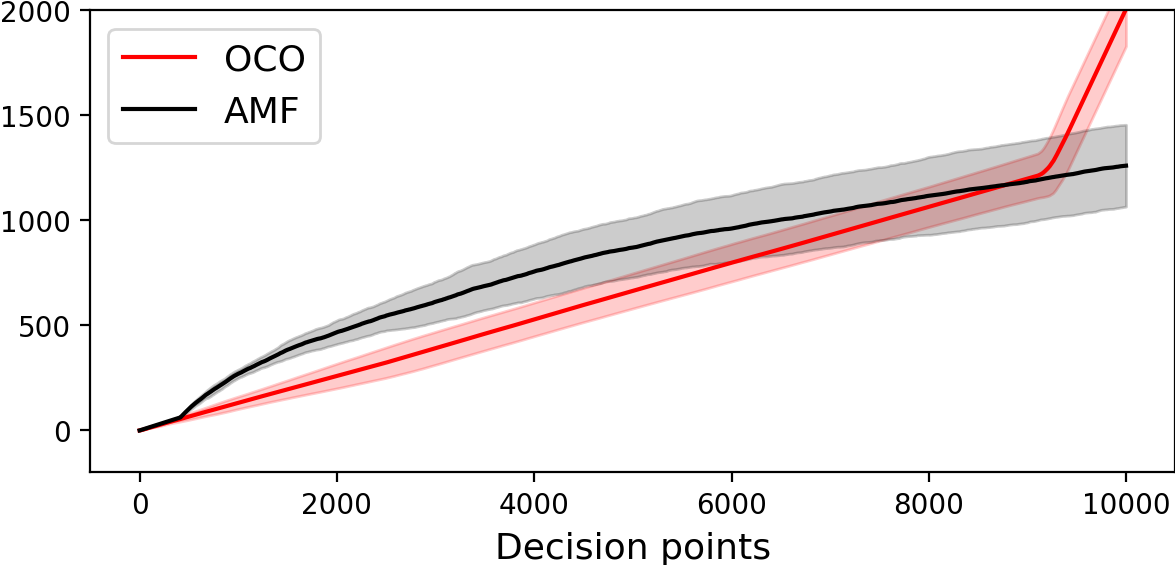}
\includegraphics[width=0.235\textwidth]{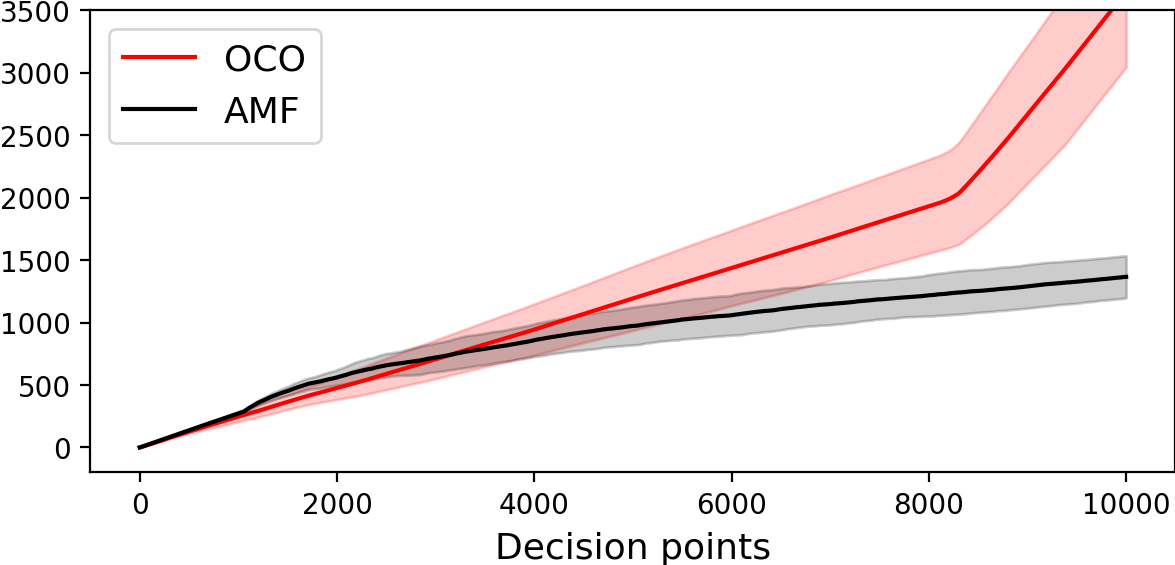}
}}
\vskip -0.1in
\caption{\label{fig:comp} Regret of \texttt{AMF} and \texttt{OCO} algorithms for $K=20$ and $m=20$. 
The line and shade represent the average and standard deviation based on 20 independent experiments. 
Additional results on different $K$ and $m$ are in Section~\ref{subsec:additional_comp}.}
\vskip -0.2in
\end{figure}

\begin{figure*}[t]
\centering
\subfigure[\label{fig:sen_gamma_theta}On various $ \gamma_\theta$]{\includegraphics[width=0.30\textwidth]{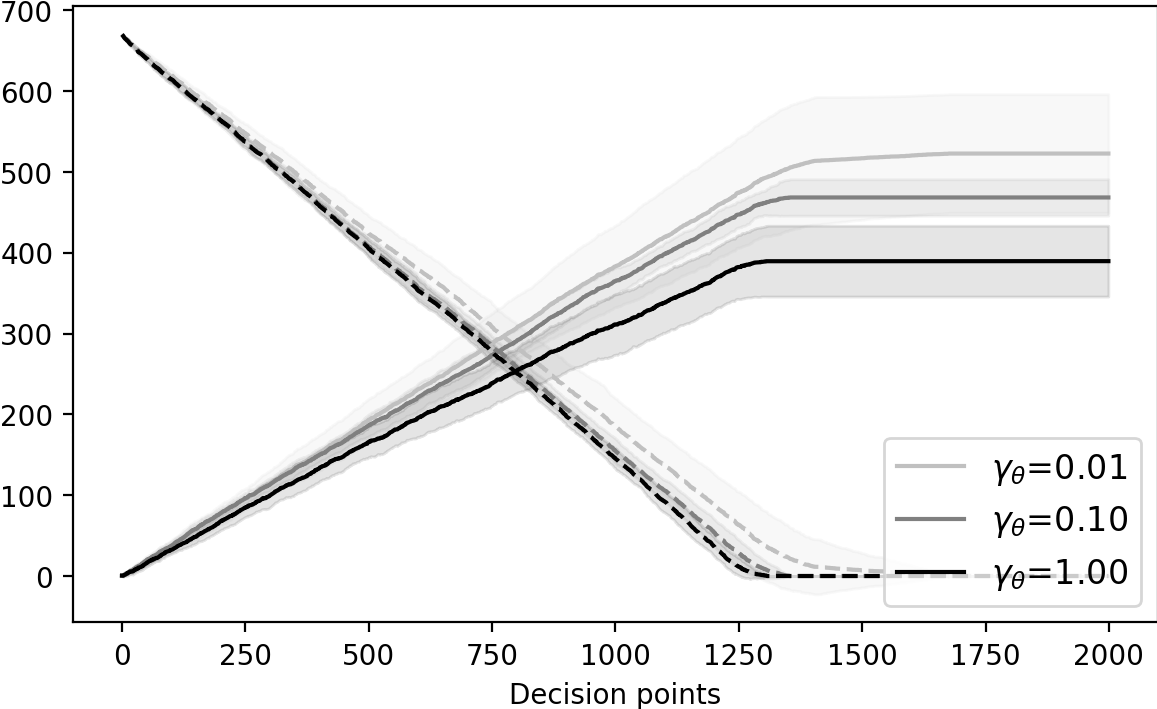}}
\subfigure[\label{fig:sen_gamma_b}On various $\gamma_b$]{\includegraphics[width=0.30\textwidth]{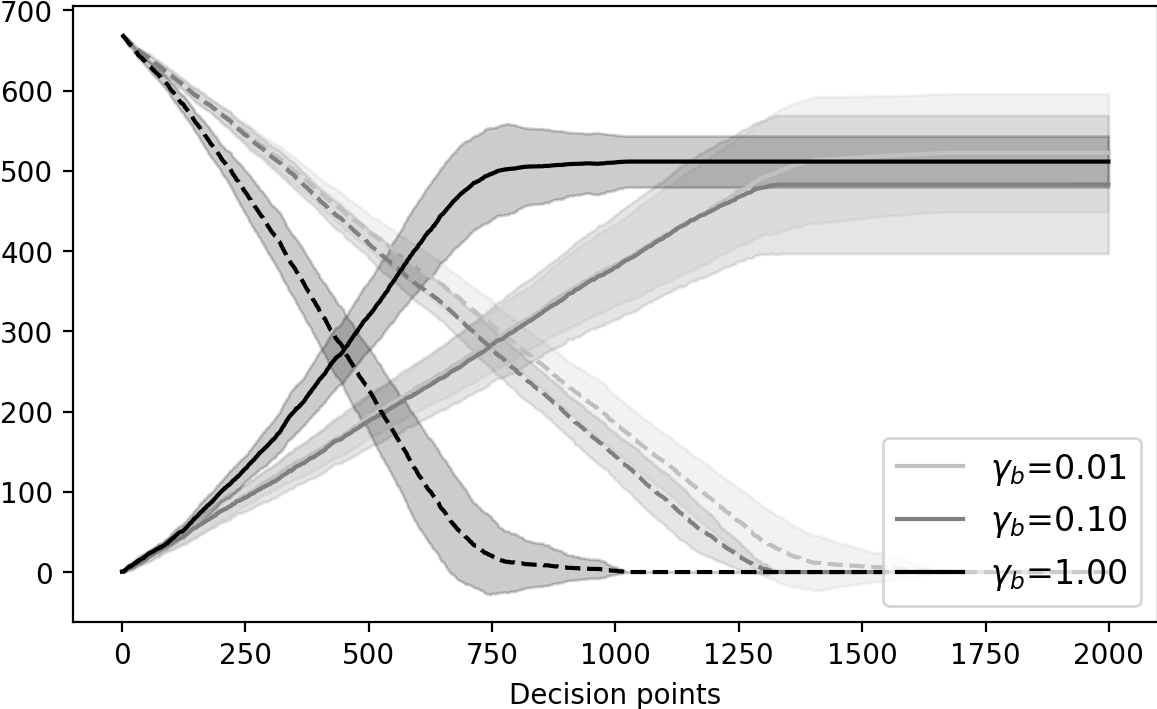}}
\subfigure[\label{fig:sen_delta}On various $\delta$]{\includegraphics[width=0.30\textwidth]{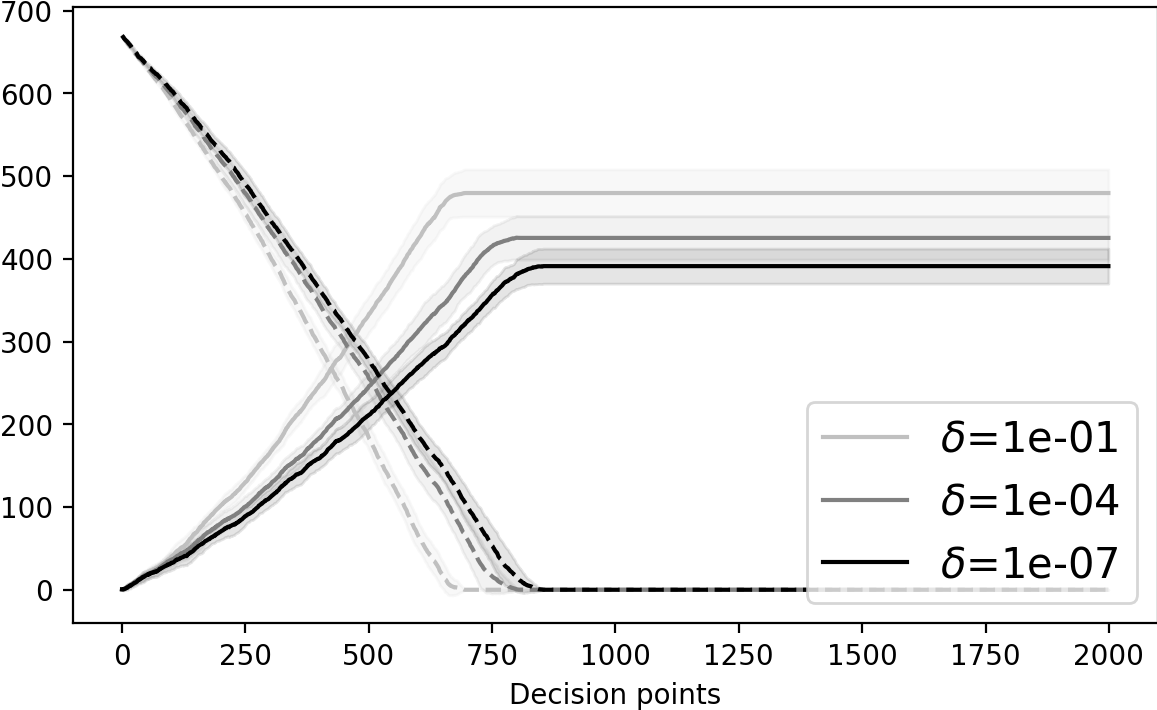}
}
\caption{\label{fig:sen} The reward and inventory of \texttt{AMF} on various hyperparameters $\gamma_{\theta}, \gamma_{b}$ and $\delta$.
The solid (resp. dashed) line represents the reward (resp. inventory).
The line and shade represent the average and standard deviation based on 10 repeated experiments, respectively.}
\end{figure*}


\section{Numerical Results}

We demonstrate the cumulative regrets with given budgets (Section~\ref{subsec:regret_d} and~\ref{subsec:regret_comp}) and the sensitivity of our proposed \texttt{AMF} to the hyperparameter choice (Section~\ref{subsec:main_sensitivity}).
For the computation of the regret, we use a setting where the instantaneous regret is computable for each round. (For the details of the setting, see Appendix~\ref{subsec:setting}.)

\subsection{Regret $\mathcal{R}_{T}^{\widehat{\pi}}$ as a function of $d$} 
\label{subsec:regret_d}
Figure~\ref{fig:proof} plots $\log(\mathcal{R}_{T}^{\widehat{\pi}})$ vs.~$\log(d)$ for a single-class ($J = 1$) LMMP for $T = \{5000, 20000\}$ and the budget $B=\sqrt{dT}$, where our $\tilde{O}(\frac{OPT}{B}\sqrt{JdT})$ regret bound implies that $\log (\mathcal{R}_{T}^{\widehat{\pi}})$ is constant over $d$.
The regression line on the plot is nearly flat and the slope of the best-fit line is $0.136$ (resp. $0.008$) for $T=5000$ (resp. $T=20000$). 
The weak increase in $T=5000$ is captured by the $O(d \log JdmT)$ term in our bound, which diminishes for large $T$.



\subsection{Comparison of \texttt{AMF} with \texttt{OCO}}
\label{subsec:regret_comp}
In order to compare \texttt{AMF} with \texttt{OCO}~\citep{agrawal2016linear}, we set $J=1$.
The hyperparameters for \texttt{AMF} were set to $\gamma_{\theta}=1$,
$\gamma_b=1$ and $\delta = 0.01$. 

Figure~\ref{fig:comp}(a) (resp. (b)) plots the cumulative regret of the two algorithms with budget $B=\sqrt{d}T^{\frac{3}{4}}$ (resp. $B=\sqrt{dT}$).
Note that \texttt{OCO} requires a minimum budget $B = \sqrt{d}T^{\frac{3}{4}}$ whereas \texttt{AMF} requires a lower minimum budget of $B = \sqrt{dT}$.
The regret lines cross because AMF is allowed to skip arrivals whereas \texttt{OCO} does not skip arrivals. 
The sudden bend points at the end of the round in \texttt{OCO} show that it runs out of budget and has regret = 1.
In all cases, our algorithm performs better and the performance gap
increases as $d$ increases. 
Note that the regret plot for \texttt{OCO} never flattens out for most cases, where the regret of \texttt{AMF} flattens as $t$ increases. 
This is because our new estimator which uses contexts from \emph{all}
actions with unbiased pseudo-rewards~\eqref{eq:pseudo_rewards} for unselected actions and has a significantly faster convergence rate as compared with the estimator used in \texttt{OCO}.

\subsection{Sensitivity Analysis}
\label{subsec:main_sensitivity}
We demonstrate the sensitivity of the proposed \texttt{AMF} algorithm to its three hyperparameters: $\gamma_\theta$, $\gamma_b$, and $\delta$. 
Figure~\ref{fig:sen_gamma_theta} and~\ref{fig:sen_gamma_b} show the reward and inventory of our algorithm on various $\gamma_\theta\in\{0.01, 0.1, 1\}$ and $\gamma_b \in \{0.01, 0.1,1\}$.
We present on these sets since the variability of the reward and the inventory of the algorithm are hardly visible outside the sets.
Figure~\ref{fig:sen_delta} shows the reward and inventory of \texttt{AMF} on various $\delta \in \{10^{-1},10^{-4},10^{-7}\}$.
When $\delta \ge 10^{-1}$ (resp. $\delta \le 10^{-7}$) the reward and inventories are same with $\delta = 10^{-1}$ (resp. $\delta = 10^{-7}$). 
The change in the reward and the consumption of the proposed \texttt{AMF} is visible only when the hyperparameters change drastically.
This shows that choice of hyperparameters is not sensitive.
The effect of $\gamma_\theta$ and $\gamma_b$ diminishes fast by $n_t^{-1/2}$ term and our policy finds the order of the utilities rather than their absolute values.
For $\delta$, which controls the sampling probabilities~\eqref{eq:pseudo_action_probability} in estimators and the exploration rounds in~\eqref{eq:self_condition}, it also has a small effect.
This is because the minimum eigenvalue of $F_{n_t}$ increases in $\Omega(n_t)$-rate and reduces the effect of $\log\frac{1}{\delta}$ terms in~\eqref{eq:pseudo_action_probability} and~\eqref{eq:self_condition}.
Therefore, our algorithm guarantees a similar performance for other hyperparameters than specified in Theorem~\ref{thm:regret_bound}.
For details of the experimental settings and recommendation of the specific hyperparameter choices, see Appendix~\ref{subsec:sensitivity}.

\section{Conclusion}

We introduce a new problem class  LMMP  that extends upon LinCBwK and online revenue management to a multi-class setting.  
To address this problem, we propose a novel estimator that utilizes unbiased pseudo-rewards and contexts of \textit{all} actions to learn class-specific parameters of all classes.  
We use this transfer-learning-based estimator to propose an algorithm in which the policy is available in closed form and the worst-case regret is $\tilde{O}(\frac{OPT}{B}\sqrt{JdT})$ when the budget $B(T)=\Omega(\sqrt{JdT})$.  
This result improves both the regret bound and the minimum budget required, resolving an open problem in LinCBwK. 
Numerical experiments demonstrate superior performances over benchmarks for the single-class case, and robustness to hyperparameters changes for multiple-class data. 

\section*{Acknowledgements}
We thank the anonymous referees for offering many helpful comments and valuable feedback.
Wonyoung Kim was supported by the National Research Foundation of Korea (NRF) grant funded by the Korea government (MSIT) (No. RS-2023-00240142) and Garud Iyengar was supported by NSF EFMA-2132142, ARPA-E PERFORM Program, and ONR N000142312374.

\bibliographystyle{icml2023}
\bibliography{ref}

\begin{thebibliography}{39}
\providecommand{\natexlab}[1]{#1}
\providecommand{\url}[1]{\texttt{#1}}
\expandafter\ifx\csname urlstyle\endcsname\relax
  \providecommand{\doi}[1]{doi: #1}\else
  \providecommand{\doi}{doi: \begingroup \urlstyle{rm}\Url}\fi

\bibitem[Abbasi-Yadkori et~al.(2011)Abbasi-Yadkori, P{\'a}l, and
  Szepesv{\'a}ri]{abbasi2011improved}
Abbasi-Yadkori, Y., P{\'a}l, D., and Szepesv{\'a}ri, C.
\newblock Improved algorithms for linear stochastic bandits.
\newblock In \emph{Advances in Neural Information Processing Systems}, pp.\
  2312--2320, 2011.

\bibitem[Agrawal \& Devanur(2016)Agrawal and Devanur]{agrawal2016linear}
Agrawal, S. and Devanur, N.
\newblock Linear contextual bandits with knapsacks.
\newblock \emph{Advances in Neural Information Processing Systems}, 29, 2016.

\bibitem[Agrawal \& Devanur(2014{\natexlab{a}})Agrawal and
  Devanur]{agrawal2014bandits}
Agrawal, S. and Devanur, N.~R.
\newblock Bandits with concave rewards and convex knapsacks.
\newblock In \emph{Proceedings of the fifteenth ACM conference on Economics and
  computation}, pp.\  989--1006, 2014{\natexlab{a}}.

\bibitem[Agrawal \& Devanur(2014{\natexlab{b}})Agrawal and
  Devanur]{agrawal2014fast}
Agrawal, S. and Devanur, N.~R.
\newblock Fast algorithms for online stochastic convex programming.
\newblock In \emph{Proceedings of the twenty-sixth annual ACM-SIAM symposium on
  Discrete algorithms}, pp.\  1405--1424. SIAM, 2014{\natexlab{b}}.

\bibitem[Amani et~al.(2019)Amani, Alizadeh, and Thrampoulidis]{amani2019linear}
Amani, S., Alizadeh, M., and Thrampoulidis, C.
\newblock Linear stochastic bandits under safety constraints.
\newblock In \emph{Advances in Neural Information Processing Systems}, pp.\
  9252--9262, 2019.

\bibitem[Azuma(1967)]{azuma1967weighted}
Azuma, K.
\newblock Weighted sums of certain dependent random variables.
\newblock \emph{Tohoku Mathematical Journal, Second Series}, 19\penalty0
  (3):\penalty0 357--367, 1967.

\bibitem[Badanidiyuru et~al.(2018)Badanidiyuru, Kleinberg, and
  Slivkins]{badanidiyuru2018bandits}
Badanidiyuru, A., Kleinberg, R., and Slivkins, A.
\newblock Bandits with knapsacks.
\newblock \emph{Journal of the ACM (JACM)}, 65\penalty0 (3):\penalty0 1--55,
  2018.

\bibitem[Bang \& Robins(2005)Bang and Robins]{bang2005doubly}
Bang, H. and Robins, J.~M.
\newblock Doubly robust estimation in missing data and causal inference models.
\newblock \emph{Biometrics}, 61\penalty0 (4):\penalty0 962--973, 2005.

\bibitem[Bastani \& Bayati(2020)Bastani and Bayati]{bastani2020online}
Bastani, H. and Bayati, M.
\newblock Online decision making with high-dimensional covariates.
\newblock \emph{Operations Research}, 68\penalty0 (1):\penalty0 276--294, 2020.

\bibitem[Bastani et~al.(2021)Bastani, Bayati, and Khosravi]{bastani2021mostly}
Bastani, H., Bayati, M., and Khosravi, K.
\newblock Mostly exploration-free algorithms for contextual bandits.
\newblock \emph{Management Science}, 67\penalty0 (3):\penalty0 1329--1349,
  2021.

\bibitem[Besbes \& Zeevi(2009)Besbes and Zeevi]{besbes2009dynamic}
Besbes, O. and Zeevi, A.
\newblock Dynamic pricing without knowing the demand function: Risk bounds and
  near-optimal algorithms.
\newblock \emph{Operations Research}, 57\penalty0 (6):\penalty0 1407--1420,
  2009.

\bibitem[Besbes \& Zeevi(2012)Besbes and Zeevi]{besbes2012blind}
Besbes, O. and Zeevi, A.
\newblock Blind network revenue management.
\newblock \emph{Operations research}, 60\penalty0 (6):\penalty0 1537--1550,
  2012.

\bibitem[Boyd et~al.(2004)Boyd, Boyd, and Vandenberghe]{boyd2004convex}
Boyd, S., Boyd, S.~P., and Vandenberghe, L.
\newblock \emph{Convex optimization}.
\newblock Cambridge university press, 2004.

\bibitem[Chu et~al.(2011)Chu, Li, Reyzin, and Schapire]{chu2011contextual}
Chu, W., Li, L., Reyzin, L., and Schapire, R.
\newblock Contextual bandits with linear payoff functions.
\newblock In \emph{Proceedings of the Fourteenth International Conference on
  Artificial Intelligence and Statistics}, pp.\  208--214, 2011.

\bibitem[Devanur et~al.(2011)Devanur, Jain, Sivan, and
  Wilkens]{devanur2011near}
Devanur, N.~R., Jain, K., Sivan, B., and Wilkens, C.~A.
\newblock Near optimal online algorithms and fast approximation algorithms for
  resource allocation problems.
\newblock In \emph{Proceedings of the 12th ACM conference on Electronic
  commerce}, pp.\  29--38, 2011.

\bibitem[Dimakopoulou et~al.(2019)Dimakopoulou, Zhou, Athey, and
  Imbens]{dimakopoulou2019balanced}
Dimakopoulou, M., Zhou, Z., Athey, S., and Imbens, G.
\newblock Balanced linear contextual bandits.
\newblock In \emph{Proceedings of the AAAI Conference on Artificial
  Intelligence}, volume~33, pp.\  3445--3453, 2019.

\bibitem[Efron \& Tibshirani(1994)Efron and Tibshirani]{efron1994introduction}
Efron, B. and Tibshirani, R.~J.
\newblock \emph{An introduction to the bootstrap}.
\newblock CRC press, 1994.

\bibitem[Feldman et~al.(2010)Feldman, Henzinger, Korula, Mirrokni, and
  Stein]{feldman2010online}
Feldman, J., Henzinger, M., Korula, N., Mirrokni, V.~S., and Stein, C.
\newblock Online stochastic packing applied to display ad allocation.
\newblock In \emph{European Symposium on Algorithms}, pp.\  182--194. Springer,
  2010.

\bibitem[Ferreira et~al.(2018)Ferreira, Simchi-Levi, and
  Wang]{ferreira2018online}
Ferreira, K.~J., Simchi-Levi, D., and Wang, H.
\newblock Online network revenue management using thompson sampling.
\newblock \emph{Operations research}, 66\penalty0 (6):\penalty0 1586--1602,
  2018.

\bibitem[Gallego \& Van~Ryzin(1994)Gallego and Van~Ryzin]{gallego1994optimal}
Gallego, G. and Van~Ryzin, G.
\newblock Optimal dynamic pricing of inventories with stochastic demand over
  finite horizons.
\newblock \emph{Management science}, 40\penalty0 (8):\penalty0 999--1020, 1994.

\bibitem[Good(2006)]{good2006resampling}
Good, P.~I.
\newblock \emph{Resampling methods}.
\newblock Springer, 2006.

\bibitem[Immorlica et~al.(2019)Immorlica, Sankararaman, Schapire, and
  Slivkins]{immorlica2019adversarial}
Immorlica, N., Sankararaman, K.~A., Schapire, R., and Slivkins, A.
\newblock Adversarial bandits with knapsacks.
\newblock In \emph{2019 IEEE 60th Annual Symposium on Foundations of Computer
  Science (FOCS)}, pp.\  202--219. IEEE, 2019.

\bibitem[Kannan et~al.(2018)Kannan, Morgenstern, Roth, Waggoner, and
  Wu]{kannan2018smoothed}
Kannan, S., Morgenstern, J.~H., Roth, A., Waggoner, B., and Wu, Z.~S.
\newblock A smoothed analysis of the greedy algorithm for the linear contextual
  bandit problem.
\newblock \emph{Advances in neural information processing systems}, 31, 2018.

\bibitem[Kim \& Paik(2019)Kim and Paik]{kim2019doubly}
Kim, G. and Paik, M.~C.
\newblock Doubly-robust lasso bandit.
\newblock In \emph{Advances in Neural Information Processing Systems}, pp.\
  5869--5879, 2019.

\bibitem[Kim et~al.(2021)Kim, Kim, and Paik]{kim2021doubly}
Kim, W., Kim, G.-S., and Paik, M.~C.
\newblock Doubly robust thompson sampling with linear payoffs.
\newblock In Beygelzimer, A., Dauphin, Y., Liang, P., and Vaughan, J.~W.
  (eds.), \emph{Advances in Neural Information Processing Systems}, 2021.

\bibitem[Kim et~al.(2022)Kim, Lee, and Paik]{kim2022double}
Kim, W., Lee, K., and Paik, M.~C.
\newblock Double doubly robust thompson sampling for generalized linear
  contextual bandits.
\newblock \emph{arXiv preprint arXiv:2209.06983}, 2022.

\bibitem[Kim et~al.(2023)Kim, Paik, and Oh]{kim2022squeeze}
Kim, W., Paik, M.~C., and Oh, M.-H.
\newblock Squeeze all: Novel estimator and self-normalized bound for linear
  contextual bandits.
\newblock In Ruiz, F., Dy, J., and van~de Meent, J.-W. (eds.),
  \emph{Proceedings of The 26th International Conference on Artificial
  Intelligence and Statistics}, volume 206 of \emph{Proceedings of Machine
  Learning Research}, pp.\  3098--3124. PMLR, 25--27 Apr 2023.
\newblock URL \url{https://proceedings.mlr.press/v206/kim23d.html}.

\bibitem[Lattimore \& Szepesv{\'a}ri(2020)Lattimore and
  Szepesv{\'a}ri]{lattimore2020bandit}
Lattimore, T. and Szepesv{\'a}ri, C.
\newblock \emph{Bandit algorithms}.
\newblock Cambridge University Press, 2020.

\bibitem[Lee et~al.(2016)Lee, Peres, and Smart]{lee2016}
Lee, J.~R., Peres, Y., and Smart, C.~K.
\newblock A gaussian upper bound for martingale small-ball probabilities.
\newblock \emph{Ann. Probab.}, 44\penalty0 (6):\penalty0 4184--4197, 11 2016.
\newblock \doi{10.1214/15-AOP1073}.

\bibitem[Li et~al.(2021)Li, Sun, and Ye]{li21symmetry}
Li, X., Sun, C., and Ye, Y.
\newblock The symmetry between arms and knapsacks: A primal-dual approach for
  bandits with knapsacks.
\newblock In Meila, M. and Zhang, T. (eds.), \emph{Proceedings of the 38th
  International Conference on Machine Learning}, volume 139 of
  \emph{Proceedings of Machine Learning Research}, pp.\  6483--6492. PMLR,
  18--24 Jul 2021.

\bibitem[Liu et~al.(2021)Liu, Li, Shi, and Ying]{liu2021efficient}
Liu, X., Li, B., Shi, P., and Ying, L.
\newblock An efficient pessimistic-optimistic algorithm for stochastic linear
  bandits with general constraints.
\newblock \emph{Advances in Neural Information Processing Systems},
  34:\penalty0 24075--24086, 2021.

\bibitem[Moradipari et~al.(2021)Moradipari, Amani, Alizadeh, and
  Thrampoulidis]{moradipari21safethompson}
Moradipari, A., Amani, S., Alizadeh, M., and Thrampoulidis, C.
\newblock Safe linear thompson sampling with side information.
\newblock \emph{IEEE Transactions on Signal Processing}, 69:\penalty0
  3755--3767, 2021.
\newblock \doi{10.1109/TSP.2021.3089822}.

\bibitem[Oh et~al.(2021)Oh, Iyengar, and Zeevi]{oh2021sparsity}
Oh, M.-h., Iyengar, G., and Zeevi, A.
\newblock Sparsity-agnostic lasso bandit.
\newblock In \emph{International Conference on Machine Learning}, pp.\
  8271--8280. PMLR, 2021.

\bibitem[Pacchiano et~al.(2021)Pacchiano, Ghavamzadeh, Bartlett, and
  Jiang]{pacchiano21linear}
Pacchiano, A., Ghavamzadeh, M., Bartlett, P., and Jiang, H.
\newblock Stochastic bandits with linear constraints.
\newblock In Banerjee, A. and Fukumizu, K. (eds.), \emph{Proceedings of The
  24th International Conference on Artificial Intelligence and Statistics},
  volume 130 of \emph{Proceedings of Machine Learning Research}, pp.\
  2827--2835. PMLR, 13--15 Apr 2021.

\bibitem[Sankararaman \& Slivkins(2021)Sankararaman and
  Slivkins]{sankararaman2021bandits}
Sankararaman, K.~A. and Slivkins, A.
\newblock Bandits with knapsacks beyond the worst case.
\newblock \emph{Advances in Neural Information Processing Systems},
  34:\penalty0 23191--23204, 2021.

\bibitem[Sivakumar et~al.(2020)Sivakumar, Wu, and
  Banerjee]{sivakumar2020structured}
Sivakumar, V., Wu, S., and Banerjee, A.
\newblock Structured linear contextual bandits: A sharp and geometric smoothed
  analysis.
\newblock In \emph{International Conference on Machine Learning}, pp.\
  9026--9035. PMLR, 2020.

\bibitem[Sivakumar et~al.(2022)Sivakumar, Zuo, and
  Banerjee]{sivakumar2022smoothed}
Sivakumar, V., Zuo, S., and Banerjee, A.
\newblock Smoothed adversarial linear contextual bandits with knapsacks.
\newblock In \emph{International Conference on Machine Learning}, pp.\
  20253--20277. PMLR, 2022.

\bibitem[Tropp(2012)]{tropp2012user}
Tropp, J.~A.
\newblock User-friendly tail bounds for sums of random matrices.
\newblock \emph{Foundations of computational mathematics}, 12\penalty0
  (4):\penalty0 389--434, 2012.

\bibitem[Tropp(2015)]{tropp2015introduction}
Tropp, J.~A.
\newblock An introduction to matrix concentration inequalities.
\newblock \emph{Foundations and Trends{\textregistered} in Machine Learning},
  8\penalty0 (1-2):\penalty0 1--230, 2015.

\end{thebibliography}

\clearpage{}

\appendix
\onecolumn


\section{Supplementary for Experiments}

\subsection{Settings of Parameters and Contexts for Regret Computation \label{subsec:setting}}
For numerical experiments, we devise a setting where explicit regret computation is available.
We set $J=1$ for \texttt{OCO} to be compatible with the setting.
For $x\in\Real_{+}$, let $\lceil x \rceil$ be the smallest integer greater than equal to $x$.
For parameters, we set $\theta_{\star}=(-1,\cdots,-1,\lceil d/2\rceil^{-1},\cdots,\lceil d/2\rceil^{-1})$ and 
\[
W_{\star}=\begin{pmatrix}\rho\lceil d/2\rceil^{-1} & \cdots & \rho\lceil d/2\rceil^{-1}\\
\vdots & \cdots & \vdots\\
\rho\lceil d/2\rceil^{-1} & \cdots & \rho\lceil d/2\rceil^{-1}\\
\rho & \cdots & \rho\\
\vdots & \vdots & \vdots\\
\rho & \cdots & \rho
\end{pmatrix},
\]
where the $\lceil d/2\rceil^{-1}$ and $\rho\lceil d/2\rceil^{-1}$ terms are in the first $\lceil d/2\rceil$ entries.
For contexts, we set the optimal action by $(0,\cdots,0,1,\cdots,1)$, and for other actions, we set  $(U_{0,0.05},\cdots,U_{0,0.05},U_{-0.05,0},\cdots,U_{-0.05,0})$,where $U_{a,b}$ is the Uniform random variable supported on $[a,b]$.
Then we have the optimal arm with reward 1 and consumption $\rho$, while other arms have reward less than 1 and consumption more than $\rho$.

\subsection{Experiment Settings for Sensitivity Analysis \label{subsec:sensitivity}}

The settings of the experiment in Section~\ref{subsec:main_sensitivity} is described as follows.
The number of classes is $J=3$ with a uniform prior $p=(1/3,1/3,1/3)^{\top}$ and every $d=5$ elements of $K=10$ contexts are generated from the uniform distribution on $[\frac{kj}{KJ}-1,\frac{kj}{KJ}+1]$ for $k\in[K]$ and $j\in[J]$. 
The costs are generated from the uniform distribution on $[\frac{k(J-j+1)-1}{KJ},\frac{k(J-j+1)+1}{KJ}]$ for $k\in[K]$ and $j\in[J]$.
Each element of $\Parameter{j}$ and $\RParameter{j}$ is generated from $U_{0,1}$ and fixed throughout the experiment.
The generated rewards and consumption vectors are not truncated to one to impose greater variability, as our algorithm does not show apparent sensitivity on bounded rewards and consumption vectors.
The algorithm consumes the budget faster than in previous experiments because the consumption vector is not bounded to 1.

Based on the experiments, we recommend using grid search on $\gamma_{\theta}\times\gamma_{b} \in [0,1]^2$ to maximize the reward.

However, we recommend using $\delta=0.1$, which is greater than the specified value in Theorem \ref{thm:regret_bound} for the algorithm to start using its policy in earlier rounds.

\begin{figure*}[ht]
\centering
\subfigure[Regret comparison under $K=10$ and $m=10$]{{\includegraphics[width=0.24\textwidth]{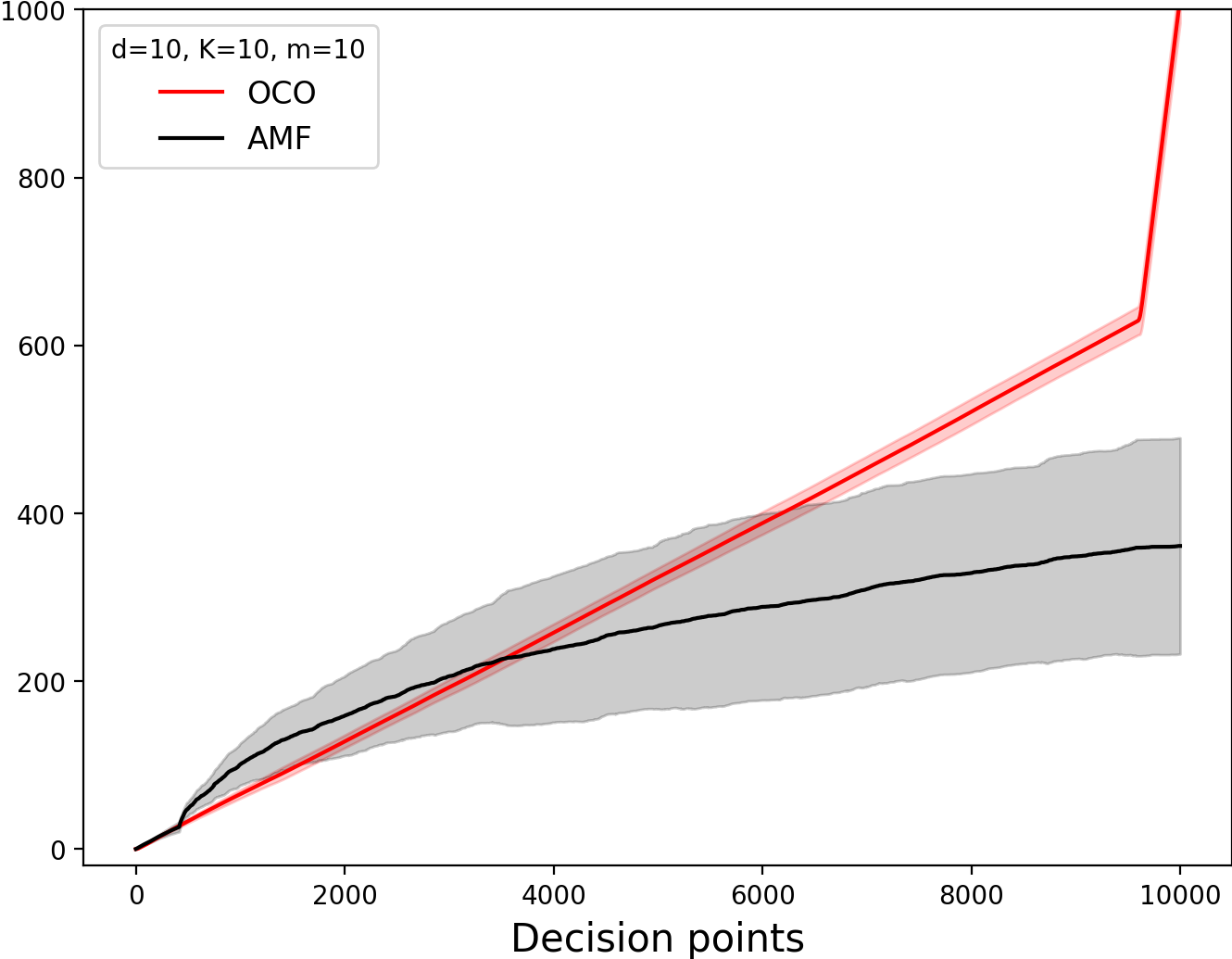}
\includegraphics[width=0.24\textwidth]{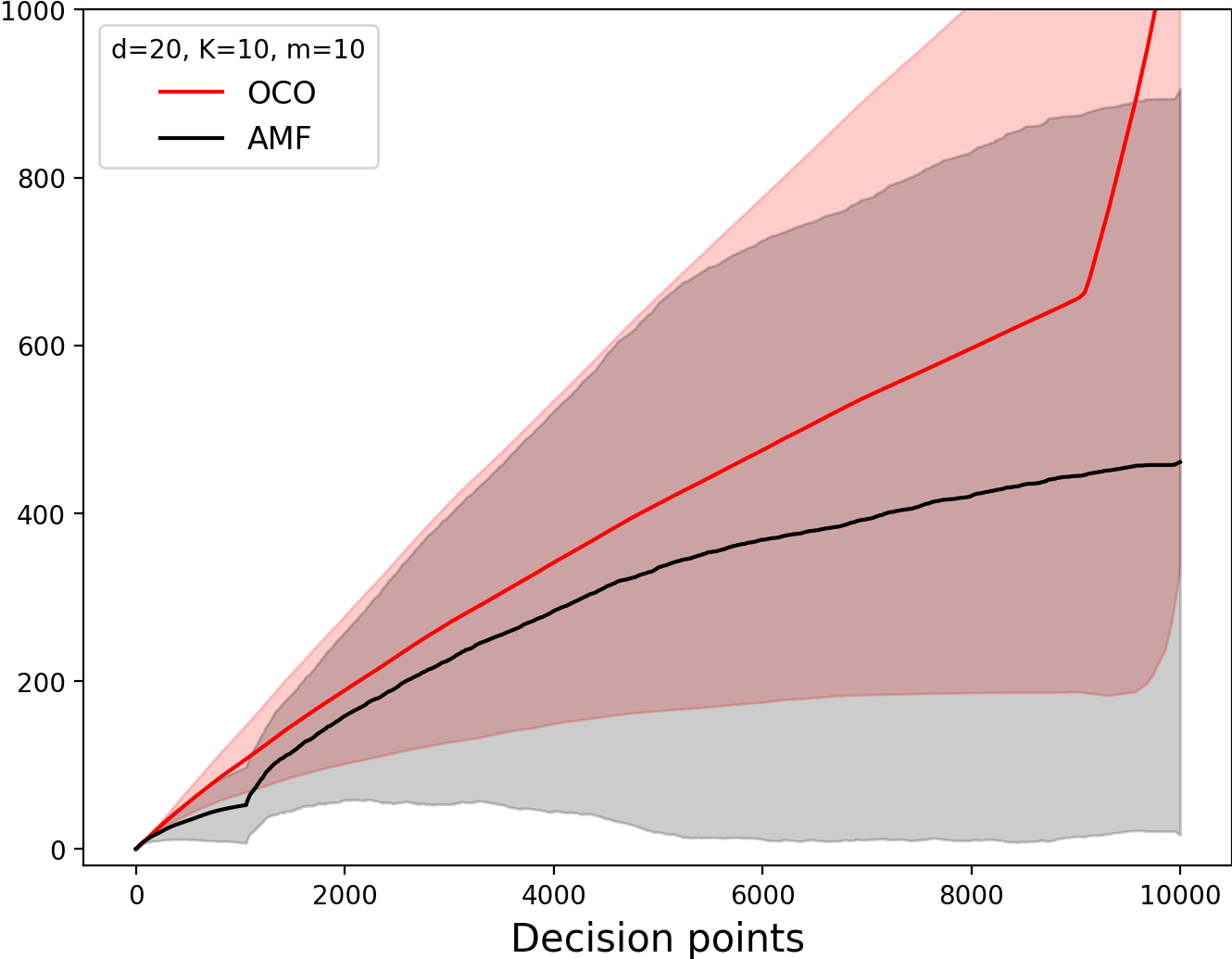}
}}
\subfigure[Regret comparison under $K=20$ and $m=10$]{{\includegraphics[width=0.24\textwidth]{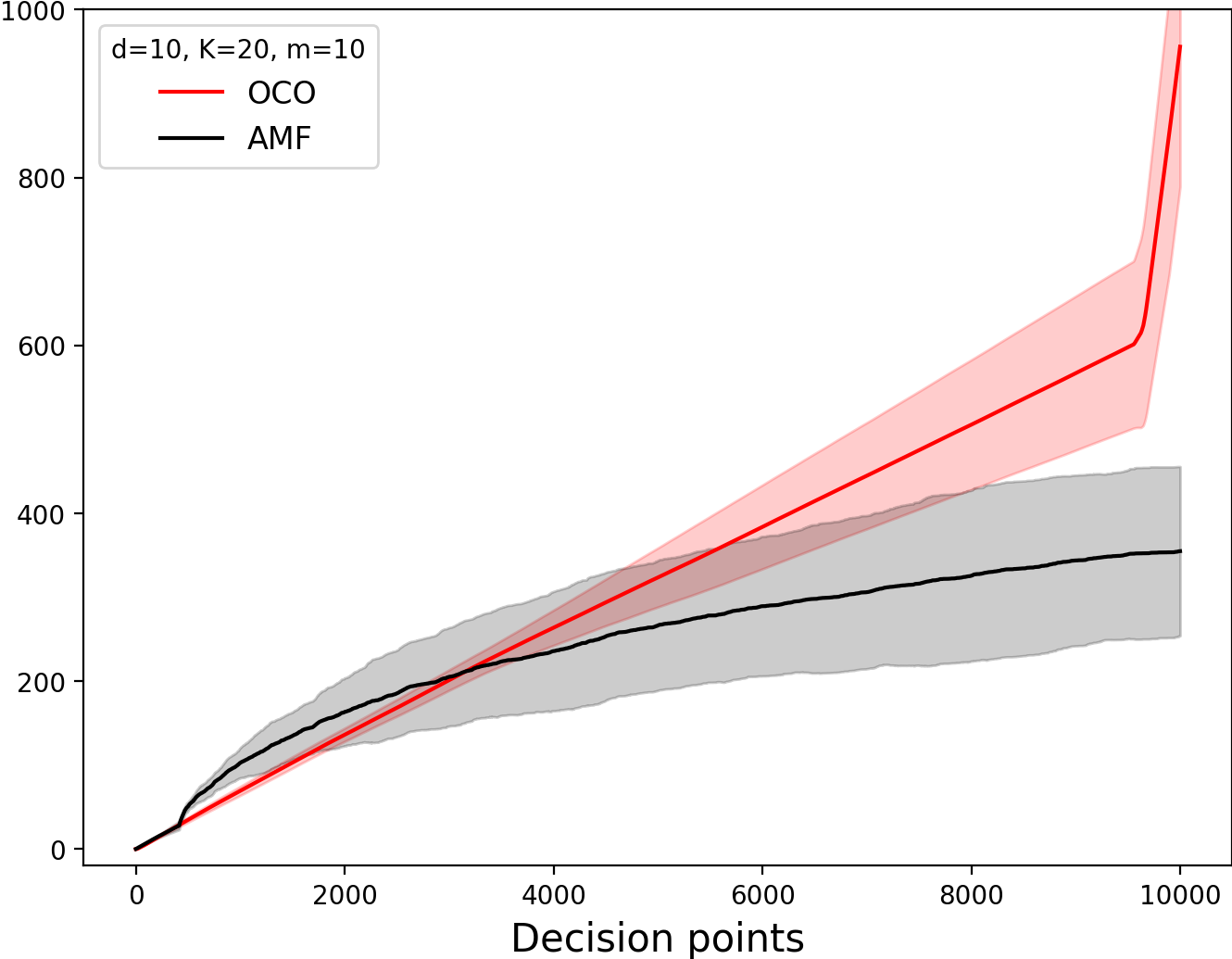}
\includegraphics[width=0.24\textwidth]{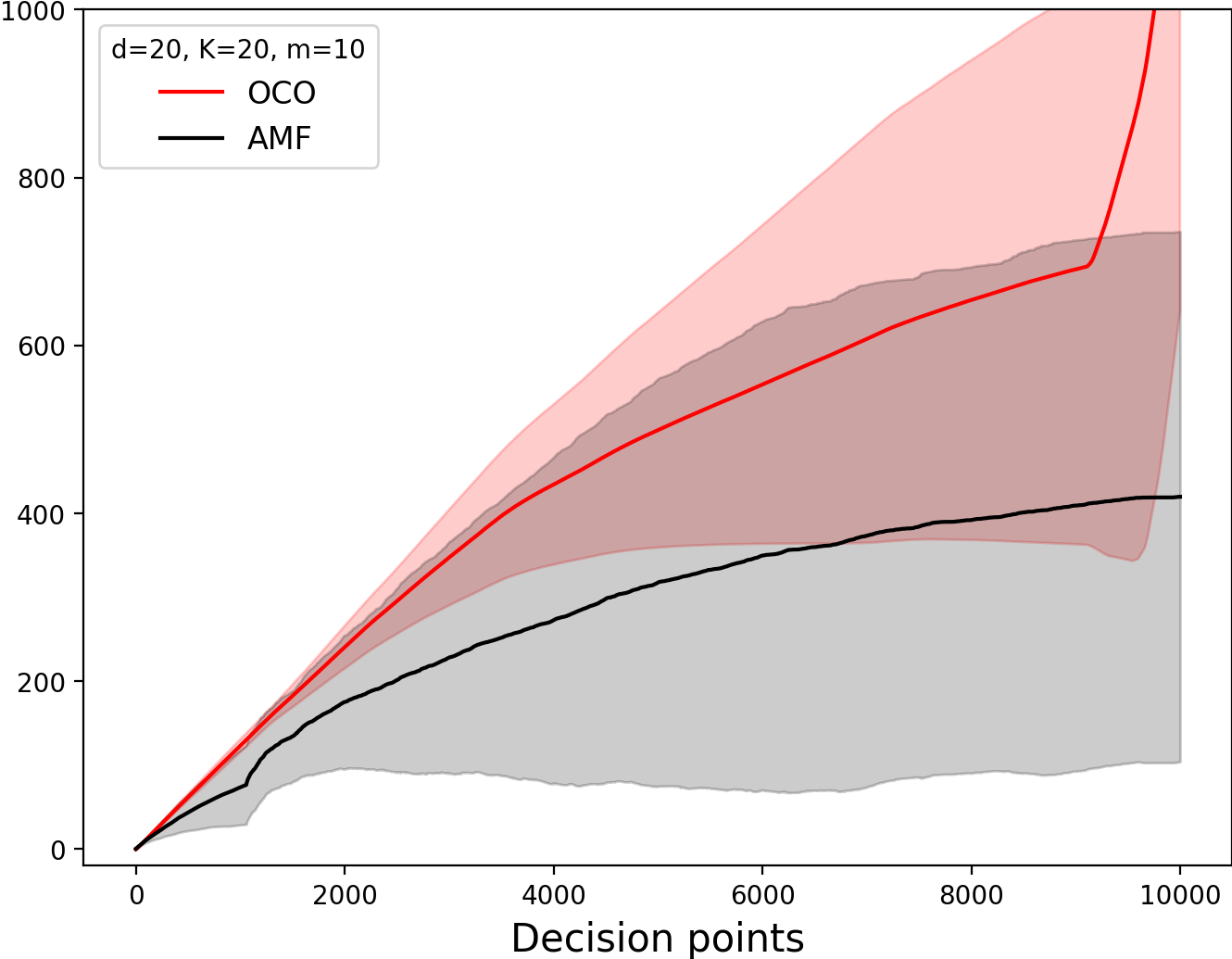}
}}
\\
\subfigure[Regret comparison under $K=10$ and $m=20$]{{\includegraphics[width=0.24\textwidth]{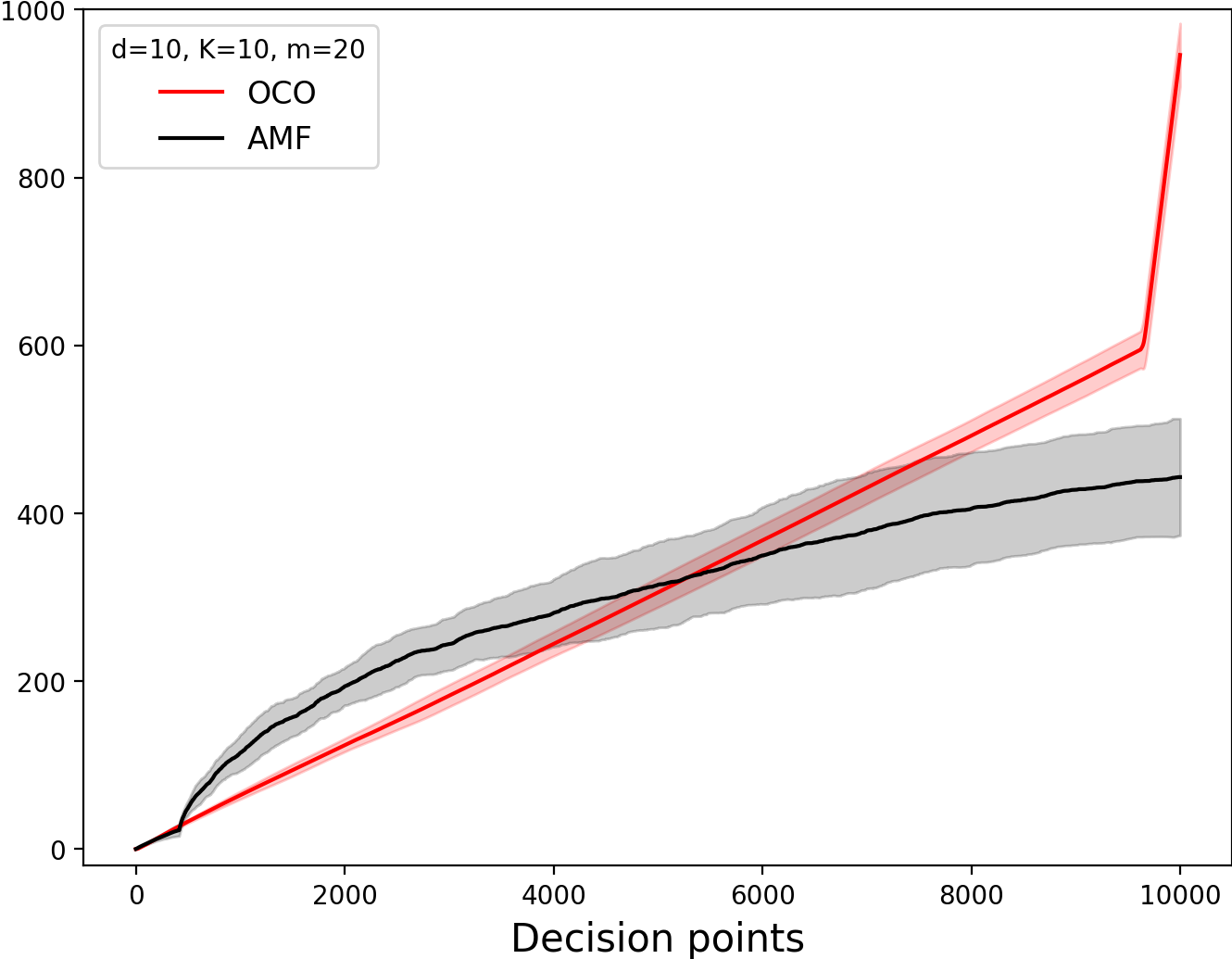}
\includegraphics[width=0.24\textwidth]{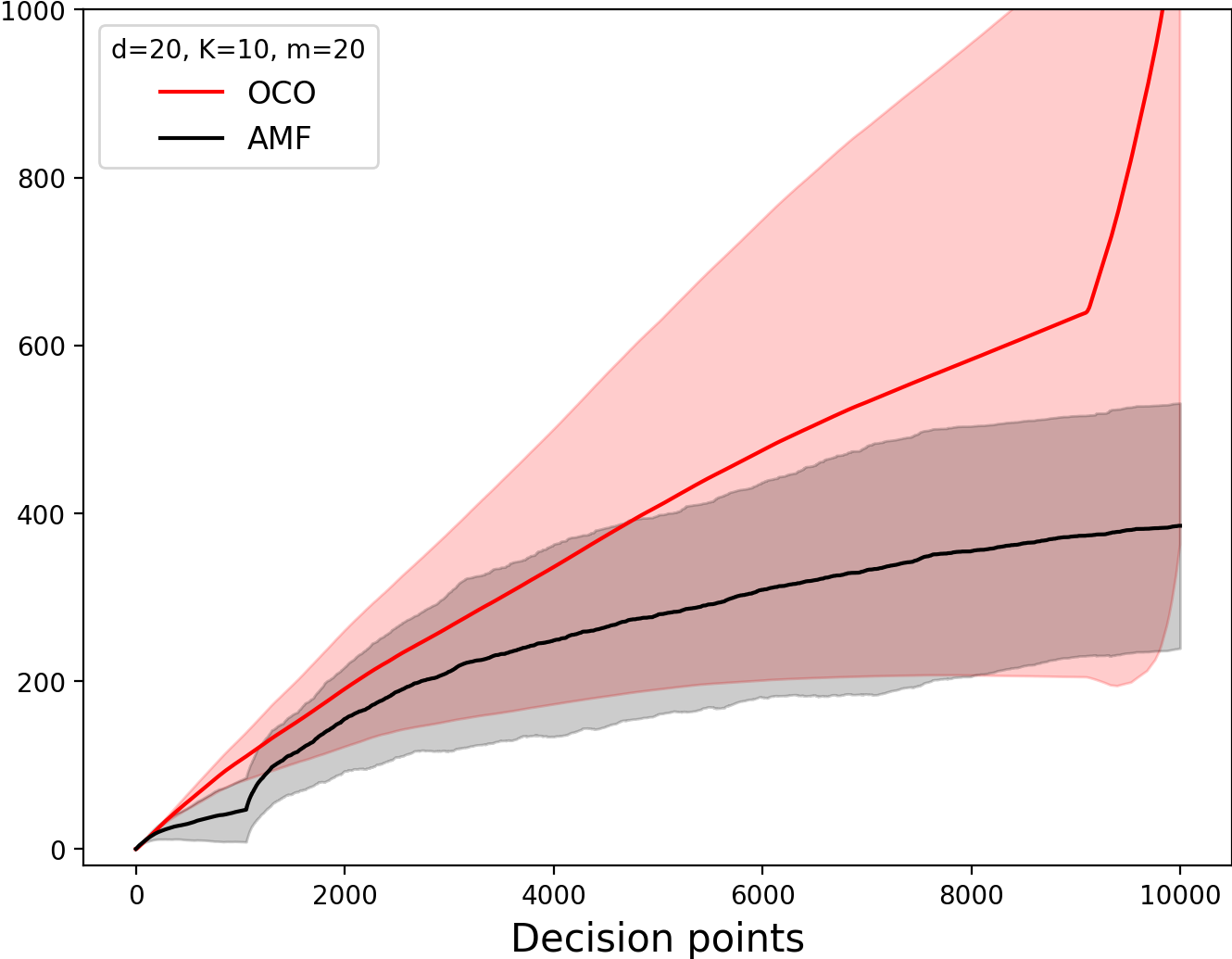}
}}
\subfigure[Regret comparison under $K=20$ and $m=20$]{{\includegraphics[width=0.24\textwidth]{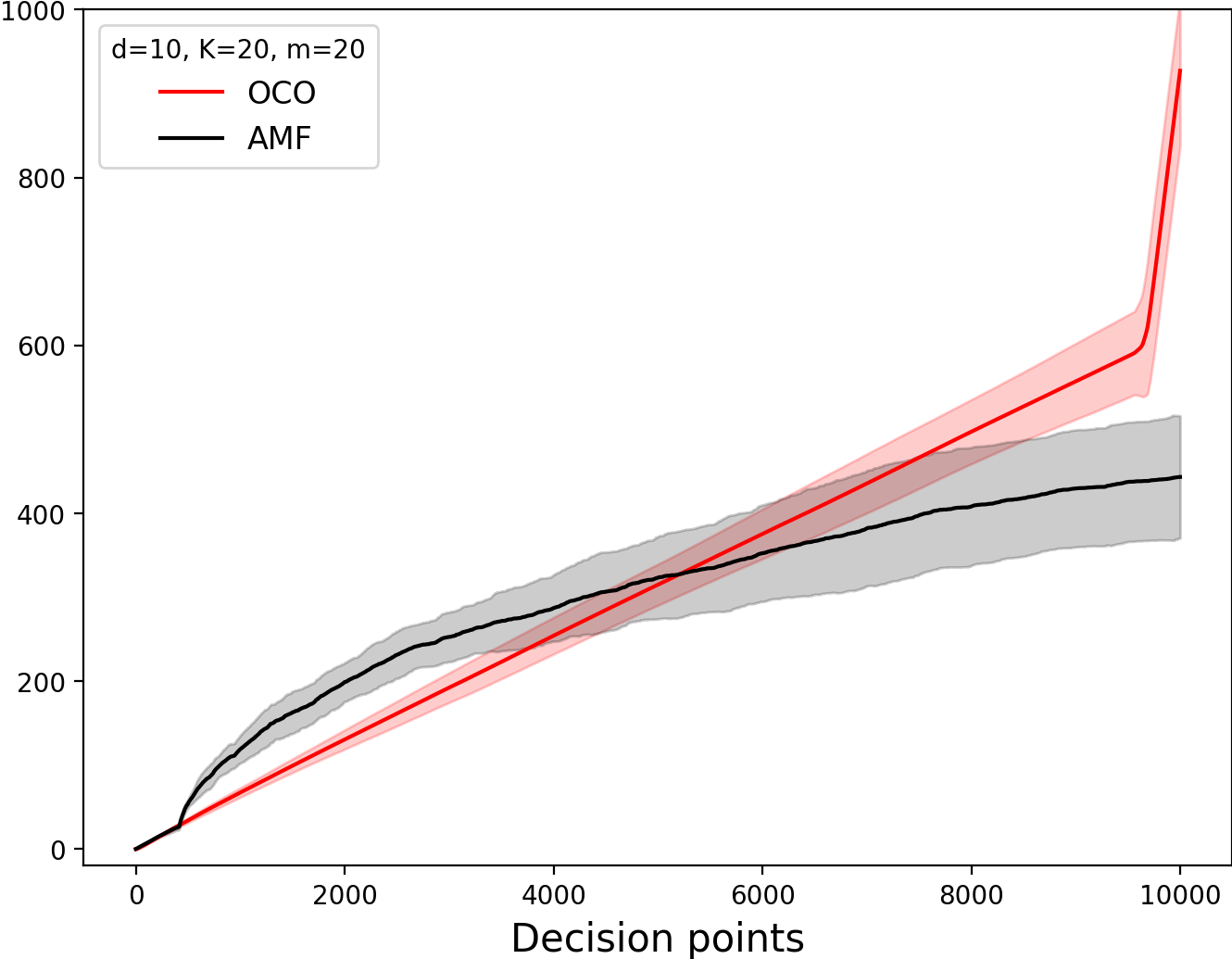}
\includegraphics[width=0.24\textwidth]{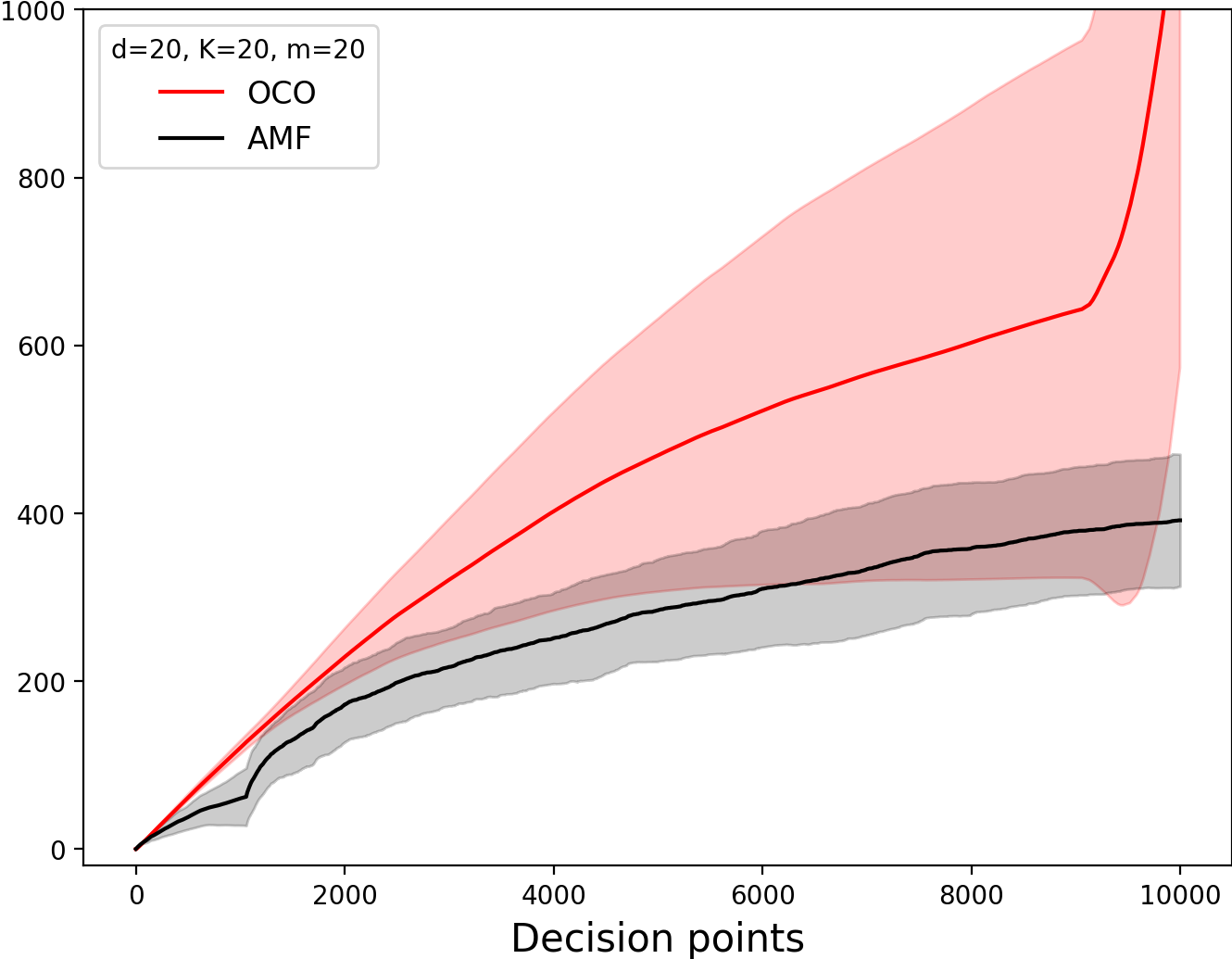}
}}
\caption{\label{fig:additional_comp_B075}Regret comparison of \texttt{AMF} and \texttt{OCO} algorithms under $B=dT^{3/4}$. 
The line and shade represent the average and standard deviation based on 20 repeated experiments.}
\end{figure*}

\begin{figure*}[ht]
\centering
\subfigure[Regret comparison under $K=10$ and $m=10$]{{\includegraphics[width=0.24\textwidth]{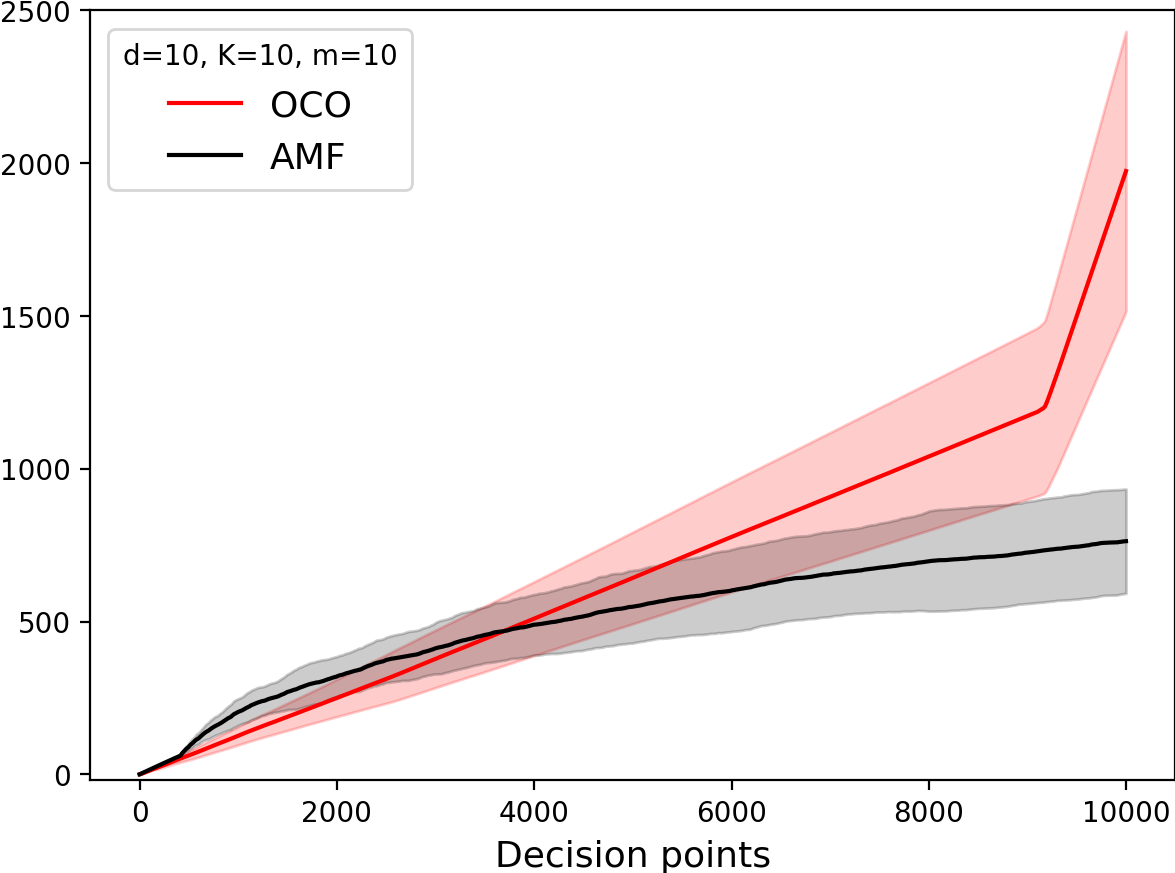}
\includegraphics[width=0.24\textwidth]{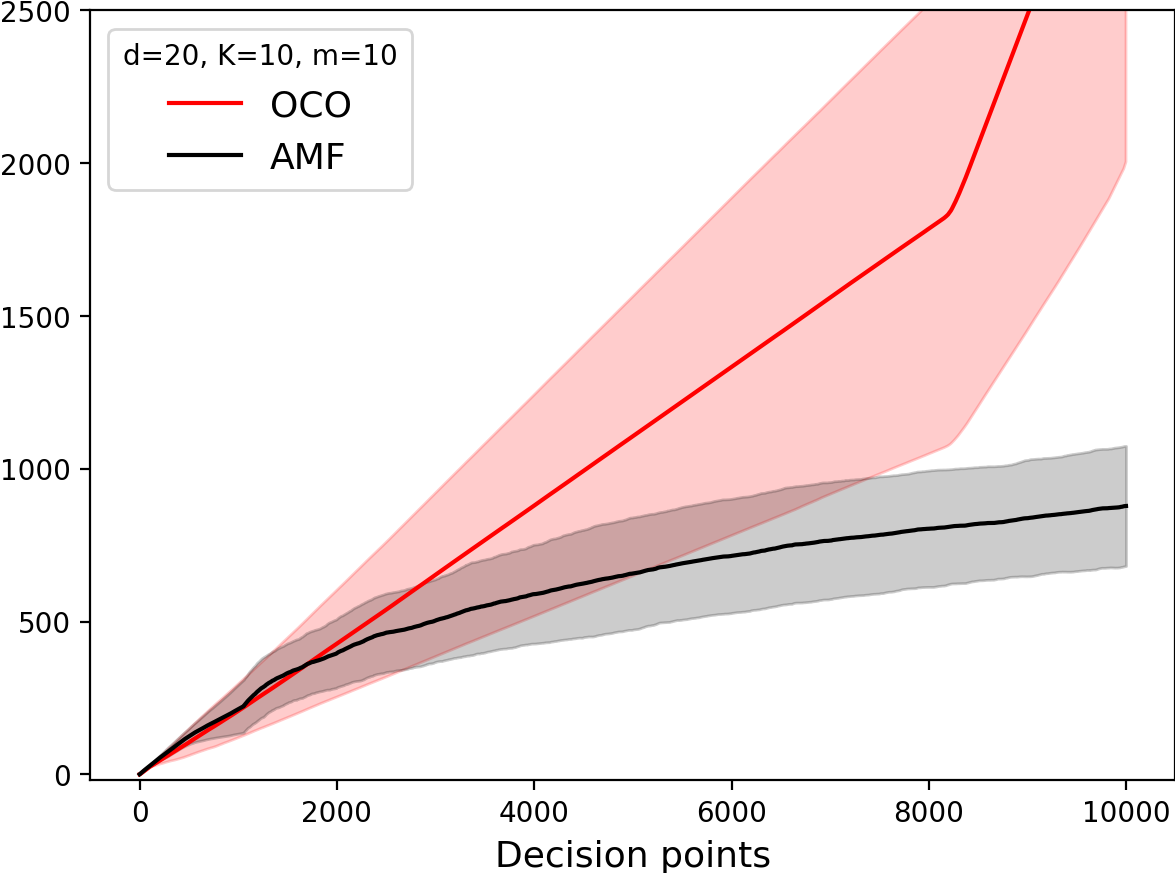}
}}
\subfigure[Regret comparison under $K=20$ and $m=10$]{{\includegraphics[width=0.24\textwidth]{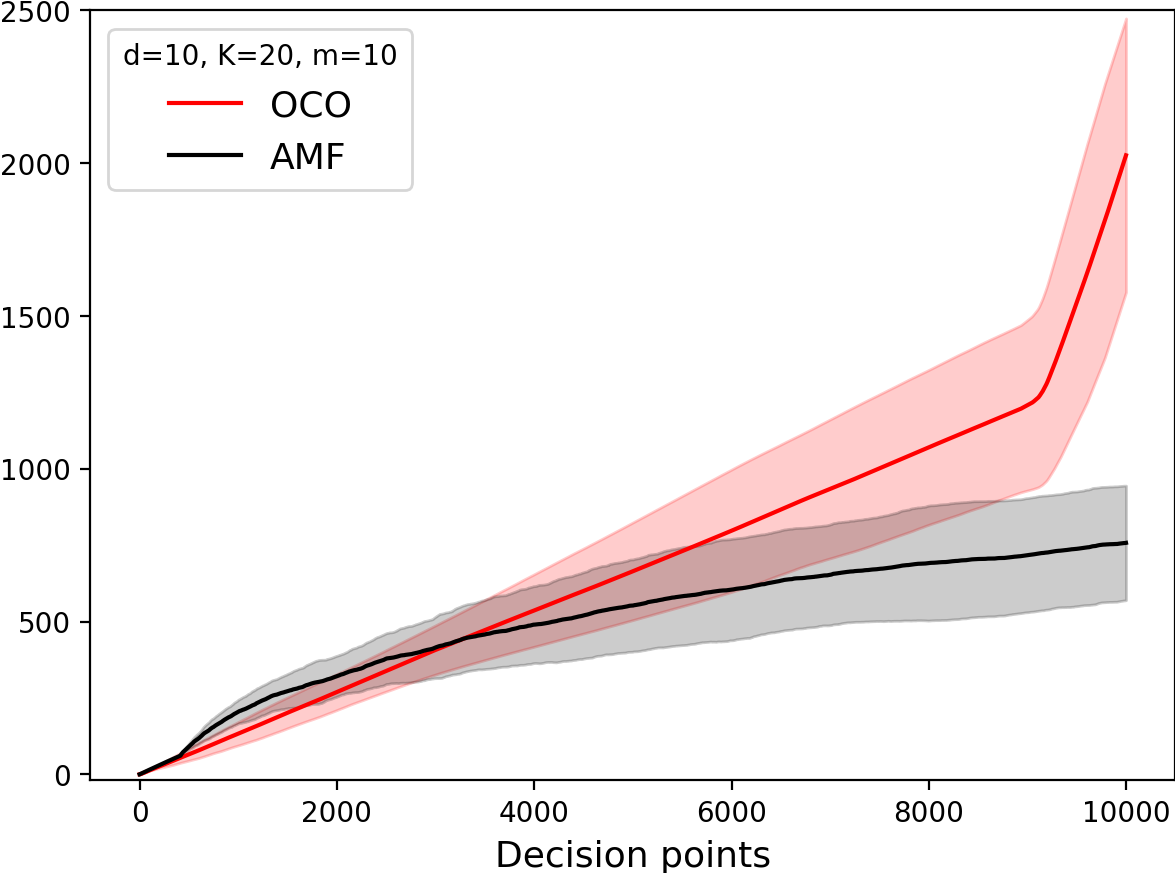}
\includegraphics[width=0.24\textwidth]{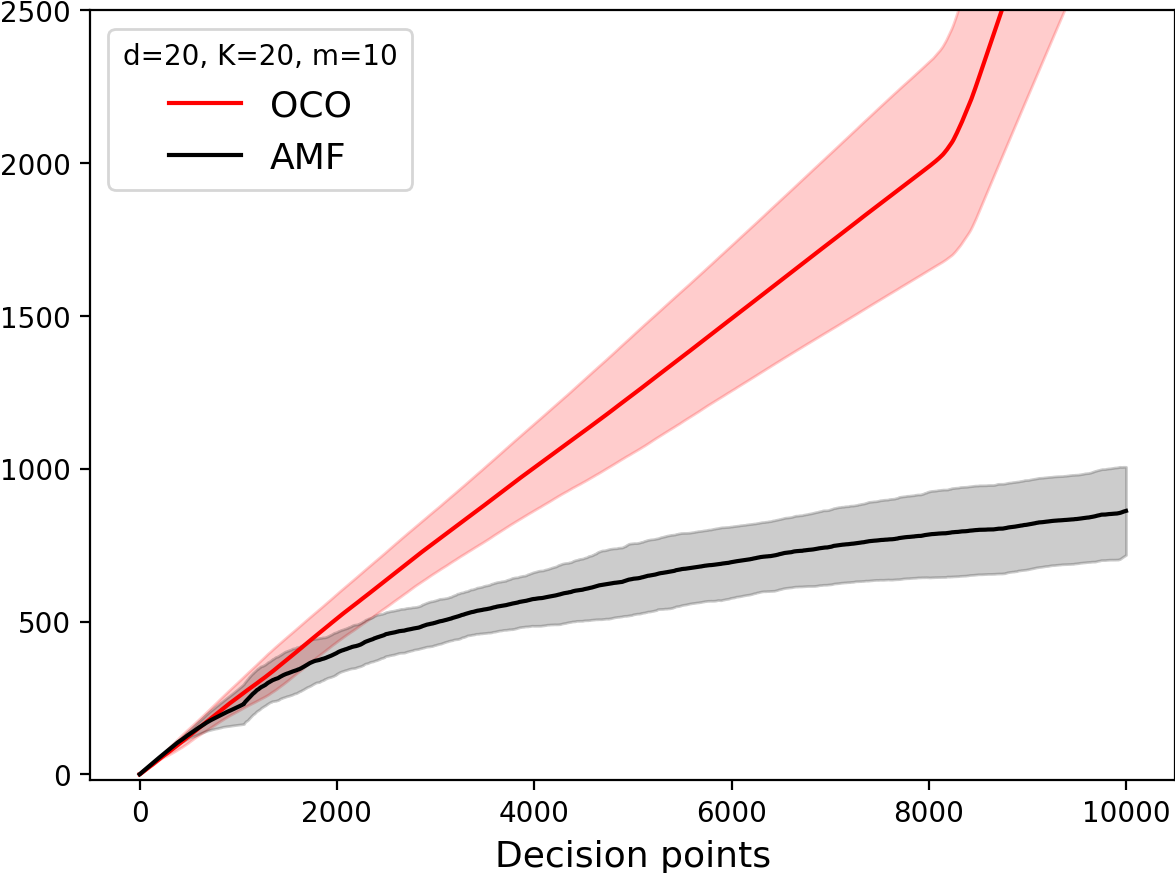}
}}
\\
\subfigure[Regret comparison under $K=10$ and $m=20$]{{\includegraphics[width=0.24\textwidth]{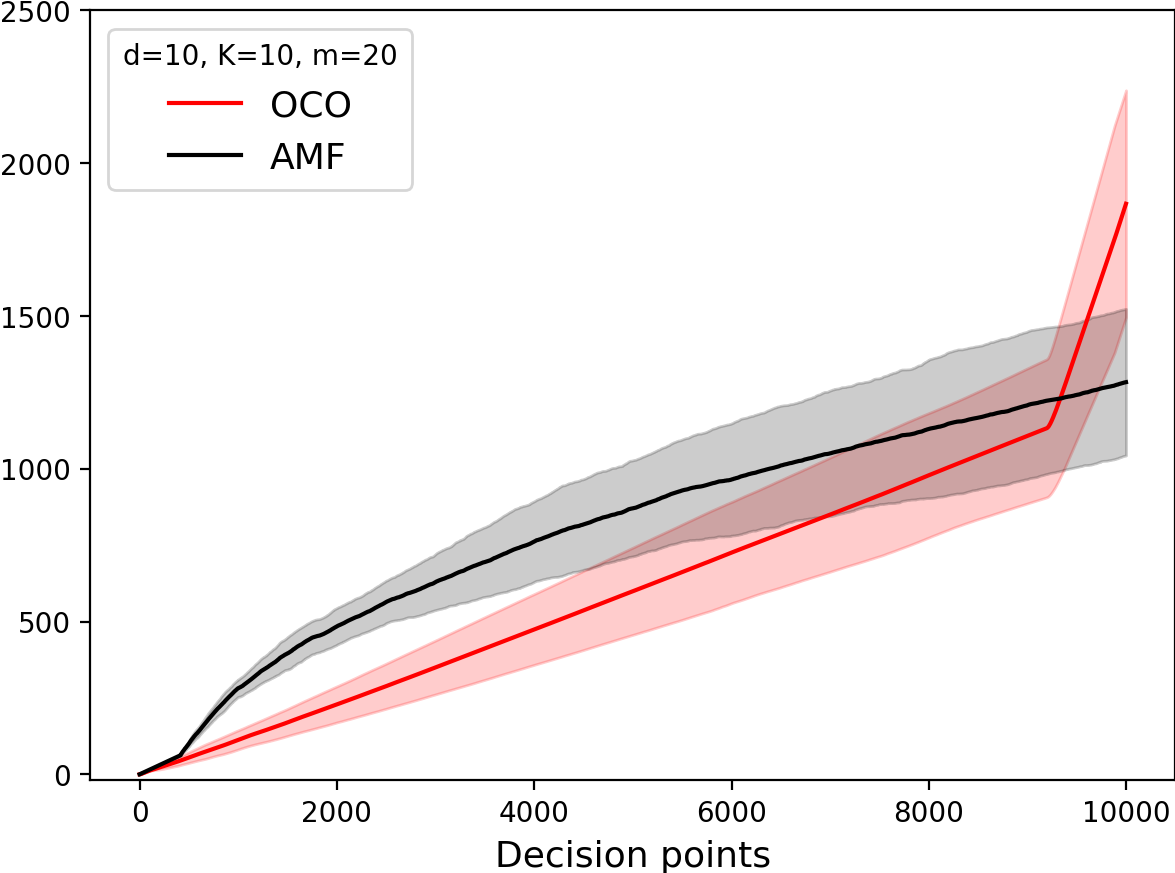}
\includegraphics[width=0.24\textwidth]{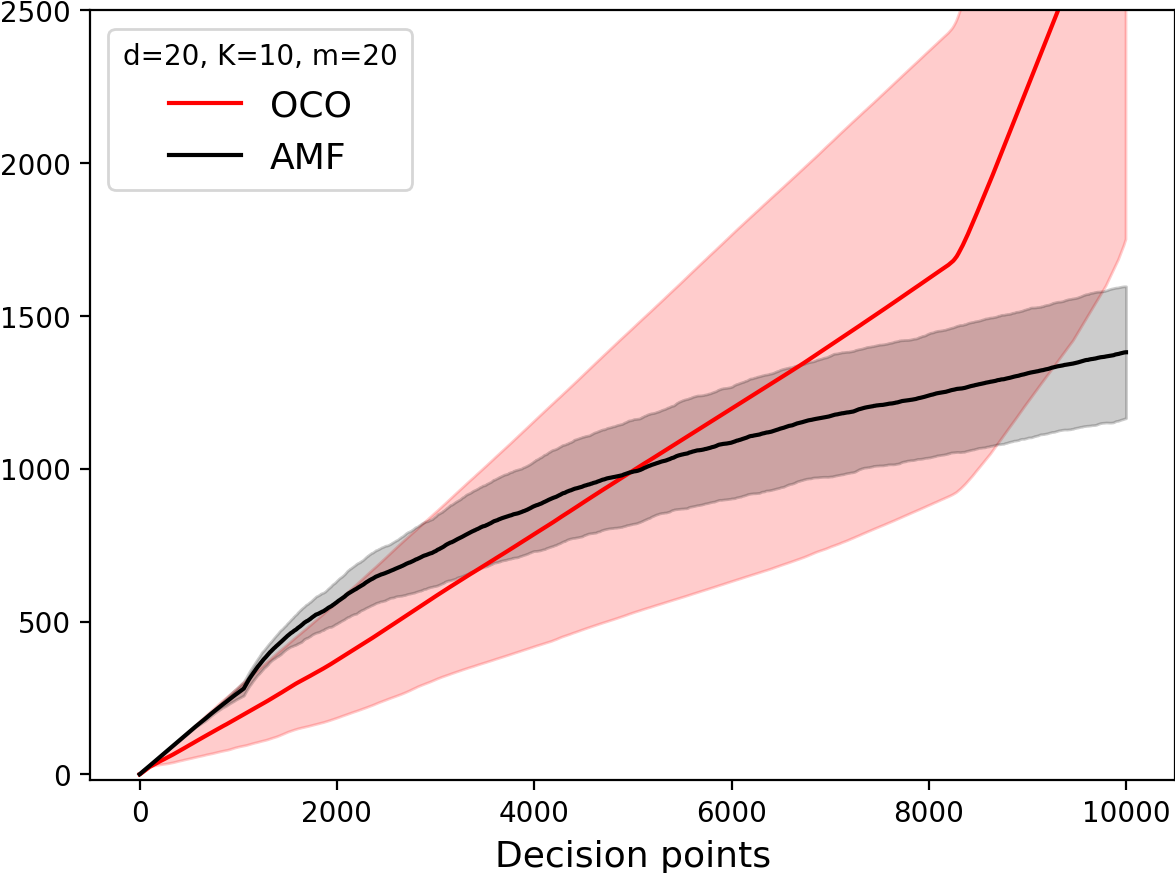}
}}
\subfigure[Regret comparison under $K=20$ and $m=20$]{{\includegraphics[width=0.24\textwidth]{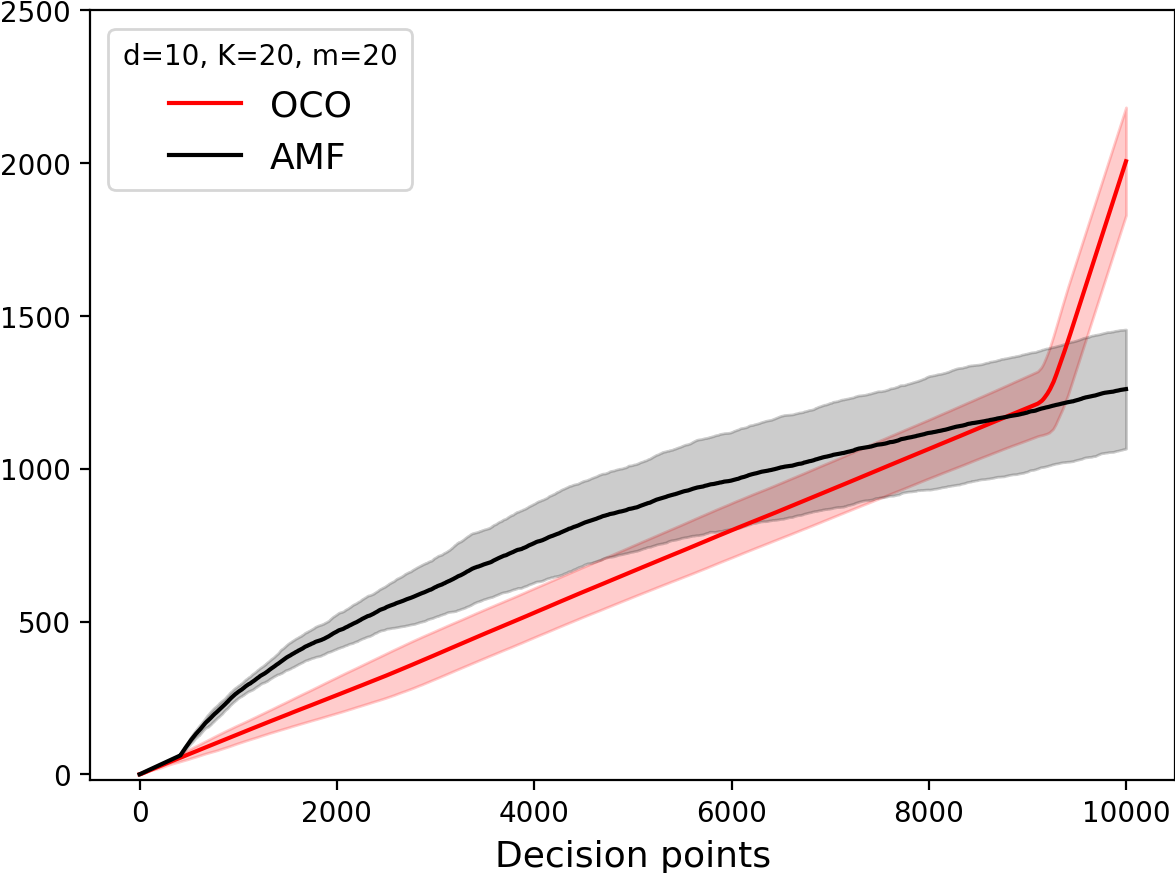}
\includegraphics[width=0.24\textwidth]{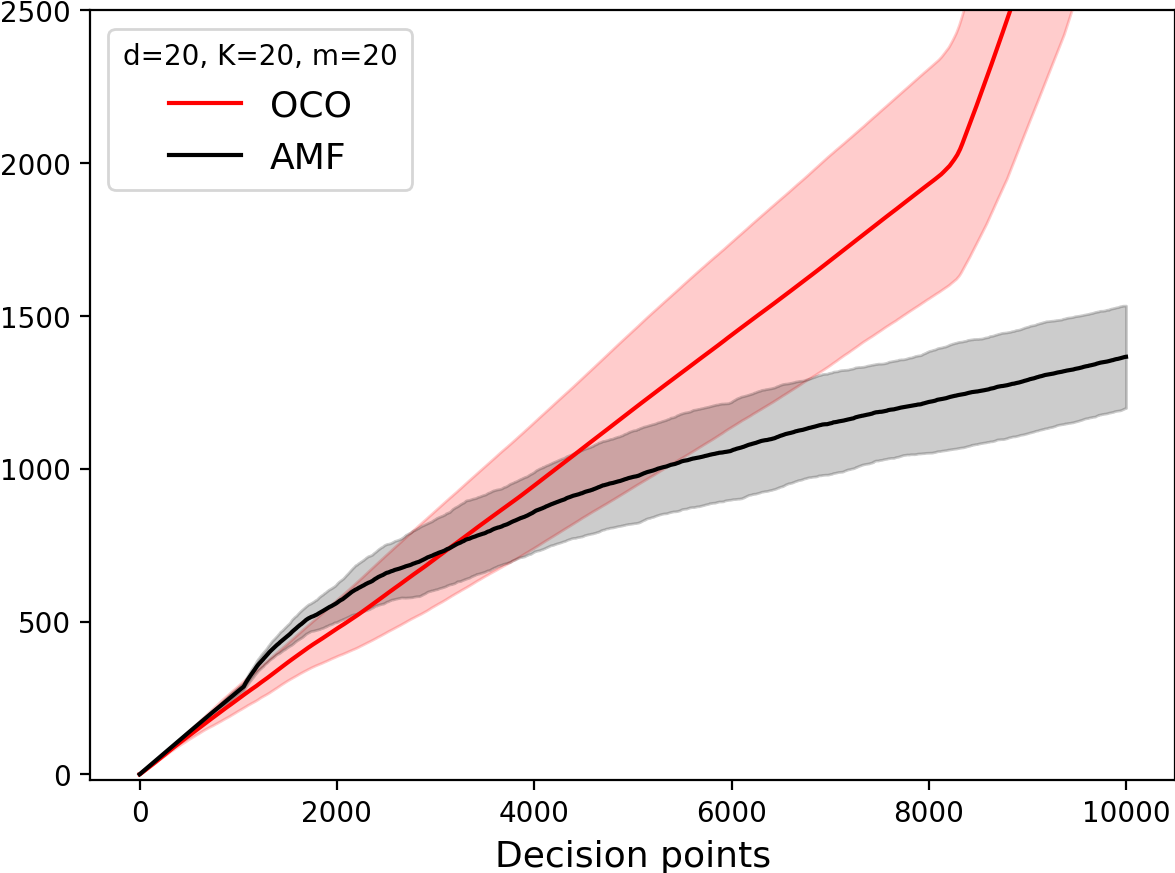}
}}
\caption{\label{fig:additional_comp_B05}Regret comparison of \texttt{AMF} and \texttt{OCO} algorithms under $B=\sqrt{dT}$. 
The line and shade represent the average and standard deviation based on 20 repeated experiments.}
\end{figure*}

\subsection{Additional Results on Regret Comparison.}
\label{subsec:additional_comp}

Figure \ref{fig:additional_comp_B075} (a)-(d) show the regret comparison of \texttt{AMF} and \texttt{OCO} on different terms of $K=10,20$, $m=10,20$, and $B=dT^{3/4}$. 
Similar to the results in Figure~\ref{fig:compB075}, our algorithm has less regret than \texttt{OCO} in all cases, especially at the end of the rounds. 
The crossing line occurs when our algorithm skips in the middle round when $\rho_{t}<0$ while \texttt{OCO} does not skip until the inventory runs out.

Figure~\ref{fig:additional_comp_B05} (a)-(d) show the regret of \texttt{AMF} and \texttt{OCO} algorithm on various $K=10, 20$ and $m=10, 20$ with budget $B=\sqrt{dT}$.
Even in the smaller budget, our algorithm \texttt{AMF} does not run out the inventory and gains more reward than \texttt{OCO}.
The gap of the performance tends to be larger than $B=\sqrt{d}T^{\frac{3}{4}}$ case.

\section{Missing Proofs}

\subsection{Proof of Theorem \ref{thm:self}}
\begin{proof}
Because the construction of $\widehat{\Theta}_{t}$ and $\widehat{\mathbf{W}}_{t}$ is the same,
the bound for the $\widehat{\mathbf{W}}_{t}$ follows immediately from the bound for $\widehat{\Theta}_{t}$ by
replacing $\{\Reward{\Class{\AdmitRound{\nu}}}{\Action{\AdmitRound{\nu}}}{\AdmitRound{\nu}}:\nu\in[n_{t}]\}$
with $m$ entries of $\{\Consumption{\Class{\AdmitRound{\nu}}}{\Action{\AdmitRound{\nu}}}{\AdmitRound{\nu}}:\nu\in[n_{t}]\}$.
Thus, it is sufficient to prove the bound for $\widehat{\Theta}_{t}$. 

\paragraph{Step 1. Estimation error decomposition:} Let us fix $t\in[T]$
throughout the proof. For each $\nu\in[n_{t}]$ and $k\in[K]$, denote
$\mathbf{X}_{k,\nu}:=\StackedContext k{\nu}\StackedContext k{\nu}^{\top}$.
Then we can write 
\begin{align*}
V_{n_{t}}:=	&\sum_{\nu\in\Psi_{n_{t}}}\sum_{k=1}^{K}\mathbf{X}_{k,\nu}+\sum_{\nu\notin\Psi_{n_{t}}}\mathbf{X}_{\Action{\AdmitRound{\nu}},\nu}+I_{J\cdot d},\\
A_{n_{t}}:=	&\sum_{\nu\in\Psi_{n_{t}}}\sum_{k=1}^{K}\frac{\Indicator{\PseudoAction{\nu}=k}}{\NewProb k{\nu}}\mathbf{X}_{k,\nu}+\sum_{\nu\notin\Psi_{n_{t}}}\mathbf{X}_{k,\nu}+I_{J\cdot d}.
\end{align*}
Denote the errors $\tilde{\eta}_{k,\nu}:=\TildeR k{\nu}-\StackedContext k{\nu}^{\top}\Theta_{\star}$ and $\Error k{\nu}:=\Reward{\Class{\AdmitRound{\nu}}}k{\AdmitRound{\nu}}-\StackedContext k{\nu}^{\top}\Theta_{\star}$.
By the definition of the estimator $\widehat{\Theta}_{n_t}$, 
\begin{equation}
\begin{split} 
&\norm{\StackedEstimator{n_{t}}-\Theta^{*}}_{V_{n_{t}}}\\
&=\norm{V_{n_{t}}^{-1/2}\left\{ -\Theta^{*}+\sum_{\nu\in\Psi_{n_{t}}}\sum_{k=1}^{K}\tilde{\eta}_{k,\nu}\StackedContext k{\nu}+\sum_{\nu\notin\Psi_{n_{t}}}\Error k{\nu}\StackedContext{\Action{\AdmitRound{\nu}}}{\nu}\right\} }_{2}\\
&\le\Maxeigen{V_{n_{t}}^{-1/2}}\norm{\Theta^{*}}_{2}+\norm{V_{n_{t}}^{-1/2}\left\{ \sum_{\nu\in\Psi_{n_{t}}}\sum_{k=1}^{K}\tilde{\eta}_{k,\nu}\StackedContext k{\nu}+\sum_{\nu\notin\Psi_{n_{t}}}\Error k{\nu}\StackedContext{\Action{\AdmitRound{\nu}}}{\nu}\right\} }_{2}\\
&\le\sqrt{Jd}+\norm{V_{n_{t}}^{-1/2}\left\{ \sum_{\nu\in\Psi_{n_{t}}}\sum_{k=1}^{K}\tilde{\eta}_{k,\nu}\StackedContext k{\nu}+\sum_{\nu\notin\Psi_{n_{t}}}\Error k{\nu}\StackedContext{\Action{\AdmitRound{\nu}}}{\nu}\right\} }_{2},
\end{split}
\label{eq:self1}
\end{equation}
where and the last inequality holds because $\norm{\Parameter j}_{2}\le\sqrt{d}$.
Plugging in $\TildeR k{\nu}$ defined in~\eqref{eq:pseudo_rewards},
\[
\tilde{\eta}_{k,\nu}\StackedContext k{\nu}=\left(1-\frac{\Indicator{\PseudoAction{\nu}=k}}{\NewProb k{\nu}}\right)\StackedContext k{\nu}\StackedContext k{\nu}^{\top}\left(\check{\Theta}_{t}-\Theta^{*}\right)+\frac{\Indicator{\PseudoAction{\nu}=k}}{\NewProb k{\nu}}\Error k{\nu}\StackedContext k{\nu},
\]
and the term $\sum_{\nu\in\Psi_{n_{t}}}\sum_{k=1}^{K}\tilde{\eta}_{k,\nu}\StackedContext k{\nu}$ is decomposed as,
\begin{equation}
\sum_{\nu\in\Psi_{n_{t}}}\sum_{k=1}^{K}\tilde{\eta}_{k,\nu}\StackedContext k{\nu}=\sum_{\nu\in\Psi_{n_{t}}}\sum_{k=1}^{K}\left\{ \left(1-\frac{\Indicator{\PseudoAction{\nu}=k}}{\NewProb k{\nu}}\right)\mathbf{X}_{k,\nu}\left(\check{\Theta}_{t}-\Theta^{*}\right)+\frac{\Indicator{\PseudoAction{\nu}=k}}{\NewProb k{\nu}}\Error k{\nu}\StackedContext k{\nu}\right\}. 
\label{eq:self2}
\end{equation}
By definition of the IPW estimator $\check{\Theta}_{t}$,
\begin{equation}
\begin{split}
&\sum_{\nu\in\Psi_{n_{t}}}\sum_{k=1}^{K}\left(1-\frac{\Indicator{\PseudoAction{\nu}=k}}{\NewProb k{\nu}}\right)\mathbf{X}_{k,\nu}\left(\check{\Theta}_{t}-\Theta^{*}\right)\\&=\left\{ \sum_{\nu\in\Psi_{n_{t}}}\sum_{k=1}^{K}\left(1-\frac{\Indicator{\PseudoAction{\nu}=k}}{\NewProb k{\nu}}\right)\mathbf{X}_{k,\nu}\right\} A_{n_{t}}^{-1}\left(-\Theta^{*}+\sum_{\nu\in\Psi_{n_{t}}}\sum_{k=1}^{K}\frac{\Indicator{\PseudoAction{\nu}=k}}{\NewProb k{\nu}}\Error k{\nu}\StackedContext k{\nu}+\sum_{\nu\notin\Psi_{n_{t}}}\Error{\Action{\AdmitRound{\nu}}}{\nu}\StackedContext{\Action{\AdmitRound{\nu}}}{\nu}\right)\\&=\left(V_{n_{t}}-A_{n_{t}}\right)A_{n_{t}}^{-1}\left(-\Theta^{*}+\sum_{\nu\in\Psi_{n_{t}}}\sum_{k=1}^{K}\frac{\Indicator{\PseudoAction{\nu}=k}}{\NewProb k{\nu}}\Error k{\nu}\StackedContext k{\nu}+\sum_{\nu\notin\Psi_{n_{t}}}\Error{\Action{\AdmitRound{\nu}}}{\nu}\StackedContext{\Action{\AdmitRound{\nu}}}{\nu}\right)\\&:=\left(V_{n_{t}}-A_{n_{t}}\right)A_{n_{t}}^{-1}\left(-\Theta^{*}+S_{n_{t}}\right),
\end{split}
\label{eq:self3}
\end{equation}
where
\[
S_{n_t}:=\sum_{\nu\in\Psi_{n_{t}}}\sum_{k=1}^{K}\frac{\Indicator{\PseudoAction{\nu}=k}}{\NewProb k{\nu}}\Error k{\nu}\StackedContext k{\nu}+\sum_{\nu\notin\Psi_{n_{t}}}\Error{\Action{\AdmitRound{\nu}}}{\nu}\StackedContext{\Action{\AdmitRound{\nu}}}{\nu},
\]
then,
\begin{align*}
\norm{\StackedEstimator{n_{t}}-\Theta^{*}}_{V_{n_{t}}}\underset{\eqref{eq:self1}}{\le}&\sqrt{Jd}+\norm{V_{n_{t}}^{-1/2}\left\{ \sum_{\nu\in\Psi_{n_{t}}}\sum_{k=1}^{K}\tilde{\eta}_{k,\nu}\StackedContext k{\nu}+\sum_{\nu\notin\Psi_{n_{t}}}\Error k{\nu}\StackedContext{\Action{\AdmitRound{\nu}}}{\nu}\right\} }_{2}\\
\underset{\eqref{eq:self2},\eqref{eq:self3}}{=}&\sqrt{Jd}+\norm{V_{n_{t}}^{-1/2}\left\{ \left(V_{n_{t}}-A_{n_{t}}\right)A_{n_{t}}^{-1}\left(-\Theta^{*}+S_{n_{t}}\right)+S_{n_{t}}\right\} }_{2}.
\end{align*}
By triangular inequality,
\begin{equation}
\begin{split}
\norm{\StackedEstimator t-\Theta^{*}}_{V_{n_{t}}}\le&\sqrt{Jd}+\norm{V_{n_{t}}^{-1/2}\left\{ \left(V_{n_{t}}-A_{n_{t}}\right)A_{n_{t}}^{-1}\left(-\Theta^{*}+S_{n_{t}}\right)+S_{n_{t}}\right\} }_{2}\\\le&\sqrt{Jd}+\norm{V_{n_{t}}^{-1/2}\left(V_{n_{t}}-A_{n_{t}}\right)A_{n_{t}}^{-1}\left(-\Theta^{*}+S_{n_{t}}\right)}_{2}+\norm{S_{n_{t}}}_{V_{n_{t}}^{-1}}\\\le&\sqrt{Jd}+\norm{\left(V_{n_{t}}^{1/2}A_{n_{t}}^{-1}V_{n_{t}}^{1/2}-I_{J\cdot d}\right)\left(-V_{n_{t}}^{-1/2}\Theta^{*}+V_{t}^{-1/2}S_{n_{t}}\right)}_{2}+\norm{S_{n_{t}}}_{V_{n_{t}}^{-1}}\\\le&\sqrt{Jd}+\norm{V_{n_{t}}^{1/2}A_{n_{t}}^{-1}V_{n_{t}}^{1/2}-I_{J\cdot d}}_{2}\norm{-V_{n_{t}}^{-1/2}\Theta^{*}+V_{t}^{-1/2}S_{n_{t}}}_{2}+\norm{S_{n_{t}}}_{V_{n_{t}}^{-1}}\\\le&\sqrt{Jd}+\norm{V_{n_{t}}^{1/2}A_{n_{t}}^{-1}V_{n_{t}}^{1/2}-I_{J\cdot d}}_{2}\left(\sqrt{Jd}+\norm{S_{n_{t}}}_{V_{n_{t}}^{-1}}\right)+\norm{S_{n_{t}}}_{V_{n_{t}}^{-1}}\\=&\left(\norm{V_{n_{t}}^{1/2}A_{n_{t}}^{-1}V_{n_{t}}^{1/2}-I_{J\cdot d}}_{2}+1\right)\left(\sqrt{Jd}+\norm{S_{n_{t}}}_{V_{n_{t}}^{-1}}\right).
\end{split}
\label{eq:self4}
\end{equation}

\paragraph{Step 2. Bounding the $\|\cdot\|_2$ of the matrix in \eqref{eq:self4}}
We claim that 
\begin{equation}
V_{n_{t}}^{1/2}A_{n_{t}}^{-1}V_{n_{t}}^{1/2}\succeq\frac{1}{8}I_{J\cdot d}
\label{eq:self_claim}
\end{equation}
Define $F_{n_{t}}:=\sum_{\nu=1}^{n_{t}}\sum_{k=1}^{K}\mathbf{X}_{k,\nu}+16Kd\log(\frac{Jd}{\delta}) I_{J\cdot d}$. 
Then we have $V_{n_{t}}\preceq F_{n_{t}}$ and $V_{n_{t}}^{1/2}A_{n_{t}}^{-1}V_{n_{t}}^{1/2}\succeq F_{n_{t}}^{-1/2}A_{n_{t}}F_{n_{t}}^{-1/2}$.
Now we decompose the matrix $A_{n_{t}}$ as
\begin{equation}
\begin{split}
F_{n_{t}}^{-1/2}A_{n_{t}}F_{n_{t}}^{-1/2}=&F_{n_{t}}^{-1/2}\left\{ \sum_{\nu=1}^{n_{t}}\sum_{k=1}^{K}\frac{\Indicator{\PseudoAction{\nu}=k}}{\NewProb k{\nu}}\mathbf{X}_{k,\nu}+I_{J\cdot d}\right\} F_{n_{t}}^{-1/2}
\\
&+F_{n_{t}}^{-1/2}\left[\sum_{\nu\notin\Psi_{n_{t}}}\left\{ \mathbf{X}_{\Action{\AdmitRound{\nu}},\nu}-\sum_{k=1}^{K}\frac{\Indicator{\PseudoAction{\nu}=k}}{\NewProb k{\nu}}\mathbf{X}_{k,\nu}\right\} \right]F_{n_{t}}^{-1/2}
\\
=&F_{n_{t}}^{-1/2}\left\{ \sum_{\nu=1}^{n_{t}}\sum_{k=1}^{K}\frac{\Indicator{\PseudoAction{\nu}=k}}{\NewProb k{\nu}}\mathbf{X}_{k,\nu}+16Kd\log\frac{Jd}{\delta}I_{J\cdot d}\right\} F_{n_{t}}^{-1/2}
\\
&+F_{n_{t}}^{-1/2}\left[\sum_{\nu\notin\Psi_{n_{t}}}\left\{ \mathbf{X}_{\Action{\AdmitRound{\nu}},\nu}-\sum_{k=1}^{K}\frac{\Indicator{\PseudoAction{\nu}=k}}{\NewProb k{\nu}}\mathbf{X}_{k,\nu}\right\} +\left(1-16Kd\log\frac{Jd}{\delta}\right)I_{J\cdot d}\right]F_{n_{t}}^{-1/2}
\end{split}
\label{eq:self5}
\end{equation}
For each $\nu\in[n_{t}]$, the matrix $\sum_{k=1}^{K}\frac{\Indicator{\PseudoAction{\nu}=k}}{\NewProb k{\nu}}F_{n_{t}}^{-1/2}\mathbf{X}_{k,\nu}F_{n_{t}}^{-1/2}$ symmetric positive definite and
\begin{equation}
\begin{split}\Maxeigen{8\log\frac{Jd}{\delta}\sum_{k=1}^{K}\frac{\Indicator{\PseudoAction{\nu}=k}}{\NewProb k{\nu}}F_{n_{t}}^{-1/2}\mathbf{X}_{k,\nu}F_{n_{t}}^{-1/2}}\le & 8\log\frac{Jd}{\delta}\sum_{k=1}^{K}\frac{\Indicator{\PseudoAction{\nu}=k}}{\NewProb k{\nu}}\Maxeigen{F_{n_{t}}^{-1}}\\
\le & 8\log\frac{Jd}{\delta}\frac{\Mineigen{F_{\nu}}}{16\log\left(\frac{Jd}{\delta}\right)}\Maxeigen{F_{n_{t}}^{-1}}\\
\le & \frac{1}{2}\frac{\Mineigen{F_{\nu}}}{\Mineigen{F_{n_{t}}}}\\
\le & \frac{1}{2}.
\end{split}
\label{eq:self6}
\end{equation}
With the filtration $\Filtration 0:=\History t$ and $\Filtration n:=\Filtration 0\cup\{\PseudoAction{\nu}:\nu\in[n]\}$,
we use Lemma \ref{lem:matrix_neg} to have with probability at least $1-\delta$,
\begin{align*}
&8\log\frac{Jd}{\delta}F_{n_{t}}^{-1/2}\left\{ \sum_{\nu=1}^{n_{t}}\sum_{k=1}^{K}\frac{\Indicator{\PseudoAction{\nu}=k}}{\NewProb k{\nu}}\mathbf{X}_{k,\nu}+16Kd\log\frac{Jd}{\delta}I_{J\cdot d}\right\} F_{n_{t}}^{-1/2}
\\
&\succeq8\log\frac{Jd}{\delta}F_{n_{t}}^{-1/2}\left\{ \sum_{\nu=1}^{n_{t}}\sum_{k=1}^{K}\mathbf{X}_{k,\nu}+I_{J\cdot d}\right\} F_{n_{t}}^{-1/2}
\\
&=4\log\frac{Jd}{\delta}I_{J\cdot d}-\log\frac{Jd}{\delta}I_{J\cdot d}\\&=3\log\frac{Jd}{\delta}I_{J\cdot d},
\end{align*}
which implies
\begin{equation}
F_{n_{t}}^{-1/2}\left\{ \sum_{\nu=1}^{n_{t}}\sum_{k=1}^{K}\frac{\Indicator{\PseudoAction{\nu}=k}}{\NewProb k{\nu}}\mathbf{X}_{k,\nu}+16Kd\log\frac{Jd}{\delta}I_{J\cdot d}\right\} F_{n_{t}}^{-1/2}\succeq\frac{3}{8}I_{J\cdot d},
\label{eq:self7}
\end{equation}
and the left hand side of \eqref{eq:self5} is bounded as
\begin{equation}
F_{n_{t}}^{-1/2}A_{n_{t}}F_{n_{t}}^{-1/2}\succeq\frac{3}{8}I_{J\cdot d}+F_{n_{t}}^{-1/2}\left[\sum_{\nu\notin\Psi_{n_{t}}}\left\{ \mathbf{X}_{\Action{\AdmitRound{\nu}},\nu}-\sum_{k=1}^{K}\frac{\Indicator{\PseudoAction{\nu}=k}}{\NewProb k{\nu}}\mathbf{X}_{k,\nu}\right\} +\left(1-16Kd\log\frac{Jd}{\delta}\right)I_{J\cdot d}\right]F_{n_{t}}^{-1/2}.
\label{eq:self8}
\end{equation}
To bound the other term, observe that for $\nu\notin\Psi_{n_{t}}$,
\[
\CE{\sum_{k=1}^{K}\frac{\Indicator{\PseudoAction{\nu}=k}}{\NewProb k{\nu}}\mathbf{X}_{k,\nu}}{\History t}=\sum_{i\neq\Action{\AdmitRound{\nu}}}\sum_{k=1}^{K}\NewProb i{\nu}\frac{\Indicator{i=k}}{\NewProb k{\nu}}\mathbf{X}_{k,\nu}=\sum_{k\neq\Action{\AdmitRound{\nu}}}\mathbf{X}_{k,\nu}.
\]
Because \eqref{eq:self6} holds for $\nu\notin\Psi_{n_{t}}$, we can use
Lemma \ref{lem:matrix_neg}, to have with probability at least $1-\delta$
\[
8\log\frac{Jd}{\delta}F_{n_{t}}^{-1/2}\left[\sum_{\nu\notin\Psi_{n_{t}}}\sum_{k=1}^{K}\frac{\Indicator{\PseudoAction{\nu}=k}}{\NewProb k{\nu}}\mathbf{X}_{k,\nu}\right]F_{n_{t}}^{-1/2}\preceq12\log\frac{Jd}{\delta}F_{n_{t}}^{-1/2}\left[\sum_{\nu\notin\Psi_{n_{t}}}\sum_{k\neq\Action{\AdmitRound{\nu}}}\mathbf{X}_{k,\nu}\right]F_{n_{t}}^{-1/2}+\log\frac{Jd}{\delta}I_{J\cdot d}.
\]
Rearranging the terms,
\[
F_{n_{t}}^{-1/2}\left[\sum_{\nu\notin\Psi_{n_{t}}}\sum_{k=1}^{K}\frac{\Indicator{\PseudoAction{\nu}=k}}{\NewProb k{\nu}}\mathbf{X}_{k,\nu}\right]F_{n_{t}}^{-1/2}\preceq\frac{3}{2}F_{n_{t}}^{-1/2}\left[\sum_{\nu\notin\Psi_{n_{t}}}\sum_{k\neq\Action{\AdmitRound{\nu}}}\mathbf{X}_{k,\nu}\right]F_{n_{t}}^{-1/2}+\frac{1}{8}I_{J\cdot d}.
\]
Thus the second term in \eqref{eq:self8} is bounded as, 
\begin{equation}
\begin{split} & F_{n_{t}}^{-1/2}\left[\sum_{\nu\notin\Psi_{n_{t}}}\left\{ \mathbf{X}_{\Action{\AdmitRound{\nu}},\nu}-\sum_{k=1}^{K}\frac{\Indicator{\PseudoAction{\nu}=k}}{\NewProb k{\nu}}\mathbf{X}_{k,\nu}\right\} \right]F_{n_{t}}^{-1/2}\\
 & \succeq F_{n_{t}}^{-1/2}\left[\sum_{\nu\notin\Psi_{n_{t}}}\left\{ \mathbf{X}_{\Action{\AdmitRound{\nu}},\nu}-\frac{3}{2}\sum_{\nu\notin\Psi_{t}}\sum_{k\neq\Action{\AdmitRound{\nu}}}\mathbf{X}_{k,\nu}\right\} \right]F_{n_{t}}^{-1/2}-\frac{1}{8}I_{J\cdot d}\\
 & \succeq-\frac{3}{2}F_{n_{t}}^{-1/2}\left[\sum_{\nu\notin\Psi_{n_{t}}}\sum_{k=1}^{K}\mathbf{X}_{k,\nu}\right]F_{n_{t}}^{-1/2}-\frac{1}{8}I_{J\cdot d}\\
 & \succeq-\left\{ \frac{3dK}{2}\abs{\Psi_{n_{t}}^{c}}\Maxeigen{F_{n_{t}}^{-1}}+\frac{1}{8}\right\} I_{J\cdot d},
\end{split}
\end{equation}
where the last inequality holds by $\Maxeigen{\mathbf{X}_{k,\nu}}\le d$.
Plugging in~\eqref{eq:self8},
\begin{equation}
\begin{split}
F_{n_{t}}^{-1/2}A_{n_{t}}F_{n_{t}}^{-1/2}\succeq&\frac{1}{4}I_{J\cdot d}-\left\{ \frac{3dK}{2}\abs{\Psi_{n_{t}}^{c}}\Maxeigen{F_{n_{t}}^{-1}}\right\} I_{J\cdot d}-\left(16Kd\log\frac{Jd}{\delta}-1\right)F_{n_{t}}^{-1}
\\
\succeq&\frac{1}{4}I_{J\cdot d}-\left(\frac{3dK}{2}\abs{\Psi_{n_{t}}^{c}}+16Kd\log\frac{Jd}{\delta}\right)\Maxeigen{F_{n_{t}}^{-1}}I_{J\cdot d}.
\label{eq:self9}
\end{split}
\end{equation}
By Lemma \ref{lem:matrix_neg}, with probability at least $1-\delta$,
\[
\frac{1}{2}\abs{\Psi_{n_{t}}^{c}}=\frac{1}{2}\sum_{\nu=1}^{n_{t}}\Indicator{\PseudoAction{\nu}\neq\Action{\AdmitRound{\nu}}}\le \frac{3}{2}\sum_{\nu=1}^{n_{t}}\sum_{k\neq\Action{\AdmitRound{\nu}}}\NewProb k{\nu}+\log\frac{1}{\delta},
\]
which implies
\begin{equation}
\begin{split}
&\left(\frac{3dK}{2}\abs{\Psi_{n_{t}}^{c}}+16Kd\log\frac{Jd}{\delta}\right)\Maxeigen{F_{n_{t}}^{-1}}
\\
&\le\frac{dK}{2\Mineigen{F_{n_{t}}}}\left(9\sum_{\nu=1}^{n_{t}}\sum_{k\neq\Action{\AdmitRound{\nu}}}\NewProb k{\nu}+3\log\frac{1}{\delta}+32\log\frac{Jd}{\delta}\right)
\\
&\le\frac{dK}{2\Mineigen{F_{n_{t}}}}\left(9\sum_{\nu=1}^{n_{t}}\sum_{k\neq\Action{\AdmitRound{\nu}}}\NewProb k{\nu}+35\log\frac{Jd}{\delta}\right)
\\
&=\frac{dK}{2\Mineigen{F_{n_{t}}}}\left(\sum_{\nu=1}^{n_{t}}\frac{144\left(K-1\right)\log\frac{Jd}{\delta}}{\Mineigen{F_{\nu}}}+35\log\frac{Jd}{\delta}\right)
\\
&\le\frac{1}{8},
\end{split}
\label{eq:self10}
\end{equation}
where the last inequality holds by the assumption \eqref{eq:self_condition}.
Plugging in \eqref{eq:self9}, with probability at least $1-2\delta$,
\[
F_{n_{t}}^{-1/2}\left[\sum_{\nu\notin\Psi_{n_{t}}}\left\{ \mathbf{X}_{\Action{\AdmitRound{\nu}},\nu}-\sum_{k=1}^{K}\frac{\Indicator{\PseudoAction{\nu}=k}}{\NewProb k{\nu}}\mathbf{X}_{k,\nu}\right\} \right]F_{n_{t}}^{-1/2}\succeq-\frac{1}{4}I_{J\cdot d}.
\]
With \eqref{eq:self8},
\[
F_{n_{t}}^{-1/2}A_{n_{t}}F_{n_{t}}^{-1/2}\succeq\frac{1}{8}I_{J\cdot d},
\]
which proves \eqref{eq:self_claim} and the claim implies 
\[
\norm{V_{n_{t}}^{1/2}A_{n_{t}}^{-1}V_{n_{t}}^{1/2}-I_{J\cdot d}}_{2} \le 7.
\]

\paragraph{Step 3. Bounding the self-normalized vector-valued martingale $S_{n_t}$}
Let $\Filtration 0$ be a sigma algebra generated by contexts $\{\Context{\Class s}ks:k\in[K],s\in[t]\}$,
and $\Psi_{t}$. Define filtration as $\Filtration{\nu}:=\sigma(\Filtration 0\cup\History{\AdmitRound{\nu+1}})$.
Then $S_{\nu}$ is a $\Real^{J\cdot d}$-valued martingale because
\begin{align*}
\CE{S_{\nu}-S_{\nu-1}}{\Filtration{\nu-1}}&=\CE{\Indicator{\nu\in\Psi_{n_{t}}}\sum_{k=1}^{K}\frac{\Indicator{\PseudoAction{\nu}=k}}{\NewProb k{\nu}}\Error k{\nu}\StackedContext k{\nu}+\Indicator{\nu\notin\Psi_{n_{t}}}\Error{\Action{\AdmitRound{\nu}}}{\AdmitRound{\nu}}\StackedContext k{\nu}}{\Filtration{\nu-1}}\\
&=\CE{\Indicator{\nu\in\Psi_{n_{t}}}\sum_{k=1}^{K}\frac{\Indicator{\Action{\AdmitRound{\nu}}=k}}{\NewProb k{\nu}}\Error k{\nu}\StackedContext k{\nu}+\Indicator{\nu\notin\Psi_{n_{t}}}\Error{\Action{\AdmitRound{\nu}}}{\AdmitRound{\nu}}\StackedContext k{\nu}}{\Filtration{\nu-1}}\\
&=\CE{\left\{ \frac{\Indicator{\nu\in\Psi_{n_{t}}}}{\NewProb{\Action{\AdmitRound{\nu}}}{\nu}}+\Indicator{\nu\notin\Psi_{n_{t}}}\right\} \Error{\Action{\AdmitRound{\nu}}}{\nu}\StackedContext k{\nu}}{\Filtration{\nu-1}}\\
&=\CE{\left\{ \frac{\Indicator{\nu\in\Psi_{n_{t}}}}{\NewProb{\Action{\AdmitRound{\nu}}}{\nu}}+\Indicator{\nu\notin\Psi_{n_{t}}}\right\} \Error{\Action{\AdmitRound{\nu}}}{\nu}\StackedContext k{\nu}}{\History{\AdmitRound{\nu}}}\\
&=0,
\end{align*}
where the second equality holds by definition of $\Psi_{n_t}$ and
the fourth inequality holds because the distribution of $\{\Context{\Class s}ks:k\in[K],s\in(\AdmitRound{\nu},t]\}$
is independent of $\History{\AdmitRound{\nu}}$ by Assumption \ref{assum:independent_contexts}.
By Assumption~\ref{assum:error}, for any $\lambda\in\Real,$
\begin{align*}
\CE{\exp\left[\lambda\left\{ \frac{\Indicator{\nu\in\Psi_{n_{t}}}}{\NewProb{\Action{\AdmitRound{\nu}}}{\nu}}\!+\!\Indicator{\nu\notin\Psi_{n_{t}}}\right\} \Error k{\Action{\AdmitRound{\nu}}}\right]}{\Filtration{\nu-1}}&\le\CE{\exp\left[\frac{\lambda^{2}\sigma^{2}}{2}\left\{ \frac{\Indicator{\nu\in\Psi_{n_{t}}}}{\NewProb{\Action{\AdmitRound{\nu}}}{\nu}}\!+\!\Indicator{\nu\notin\Psi_{n_{t}}}\right\} ^{2}\right]}{\Filtration{\nu-1}}\\
&\le\exp\left[2\lambda^{2}\sigma_{r}^{2}\right],
\end{align*}
Thus, $\left\{ \frac{\Indicator{\nu\in\Psi_{n_{t}}}}{\NewProb{\Action{\AdmitRound{\nu}}}{\nu}}+\Indicator{\nu\notin\Psi_{n_{t}}}\right\} \Error k{\Action{\AdmitRound{\nu}}}$ is $2\sigma_r$-sub-Gaussian. 
Because
\begin{align*}
\norm{S_{n_t}}_{V_{t}^{-1}}= & \norm{\sum_{\nu\in\Psi_{n_{t}}}\sum_{k=1}^{K}\frac{\Indicator{\PseudoAction{\nu}=k}}{\NewProb k{\nu}}\Error k{\nu}\StackedContext k{\nu}+\sum_{\nu\notin\Psi_{n_{t}}}\Error{\Action{\AdmitRound{\nu}}}{\nu}\StackedContext{\Action{\AdmitRound{\nu}}}{\nu}}_{V_{n_{t}}^{-1}}\\
= & \norm{\sum_{\nu=1}^{n_{t}}\left\{ \frac{\Indicator{\nu\in\Psi_{n_{t}}}}{\NewProb{\Action{\AdmitRound{\nu}}}{\nu}}+\Indicator{\nu\notin\Psi_{n_{t}}}\right\} \Error{\Action{\AdmitRound{\nu}}}{\nu}\StackedContext k{\nu}}_{V_{n_{t}}^{-1}}\\
= & \norm{\sum_{\nu=1}^{n_{t}}\left\{ \frac{\Indicator{\nu\in\Psi_{n_{t}}}}{\NewProb{\Action{\AdmitRound{\nu}}}{\nu}}+\Indicator{\nu\notin\Psi_{n_{t}}}\right\} \Error{\Action{\AdmitRound{\nu}}}{\nu}V_{n_{t}}^{-1/2}\StackedContext k{\nu}}_{2},
\end{align*}
by Lemma \ref{lem:eta_x_bound}, with probability at least $1-\delta$,
\begin{align*}
\norm{S_{t}}_{V_{t}^{-1}}\le & \norm{\sum_{\nu=1}^{n_{t}}\left\{ \frac{\Indicator{\nu\in\Psi_{n_{t}}}}{\NewProb{\Action{\AdmitRound{\nu}}}{\nu}}+\Indicator{\nu\notin\Psi_{t}}\right\} \Error{\Action{\AdmitRound{\nu}}}{\nu}V_{n_{t}}^{-1/2}\StackedContext k{\nu}}_{2},\\
\le & 12\sigma_{r}\sqrt{\sum_{\nu=1}^{n_{t}}\norm{V_{n_{t}}^{-1/2}\StackedContext k{\nu}}_{2}^{2}\log\frac{4}{\delta}}\\
\le & 12\sigma_{r}\sqrt{Jd\log\frac{4}{\delta}},
\end{align*}
where the last inequality holds because
\begin{align*}
\sum_{\nu=1}^{n_{t}}\norm{V_{n_{t}}^{-1/2}\StackedContext k{\nu}}_{2}^{2}&=\sum_{\nu=1}^{n_{t}}\StackedContext k{\nu}^{\top}V_{n_{t}}^{-1}\StackedContext k{\nu}=\Trace{\sum_{\nu=1}^{n_{t}}\StackedContext k{\nu}^{\top}V_{n_{t}}^{-1}\StackedContext k{\nu}}\\&=\Trace{\sum_{\nu=1}^{n_{t}}\StackedContext k{\nu}\StackedContext k{\nu}^{\top}V_{n_{t}}^{-1}}\le\Trace{V_{n_{t}}V_{n_{t}}^{-1}}=Jd.
\end{align*}
With \eqref{eq:self4}, the proof is completed
\end{proof}

\subsection{Proof of Theorem~\ref{thm:u_convergence}}
\begin{proof}
Similar to the proof of Theorem~\ref{thm:self}, the bound for consumption vector immediately follows from the bound for the utilities.
Therefore we provide the proof for the utility bound.

\paragraph{Step 1. Decomposition:} 
For each $k\in[K]$ and $j\in[J]$, 
\begin{align*}
\abs{\OptUtility jk-\TildeUtility jk{t+1}}\le & \abs{\Expectation\left[\mathbf{x}_{k}^{(j)}\right]^{\top}\Parameter j-\left(\frac{1}{\sum_{s=1}^{t+1}\Indicator{\Class s=j}}\sum_{s=1}^{t+1}\Indicator{\Class s=j}\left\{ \Context jks\right\} ^{\top}\Estimator jt\right)}
\\
\le & \abs{\Expectation\left[\mathbf{x}_{k}^{(j)}\right]^{\top}\Parameter j-\left(\frac{1}{\sum_{s=1}^{t+1}\Indicator{\Class s=j}}\sum_{s=1}^{t+1}\Indicator{\Class s=j}\left\{ \Context jks\right\} ^{\top}\Parameter j\right)}
\\ 
& +\abs{\frac{1}{\sum_{s=1}^{t+1}\Indicator{\Class s=j}}\sum_{s=1}^{t+1}\Indicator{\Class s=j}\left(\Estimator jt-\Parameter j\right)^{\top}\Context jks}\\
& \abs{\frac{1}{\sum_{s=1}^{t+1}\Indicator{\Class s=j}}\sum_{s=1}^{t+1}\Indicator{\Class s=j}\left(\Expectation\left[\mathbf{x}_{k}^{(j)}\right]^{\top}\Parameter j-\left\{ \Context jks\right\} ^{\top}\Parameter j\right)}
\\
& +\abs{\frac{1}{\sum_{s=1}^{t+1}\Indicator{\Class s=j}}\sum_{s=1}^{t+1}\Indicator{\Class s=j}\left(\Estimator jt-\Parameter j\right)^{\top}\Context jks}.
\end{align*}
Taking maximum over $k\in[K]$ gives the decomposition,
\begin{equation}
\begin{split}
\max_{k\in[K]}\abs{\OptUtility jk-\TildeUtility jk{t+1}}
\le & \max_{k\in[K]}\abs{\frac{1}{\sum_{s=1}^{t+1}\Indicator{\Class s=j}}\sum_{s=1}^{t+1}\Indicator{\Class s=j}\left(\Expectation\left[\mathbf{x}_{k}^{(j)}\right]^{\top}\Parameter j-\left\{ \Context jks\right\} ^{\top}\Parameter j\right)}
\\
& +\max_{k\in[K]}\abs{\frac{1}{\sum_{s=1}^{t+1}\Indicator{\Class s=j}}\sum_{s=1}^{t+1}\Indicator{\Class s=j}\left(\Estimator jt-\Parameter j\right)^{\top}\Context jks}.
\end{split}
\label{eq:util1}
\end{equation}

\paragraph{Step 2. Bounding the difference between expectation and empirical distribution:}
The random variables $\left\{ \left\{ \Context jks\right\} ^{\top}\Parameter j:s\in[t]\right\}$ are IID by Assumption \ref{assum:independent_contexts}.
Using Lemma \ref{lem:Azuma_Hoeffding_ineqaulity}, 
\begin{align*}
&\abs{\frac{1}{\sum_{s=1}^{t+1}\Indicator{\Class s=j}}\sum_{s=1}^{t+1}\Indicator{\Class s=j}\left(\Expectation\left[\mathbf{x}_{k}^{(j)}\right]^{\top}\Parameter j-\left\{ \Context jks\right\} ^{\top}\Parameter j\right)}\\&=\frac{1}{\sum_{s=1}^{t+1}\Indicator{\Class s=j}}\abs{\sum_{s=1}^{t+1}\Indicator{\Class s=j}\left(\Expectation\left[\mathbf{x}_{k}^{(j)}\right]^{\top}\Parameter j-\left\{ \Context jks\right\} ^{\top}\Parameter j\right)}\\&\le\frac{4}{\sqrt{\sum_{s=1}^{t+1}\Indicator{\Class s=j}}}\sqrt{\log JKT}.
\end{align*}
with probability at least $1-3(JKT)^{-1}$. 
By Lemma \ref{lem:matrix_neg}, with probability at least $1-(JT)^{-1}$,
\begin{equation}
\sum_{s=1}^{t+1}\Indicator{\Class s=j}\ge\frac{1}{2}p_{j}\left(t+1\right)-2\log J T\ge\frac{1}{4}p_{j}\left(t+1\right),
\label{eq:util2}
\end{equation}
where the last inequality holds by the assumption $t\ge8d\alpha^{-1} p_{\min}^{-1}\log JT$.
Summing up the probability bounds, with probability at least $1-5T^{-1}$,
\begin{align*}
\max_{k\in[K]}\abs{\frac{1}{\sum_{s=1}^{t+1}\Indicator{\Class s=j}}\sum_{s=1}^{t+1}\Indicator{\Class s=j}\left(\Expectation\left[\mathbf{x}_{k}^{(j)}\right]^{\top}\Parameter j-\left\{ \Context jks\right\} ^{\top}\Parameter j\right)}\le&\frac{4}{\sqrt{\sum_{s=1}^{t+1}\Indicator{\Class s=j}}}\sqrt{\log JKT}
\\
\le&\frac{8}{\sqrt{p_{j}\left(t+1\right)}}\sqrt{\log JKT}.
\end{align*}
Plugging in the decomposition \eqref{eq:util1}, for each $j\in[J]$, 
\begin{align*}
\max_{k\in[K]}\abs{\OptUtility jk-\TildeUtility jk{t+1}}\le&\frac{16}{\sqrt{p_{j}\left(t+1\right)}}\sqrt{\log JKT}\\&+\max_{k\in[K]}\abs{\frac{1}{\sum_{s=1}^{t+1}\Indicator{\Class s=j}}\sum_{s=1}^{t+1}\Indicator{\Class s=j}\left(\Estimator j{t-1}-\Parameter j\right)^{\top}\Context jks}.
\end{align*}
Taking square and summing up over $j\in[J]$ gives
\begin{equation}
\begin{split}
\sum_{j=1}^{J}p_{j}\max_{k\in[K]}\abs{\OptUtility jk-\TildeUtility jk{t+1}}^{2}\le&\frac{16J\log JKT}{t+1}\\&+\sum_{j=1}^{J}p_{j}\max_{k\in[K]}\abs{\frac{1}{\sum_{s=1}^{t+1}\Indicator{\Class s=j}}\sum_{s=1}^{t+1}\Indicator{\Class s=j}\left(\Estimator jt-\Parameter j\right)^{\top}\Context jks}^{2},
\end{split}
\label{eq:util3}
\end{equation}

\paragraph{Step 3. Bounding the prediction error:} 
By Cauchy-Schwartz inequality and \eqref{eq:util2}, 
\begin{equation}
\begin{split} & \sum_{j=1}^{J}\max_{k\in[K]}p_{j}\frac{1}{\left\{ \sum_{s=1}^{t+1}\Indicator{\Class s=j}\right\} ^{2}}\abs{\sum_{s=1}^{t+1}\Indicator{\Class s=j}\left(\Estimator jt-\Parameter j\right)^{\top}\Context jks}^{2}\\
 & \le\sum_{j=1}^{J}\max_{k\in[K]}p_{j}\frac{1}{\sum_{s=1}^{t+1}\Indicator{\Class s=j}}\sum_{s=1}^{t+1}\Indicator{\Class s=j}\left\{ \left(\Estimator jt-\Parameter j\right)^{\top}\Context jks\right\} ^{2}\\
 & \le\sum_{j=1}^{J}\max_{k\in[K]}\frac{4p_{j}}{p_{j}\left(t+1\right)}\sum_{s=1}^{t+1}\Indicator{\Class s=j}\left\{ \left(\Estimator jt-\Parameter j\right)^{\top}\Context jks\right\} ^{2}\\
 & =\frac{4}{\left(t+1\right)}\sum_{j=1}^{J}\max_{k\in[K]}\left(\Estimator jt-\Parameter j\right)^{\top}\left\{ \sum_{s=1}^{t+1}\Indicator{\Class s=j}\Context jks\left(\Context jks\right)^{\top}\right\} \left(\Estimator jt-\Parameter j\right)\\
 & \le\frac{4}{\left(t+1\right)}\sum_{j=1}^{J}\left(\Estimator jt-\Parameter j\right)^{\top}\left\{ \sum_{s=1}^{t+1}\sum_{k=1}^{K}\Indicator{\Class s=j}\Context jks\left(\Context jks\right)^{\top}\right\} \left(\Estimator jt-\Parameter j\right)\\
 & =\frac{4}{\left(t+1\right)}\left(\StackedEstimator t-\Theta^{\star}\right)^{\top}\left\{ \sum_{s=1}^{t+1}\sum_{k=1}^{K}\StackedContext ks\StackedContext ks^{\top}\right\} \left(\StackedEstimator t-\Theta^{\star}\right),
\end{split}
\label{eq:util4}
\end{equation}
where $\Theta^{\star}:=(\Parameter 1,\ldots,\Parameter J)^{T}\in\Real^{J\cdot d}$
and 
\[
\StackedContext ks:=\begin{pmatrix}\mathbf{0}_{d}\\
\vdots\\
\Context{\Class s}ks\\
\mathbf{0}_{d}
\end{pmatrix}\in\Real^{J\cdot d},
\]
where the context $\Context{\Class s}ks$ is located after $\Class s-1$ of $\mathbf{0}_{d}$ vectors. 
We claim that
\begin{equation}
\frac{1}{t+1}\sum_{s=1}^{t+1}\sum_{k=1}^{K}\StackedContext ks\StackedContext ks^{\top}\preceq2\Expectation\left[\StackedContext k1\StackedContext k1^{\top}\right]\preceq\frac{8}{\abs{\Psi_{n_{t}}}}\sum_{s\in\Psi_{n_{t}}}\sum_{k=1}^{K}\StackedContext ks\StackedContext ks^{\top},
\label{eq:util_claim}
\end{equation}
with probability at least $1-2T^{-1}$.
The matrix $\mathbf{X}_{s}:=\sum_{k=1}^{K}\StackedContext ks\StackedContext ks^{\top}$is
symmetric nonnegative definite which satisfies 
\begin{align*}
\Maxeigen{\frac{1}{2dK}\mathbf{X}_{s}} & \le\frac{1}{2}.
\end{align*}
By Lemma \ref{lem:matrix_neg}, with probability at least $1-T^{-1}$,
\[
\frac{1}{2Kd}\sum_{s=1}^{t+1}\mathbf{X}_{s}\preceq\frac{3}{4Kd}\sum_{s=1}^{t+1}\Expectation\left[\mathbf{X}_{s}\right]+\left(\log JdT\right)I_{J\cdot d},
\]
which implies
\begin{equation}
\frac{1}{t+1}\sum_{s=1}^{t+1}\mathbf{X}_{s}\preceq\frac{3}{2\left(t+1\right)}\sum_{s=1}^{t+1}\Expectation\left[\mathbf{X}_{s}\right]+\frac{2dK}{t+1}\left(\log JdT\right)I_{J\cdot d}.
\label{eq:util5}
\end{equation}
By Assumption \ref{assum:positive_definiteness}, for $s\in[t+1]$,
\begin{align*}
\Mineigen{\Expectation\left[\mathbf{X}_{s}\right]}= & \Mineigen{\begin{pmatrix}p_{1}\Expectation_{\mathbf{x}_{k}\sim\Classdistribution 1}\left[\sum_{k=1}^{K}\mathbf{x}_{k}\mathbf{x}_{k}^{\top}\right] & 0 & 0\\
0 & \ddots & 0\\
0 & 0 & p_{J}\Expectation_{\mathbf{x}_{k}\sim\Classdistribution J}\left[\sum_{k=1}^{K}\mathbf{x}_{k}\mathbf{x}_{k}^{\top}\right]
\end{pmatrix}}\\
\ge & \Mineigen{\begin{pmatrix}p_{1}K\alpha I_{d} & 0 & 0\\
0 & \ddots & 0\\
0 & 0 & p_{J}K\alpha I_{d}
\end{pmatrix}}\\
\ge & Kp_{\min}\alpha.
\end{align*}
For $t\ge8d\alpha^{-1}p_{\min}^{-1}\log JdT$ ,
\begin{align*}
\sum_{s=1}^{t+1}\Expectation\left[\mathbf{X}_{s}\right]\succeq\sum_{s=1}^{t+1}\Mineigen{\Expectation\left[\mathbf{X}_{s}\right]}I_{J\cdot d}\succeq\left(t+1\right)Kp_{\min}\alpha I_{J\cdot d}\succeq4dK\left(\log\frac{Jd}{\delta}\right)
\end{align*}
Plugging in~\eqref{eq:util5} proves the first inequality of~\eqref{eq:util_claim},
\[
\frac{1}{t+1}\sum_{s=1}^{t+1}\mathbf{X}_{s}\preceq\frac{2}{\left(t+1\right)}\sum_{s=1}^{t+1}\Expectation\left[\mathbf{X}_{s}\right]=2\Expectation\left[\mathbf{X}_{1}\right],
\]
where the equality holds because $\Expectation{\mathbf{X}_{s}} = \Expectation{\mathbf{X}_{1}}$ for all $s\in[T]$.
To prove the second inequality,
\[
\Expectation\left[\mathbf{X}_{1}\right]=\abs{\Psi_{n_{t}}}^{-1}\sum_{\nu\in\Psi_{n_{t}}}\Expectation\left[\mathbf{X}_{\AdmitRound{\nu}}\right],
\]
and by Lemma~\ref{lem:matrix_neg}, with probability at least $1-T^{-1}$,
\[
\frac{1}{2Kd}\sum_{\nu\in\Psi_{n_{t}}}\mathbf{X}_{\AdmitRound{\nu}}\succeq\frac{1}{4Kd}\sum_{\nu\in\Psi_{n_{t}}}\Expectation\left[\mathbf{X}_{\AdmitRound{\nu}}\right]-\left(\log JdT\right)I_{J\cdot d}.
\]
Rearranging the terms,
\begin{equation}
\sum_{\nu\in\Psi_{n_{t}}}\Expectation\left[\mathbf{X}_{\AdmitRound{\nu}}\right]\preceq2\sum_{\nu\in\Psi_{n_{t}}}\mathbf{X}_{\AdmitRound{\nu}}+4Kd\left(\log JdT\right)I_{J\cdot d}
\label{eq:util7}
\end{equation}
By definition of $F_{n_t}$,
\begin{equation}
\begin{split}
\sum_{\nu\in\Psi_{n_{t}}}\mathbf{X}_{\AdmitRound{\nu}}=&F_{n_{t}}-\sum_{\nu\notin\Psi_{n_{t}}}\mathbf{X}_{\AdmitRound{\nu}}-16d(K-1)\log\frac{Jd}{\delta}I_{J\cdot d}\\\succeq&F_{n_{t}}-\left(Kd\abs{\Psi_{n_{t}}^{c}}+16d(K-1)\log\frac{Jd}{\delta}\right)I_{J\cdot d}.
\end{split}
\label{eq:util8}
\end{equation}
Because the condition~\eqref{eq:self_condition} holds, we can use~\eqref{eq:self10},
\[
\left(\frac{3dK}{2}\abs{\Psi_{n_{t}}^{c}}+16Kd\log\frac{Jd}{\delta}\right)\Maxeigen{F_{n_{t}}^{-1}}\le\frac{1}{8},
\]
to obtain
\begin{align*}
Kd\abs{\Psi_{n_{t}}^{c}}+16d(K-1)\log\frac{Jd}{\delta}\le&\frac{2}{3}\left(\frac{3dK}{2}\abs{\Psi_{n_{t}}^{c}}+16Kd\log\frac{Jd}{\delta}\right)+\frac{16}{3}Kd\log\frac{Jd}{\delta}
\\
\le&\frac{1}{12\Maxeigen{F_{n_{t}}}}+\frac{16}{3}Kd\log\frac{Jd}{\delta}
\\
=&\frac{\Mineigen{F_{n_{t}}}}{12}+\frac{16}{3}Kd\log\frac{Jd}{\delta}.
\end{align*}
Plugging in~\eqref{eq:util8},
\begin{equation}
\begin{split}
\sum_{\nu\in\Psi_{n_{t}}}\mathbf{X}_{\AdmitRound{\nu}}\succeq
&F_{n_{t}}-\left\{ \frac{1}{12}\Mineigen{F_{n_{t}}}+\frac{16}{3}Kd\log\frac{Jd}{\delta}\right\} I_{J\cdot d}
\\
\succeq&\left\{ \frac{11}{12}\Mineigen{F_{n_{t}}}-\frac{16}{3}Kd\log\frac{Jd}{\delta}\right\} I_{J\cdot d}
\\
\succeq&\left\{ \frac{44}{3}\left(K-1\right)d\log\frac{Jd}{\delta}-\frac{16}{3}Kd\log\frac{Jd}{\delta}\right\} I_{J\cdot d}
\\
\succeq&\left\{ \frac{22}{3}Kd\log\frac{Jd}{\delta}-\frac{16}{3}Kd\log\frac{Jd}{\delta}\right\} I_{J\cdot d}
\\
\succeq&2Kd\log\frac{Jd}{\delta}I_{J\cdot d}
\\
\succeq&2Kd\log JdTI_{J\cdot d}
\end{split}
\label{eq:util10}
\end{equation}
where the third inequality holds by $F_{n_t} \succeq 16(K-1)\log \frac{Jd}{\delta}$ and the last inequality holds by $\delta < T^{-1}$.
Plugging in \eqref{eq:util7},
\[
\sum_{\nu\in\Psi_{n_{t}}}\Expectation\left[\mathbf{X}_{\AdmitRound{\nu}}\right]\preceq2\sum_{\nu\in\Psi_{n_{t}}}\mathbf{X}_{\AdmitRound{\nu}}+4Kd\left(\log JdT\right)I_{J\cdot d}\preceq 4\sum_{\nu\in\Psi_{n_{t}}}\mathbf{X}_{\AdmitRound{\nu}},
\]
proves the second inequality in claim~\eqref{eq:util_claim}.

From~\eqref{eq:util4},
\begin{equation}
\begin{split}
&\sum_{j=1}^{J}\max_{k\in[K]}p_{j}\frac{1}{\left\{ \sum_{s=1}^{t+1}\Indicator{\Class s=j}\right\} ^{2}}\abs{\sum_{s=1}^{t+1}\Indicator{\Class s=j}\left(\Estimator jt-\Parameter j\right)^{\top}\Context jks}^{2}
\\
&\le\frac{4}{\left(t+1\right)}\left(\StackedEstimator t-\Theta^{\star}\right)^{\top}\left\{ \sum_{s=1}^{t+1}\sum_{k=1}^{K}\StackedContext ks\StackedContext ks^{\top}\right\} \left(\StackedEstimator t-\Theta^{\star}\right)
\\
&\le\frac{32}{\abs{\Psi_{n_{t}}}}\left(\StackedEstimator t-\Theta^{\star}\right)^{\top}\left\{ \sum_{s\in\Psi_{n_{t}}}\sum_{k=1}^{K}\StackedContext ks\StackedContext ks^{\top}\right\} \left(\StackedEstimator t-\Theta^{\star}\right)
\\
&\le\frac{32}{\abs{\Psi_{n_{t}}}}\left(\StackedEstimator t-\Theta^{\star}\right)^{\top}\left\{ V_{n_{t}}\right\} \left(\StackedEstimator t-\Theta^{\star}\right)
\\
&=\frac{32}{\abs{\Psi_{n_{t}}}}\norm{\StackedEstimator t-\Theta^{\star}}_{V_{n_{t}}}^{2}
\end{split}
\label{eq:util11}
\end{equation}
On bounding the normalizing matrix, the novel Gram matrix $V_{n_t}$ plays a crucial role.
To obtain an upper bound for~\eqref{eq:util11}, we need a matrix whose eigenvalue is greater than that of:
\begin{equation}
\sum_{\nu\in\Psi_{n_{t}}}\mathbf{X}_{\AdmitRound{\nu}}=\sum_{\nu\in\Psi_{n_{t}}}\sum_{k=1}^{K}\StackedContext k{\AdmitRound{\nu}}\StackedContext k{\AdmitRound{\nu}}^{\top},
\label{eq:util12}
\end{equation}
However, with $\sum_{\nu\in\Psi_{n_{t}}}\StackedContext{\Action{\AdmitRound{\nu}}}{\AdmitRound{\nu}}\StackedContext{\Action{\AdmitRound{\nu}}}{\AdmitRound{\nu}}^{\top}$
, a Gram matrix consist of only selected contexts, we cannot bound the matrix \eqref{eq:util12}. 
Instead, by using a Gram matrix $V_{t}$, we can bound \eqref{eq:util12} as,
\begin{align*}
\sum_{\nu\in\Psi_{n_{t}}}\mathbf{X}_{\AdmitRound{\nu}}= & \sum_{\nu\in\Psi_{n_{t}}}\sum_{k=1}^{K}\StackedContext k{\AdmitRound{\nu}}\StackedContext k{\AdmitRound{\nu}}^{\top}\\
\preceq & \sum_{\nu\in\Psi_{n_{t}}}\sum_{k=1}^{K}\StackedContext k{\AdmitRound{\nu}}\StackedContext k{\AdmitRound{\nu}}^{\top}+\sum_{\nu\notin\Psi_{n_{t}}}\StackedContext{\Action{\AdmitRound{\nu}}}{\AdmitRound{\nu}}\StackedContext{\Action{\AdmitRound{\nu}}}{\AdmitRound{\nu}}^{\top}\\
\preceq & V_{n_{t}},
\end{align*}
and prove the bound~\eqref{eq:util11} to relate the prediction error to the self-normalized bound.
From \eqref{eq:util11}, by Theorem~\ref{thm:self}
\begin{align*}
\sum_{j=1}^{J}\max_{k\in[K]}p_{j}\frac{1}{\left\{ \sum_{s=1}^{t+1}\Indicator{\Class s=j}\right\} ^{2}}\abs{\sum_{s=1}^{t+1}\Indicator{\Class s=j}\left(\Estimator jt-\Parameter j\right)^{\top}\Context jks}^{2}\le&\frac{32}{\abs{\Psi_{n_{t}}}}\norm{\StackedEstimator t-\Theta^{\star}}_{V_{n_{t}}}^{2}
\\
\le&\frac{32}{\abs{\Psi_{n_{t}}}}\Selfbound{\sigma_{r}}{\delta}^{2},
\end{align*}
with probability at least $1-4(m+1)\delta$.
Because $\abs{\Psi_{n_t}} + \abs{\Psi_{n_t}^{c}} = n_t $ and~\eqref{eq:self_condition} implies
\[
\frac{3Kd}{2}\abs{\Psi_{n_{t}}^{c}}\Maxeigen{F_{n_{t}}^{-1}}\le\left(\frac{3Kd}{2}\abs{\Psi_{n_{t}}^{c}}+16Kd\log\frac{Jd}{\delta}\right)\Maxeigen{F_{n_{t}}^{-1}}\le\frac{1}{8},
\]
we obtain
\[
\abs{\Psi_{n_{t}}}\ge n_{t}-\abs{\Psi_{n_{t}}^{c}}\ge n_{t}-\frac{\Mineigen{F_{n_{t}}}}{12Kd}\ge n_{t}-\frac{\Maxeigen{F_{n_{t}}}}{12Kd}\ge n_{t}-\frac{n_{t}Kd}{12Kd}=\frac{11}{12}n_{t}.
\]
Thus,
\begin{align*}
\sum_{j=1}^{J}\max_{k\in[K]}p_{j}\frac{1}{\left\{ \sum_{s=1}^{t+1}\Indicator{\Class s=j}\right\} ^{2}}\abs{\sum_{s=1}^{t+1}\Indicator{\Class s=j}\left(\Estimator jt-\Parameter j\right)^{\top}\Context jks}^{2}\le&\frac{32}{\abs{\Psi_{n_{t}}}}\Selfbound{\sigma_{r}}{\delta}^{2}
\\
\le&\frac{12}{11}\cdot\frac{32}{n_{t}}\Selfbound{\sigma_{r}}{\delta}^{2}
\\
\le&\frac{36}{n_{t}}\Selfbound{\sigma_{r}}{\delta}^{2}.
\end{align*}
From \eqref{eq:util3},
\[
\sqrt{\sum_{j=1}^{J}p_{j}\max_{k\in[K]}\abs{\OptUtility jk-\tilde{u}_{k,t+1}^{(j)}}^{2}}\le\frac{16\sqrt{J\log JKT}}{\sqrt{t}}+\frac{6\Selfbound{\sigma_{r}}{\delta}}{\sqrt{n_{t}}}
\]
and the proof is completed.
\end{proof}

\subsection{Proof of Lemma \ref{lem:bandit_policy}}

\begin{proof}
Suppose a feasible policy $\TildePolicy jkt$ for the optimization
problem \eqref{eq:oracle_problem} satisfies
\[
\sum_{k=1}^{K+1}\TildePolicy{\Class t}kt\TildeUtility{\Class t}kt>\sum_{k=1}^{K+1}\HatPolicy{\Class t}kt\TildeUtility{\Class t}kt,
\]
which is equivalent to
\begin{equation}
\sum_{l=1}^{K+1}\TildePolicy{\Class t}{k\Order l}t\TildeUtility{\Class t}{k\Order l}t>\sum_{l=1}^{K+1}\HatPolicy{\Class t}{k\Order l}t\TildeUtility{\Class t}{k\Order l}t.\label{eq:optimal_1}
\end{equation}
Without loss of generality we assume $\HatUtility{\Class t}{k\Order l}t\ge0$
(Because $\sum_{l=1}^{K+1}\TildePolicy{\Class t}{k\Order l}t=\sum_{l=1}^{K+1}\HatPolicy{\Class t}{k\Order l}t=1$,
we can subtract $\HatUtility{\Class t}{k\Order{K+1}}t$ on both side
of \eqref{eq:optimal_1}). By the constraints on the resources, 
\[
\TildePolicy{\Class t}{k\Order 1}t\le\left(\min_{r\in[m]}\frac{\rho_{t}(r)}{\Consumption{\Class t}{k\Order 1}t(r)}\right)\wedge1=\HatPolicy{\Class t}{k\Order 1}t
\]
Suppose $\TildePolicy{\Class t}{k\Order 1}t<\HatPolicy{\Class t}{k\Order 1}t$.
Because $\sum_{l=1}^{K+1}\TildePolicy{\Class t}{k\Order l}t=\sum_{l=1}^{K+1}\HatPolicy{\Class t}{k\Order l}t=1$, by Lemma \ref{lem:ordering},
\[
\sum_{l=1}^{K+1}\TildePolicy{\Class t}{k\Order l}t\TildeUtility{\Class t}{k\Order l}t\le\sum_{l=1}^{K+1}\HatPolicy{\Class t}{k\Order l}t\TildeUtility{\Class t}{k\Order l}t,
\]
which contradicts with \eqref{eq:optimal_1}. Thus we have $\TildePolicy{\Class t}{k\Order 1}t=\HatPolicy{\Class t}{k\Order 1}t$
and 
\begin{equation}
\sum_{l=2}^{K+1}\TildePolicy{\Class t}{k\Order l}t\TildeUtility{\Class t}{k\Order l}t>\sum_{l=2}^{K+1}\HatPolicy{\Class t}{k\Order l}t\TildeUtility{\Class t}{k\Order l}t.\label{eq:optimal_2}
\end{equation}
Again, by the constraints on the resources,$\TildePolicy{\Class t}{k\Order 2}t\le\HatPolicy{\Class t}{k\Order 2}t$.
Suppose $\TildePolicy{\Class t}{k\Order 2}t<\HatPolicy{\Class t}{k\Order 2}t$.
Because $\sum_{l=2}^{K+1}\TildePolicy{\Class t}{k\Order l}t=\sum_{l=2}^{K+1}\HatPolicy{\Class t}{k\Order l}t$,
by Lemma \ref{lem:ordering},
\[
\sum_{l=2}^{K+1}\TildePolicy{\Class t}{k\Order l}t\TildeUtility{\Class t}{k\Order l}t\le\sum_{l=2}^{K+1}\HatPolicy{\Class t}{k\Order l}t\TildeUtility{\Class t}{k\Order l}t.
\]
which contradicts with \eqref{eq:optimal_2}. Thus we have $\TildePolicy{\Class t}{k\Order 2}t=\HatPolicy{\Class t}{k\Order 2}t$
Recursively, we have $\TildePolicy{\Class t}{k\Order l}t=\HatPolicy{\Class t}{k\Order l}t$
for all $l\in[K+1]$. Thus there exist no feasible solution $\TildePolicy jkt$
such that \eqref{eq:optimal_1} holds and the proof is completed. 
\end{proof}

\subsection{Proof of Lemma \ref{lem:reward_lower_bound}}
\begin{proof}
For each $t\in[T]$, denote the good events $\mathcal{G}_t:=\mathcal{E}_t\cap\mathcal{M}_{t-1}$.

\paragraph{Step 1. Bounds for the estimates $\TildeUtility{\Class t}kt$ and $\tilde{\mathbf{b}}_{k,t}^{(\Class{t})}$ :} 
For each $t\in[T]$ and $k\in[K]$,
\begin{align*}
\TildeUtility{\Class t}kt=\TildeUtility{\Class t}kt-\OptUtility{\Class t}k+\OptUtility{\Class t}k=&\frac{\Predbound{t-1}{\sigma_{r}}{\delta}}{\sqrt{p_{\Class t}}}+\HatUtility{\Class t}kt-\OptUtility{\Class t}k+\OptUtility{\Class t}k\\
\ge&\frac{\Predbound{t-1}{\sigma_{r}}{\delta}-\sqrt{p_{\Class t}}\max_{k\in[K]}\abs{\HatUtility{\Class t}kt-\OptUtility{\Class t}k}}{\sqrt{p_{\Class t}}}+\OptUtility{\Class t}k.
\end{align*}
Under the event $\mathcal{G}_t$,
\begin{align*}
\sqrt{p_{\Class t}}\max_{k\in[K]}\abs{\OptUtility{\Class t}k-\HatUtility{\Class t}kt}=&\sqrt{p_{\Class t}\max_{k\in[K]}\abs{\OptUtility{\Class t}k-\HatUtility{\Class t}kt}^{2}}\\\le&\sqrt{\sum_{j=1}^{J}p_{j}\max_{k\in[K]}\abs{\OptUtility jk-\HatUtility jkt}^{2}}\\\le&\Predbound{t-1}{\sigma_{r}}{\delta},
\end{align*}
which implies
\begin{equation}
\TildeUtility{\Class t}kt\ge\OptUtility{\Class t}k.
\label{eq:u_tilde_lower}
\end{equation}
Similarly,
\begin{equation}
\tilde{\mathbf{b}}_{k,t}^{(\Class t)}\le\mathbf{b}_{k}^{\star(\Class t)}.
\label{eq:b_tilde_upper}
\end{equation}
Another useful bound for $\TildeUtility{\Class t}kt$ is
\begin{equation}
\Expectation\left[\sum_{t\in\mathcal{U}}\sum_{k=1}^{K}\HatPolicy{\Class t}kt\abs{\TildeUtility{\Class t}kt-\OptUtility{\Class t}k}\Indicator{\mathcal{G}_{t}}\right]\le2\Predbound{t-1}{\sigma_{r}}{\delta}\sqrt{\Expectation\left[\Indicator{\Action t\in[K]}\right]}.
\label{eq:u_tilde_l2_bound}
\end{equation}
This bound is proved by the tower property of conditional expectation and Cauchy-Schwartz inequality,
\begin{align*}
\Expectation\left[\sum_{k=1}^{K}\HatPolicy{\Class t}kt\abs{\TildeUtility{\Class t}kt-\OptUtility{\Class t}k}\Indicator{\mathcal{G}_{t}}\right]=&\Expectation\left[\max_{k\in[K]}\abs{\TildeUtility{\Class t}kt-\OptUtility{\Class t}k}\sum_{k=1}^{K}\HatPolicy{\Class t}kt\Indicator{\mathcal{G}_{t}}\right]\\=&\Expectation\left[\max_{k\in[K]}\abs{\TildeUtility{\Class t}kt-\OptUtility{\Class t}k}\Indicator{\Action t\in[K]}\Indicator{\mathcal{G}_{t}}\right]\\=&\Expectation\left[\sum_{j=1}^{J}p_{j}\max_{k\in[K]}\abs{\TildeUtility jkt-\OptUtility jk}\Indicator{\Action t\in[K]}\Indicator{\mathcal{G}_{t}}\right]\\\le&\Expectation\left[\sqrt{\sum_{j=1}^{J}p_{j}\max_{k\in[K]}\abs{\TildeUtility jkt-\OptUtility jk}^{2}}\sqrt{\sum_{j=1}^{J}p_{j}\Indicator{\Action t\in[K]}}\Indicator{\mathcal{G}_{t}}\right]
\end{align*}
By definition of $\TildeUtility jkt$ and triangular inequality for $\ell_2$-norm,
\begin{align*}
\sqrt{\sum_{j=1}^{J}p_{j}\max_{k\in[K]}\abs{\TildeUtility jkt-\OptUtility jk}^{2}}\Indicator{\mathcal{G}_{t}}=&\sqrt{\sum_{j=1}^{J}p_{j}\max_{k\in[K]}\abs{\HatUtility jkt-\OptUtility jk+\frac{\Predbound{t-1}{\sigma_{r}}{\delta}}{\sqrt{p_{j}}}}^{2}}\Indicator{\mathcal{G}_{t}}\\\le&\left(\sqrt{\sum_{j=1}^{J}p_{j}\max_{k\in[K]}\abs{\HatUtility jkt-\OptUtility jk}^{2}}+\sqrt{\sum_{j=1}^{J}p_{j}\left(\frac{\Predbound{t-1}{\sigma_{r}}{\delta}}{\sqrt{p_{j}}}\right)^{2}}\right)\Indicator{\mathcal{G}_{t}}\\\le&2\Predbound{t-1}{\sigma_{r}}{\delta}\Indicator{\mathcal{G}_{t}}\\\le&2\Predbound{t-1}{\sigma_{r}}{\delta}.
\end{align*}
Then by Jensen's inequality,
\begin{align*}
\Expectation\left[\sum_{k=1}^{K}\HatPolicy{\Class t}kt\abs{\TildeUtility{\Class t}kt-\OptUtility{\Class t}k}\Indicator{\mathcal{G}_{t}}\right]\le&\Expectation\left[\sqrt{\sum_{j=1}^{J}p_{j}\max_{k\in[K]}\abs{\TildeUtility jkt-\OptUtility jk}^{2}}\sqrt{\sum_{j=1}^{J}p_{j}\Indicator{\Action t\in[K]}}\Indicator{\mathcal{G}_{t}}\right]\\\le&2\Predbound{t-1}{\sigma_{r}}{\delta}\Expectation\left[\sqrt{\sum_{j=1}^{J}p_{j}\Indicator{\Action t\in[K]}}\right]\\\le&2\Predbound{t-1}{\sigma_{r}}{\delta}\sqrt{\Expectation\left[\sum_{j=1}^{J}p_{j}\Indicator{\Action t\in[K]}\right]}\\=&2\Predbound{t-1}{\sigma_{r}}{\delta}\sqrt{\Expectation\left[\Indicator{\Action t\in[K]}\right]},
\end{align*}
which proves~\eqref{eq:u_tilde_l2_bound}.
Similarly,
\begin{equation}
\Expectation\left[\sum_{k=1}^{K}\HatPolicy{\Class t}kt\norm{\Contilde{\Class t}kt-\Optcon{\Class t}k}_{\infty}\Indicator{\mathcal{G}_{t}}\right]\le2\Predbound{t-1}{\sigma_{b}}{\delta}\sqrt{\Expectation\left[\Indicator{\Action t\in[K]}\right]}
\label{eq:b_tilde_l2_bound}
\end{equation}

\paragraph{Step 2. Reward decomposition:}
Let $\tau$ be the stopping time of the algorithm and let $\mathcal{U}:=\left\{ t\in[\tau]:\rho_{t}>0\right\} $.
Then for $t\notin\mathcal{U}$, the allocated resource is $\rho_{t}\vee0=0$
and the algorithm skips the round. Thus,
\begin{align*}
\Expectation\left[\sum_{t=1}^{T}R_{t}^{\widehat{\pi}}\right]= & \Expectation\left[\sum_{t\in\mathcal{U}}R_{t}^{\widehat{\pi}}\right].
\end{align*}
Then, the reward is decomposed as
\begin{align*}
\Expectation\left[\sum_{t\in\mathcal{U}}R_{t}^{\widehat{\pi}}\right]=&\Expectation\left[\sum_{t\in\mathcal{U}}R_{t}^{\widehat{\pi}}\Indicator{\mathcal{G}_{t}}\right]+\Expectation\left[\sum_{t\in\mathcal{U}}R_{t}^{\widehat{\pi}}\Indicator{\mathcal{G}_{t}^{c}}\right]\\\ge&\Expectation\left[\sum_{t\in\mathcal{U}}\sum_{k=1}^{K}\HatPolicy{\Class t}kt\OptUtility{\Class t}k\Indicator{\mathcal{G}_{t}}\right]-\sum_{t=1}^{T}\Probability\left(\mathcal{G}_{t}^{c}\right)\\\ge&\Expectation\left[\sum_{t\in\mathcal{U}}\sum_{k=1}^{K}\HatPolicy{\Class t}kt\TildeUtility{\Class t}kt\Indicator{\mathcal{G}_{t}}\right]-\Expectation\left[\sum_{t\in\mathcal{U}}\sum_{k=1}^{K}\HatPolicy{\Class t}kt\abs{\TildeUtility{\Class t}kt-\OptUtility{\Class t}k}\Indicator{\mathcal{G}_{t}}\right]-\sum_{t=1}^{T}\Probability\left(\mathcal{G}_{t}^{c}\right)\\\ge&\Expectation\left[\sum_{t\in\mathcal{U}}\sum_{k=1}^{K}\HatPolicy{\Class t}kt\TildeUtility{\Class t}kt\Indicator{\mathcal{G}_{t}}\right]-\sum_{t=1}^{T}\Expectation\left[\sum_{k=1}^{K}\HatPolicy{\Class t}kt\abs{\TildeUtility{\Class t}kt-\OptUtility{\Class t}k}\Indicator{\mathcal{G}_{t}}\right]-\sum_{t=1}^{T}\Probability\left(\mathcal{G}_{t}^{c}\right).
\end{align*}
By the bound~\eqref{eq:u_tilde_l2_bound},
\begin{align*}
\sum_{t=1}^{T}\Expectation\left[\sum_{k=1}^{K}\HatPolicy{\Class t}kt\abs{\TildeUtility{\Class t}kt-\OptUtility{\Class t}k}\Indicator{\mathcal{G}_{t}}\right]\le&2\sum_{t=1}^{T}\Predbound{t-1}{\sigma_{r}}{\delta}\sqrt{\Expectation\left[\Indicator{\Action t\in[K]}\right]}\\
\le&2\sqrt{T\sum_{t=1}^{T}\Predbound{t-1}{\sigma_{r}}{\delta}^{2}\Expectation\left[\Indicator{\Action t\in[K]}\right]}\\
=&2\sqrt{T\Expectation\left[\sum_{t=1}^{T}\Predbound{t-1}{\sigma_{r}}{\delta}^{2}\Indicator{\Action t\in[K]}\right]}
\end{align*}
where the last ineqaulity holds by Cauchy-Schwartz inequality.
Thus, the reward is decomposed as
\begin{equation}
\begin{split}
\Expectation\left[\sum_{t=1}^{T}R_{t}^{\widehat{\pi}}\right]=&\Expectation\left[\sum_{t\in\mathcal{U}}R_{t}^{\widehat{\pi}}\right]\\\ge&\Expectation\left[\sum_{t\in\mathcal{U}}\sum_{k=1}^{K}\HatPolicy{\Class t}kt\TildeUtility{\Class t}kt\Indicator{\mathcal{G}_{t}}\right]-2\sqrt{T\Expectation\left[\sum_{t=1}^{T}\Predbound{t-1}{\sigma_{r}}{\delta}^{2}\Indicator{\Action t\in[K]}\right]}-\sum_{t=1}^{T}\Probability\left(\mathcal{G}_{t}^{c}\right)
\end{split}
\label{eq:regret_1}
\end{equation}

\paragraph{Step 3. A lower bound for $\rho_{t}$:} Denote $u_{1}<u_{2}<\ldots<u_{\abs{\mathcal{U}}}$
the indexes in $\mathcal{U}$. For $s\notin\mathcal{U}$, we have
$\rho_{s}=\mathbf{0}_{m}$ and $\Consumption{\Class s}{\Action s}s=\mathbf{0}_{m}$.
Thus for $\nu\in[|\mathcal{U}|-1]$, 
\begin{equation}
\rho_{u_{\nu+1}}=u_{\nu+1}\rho-\sum_{s=1}^{u_{\nu+1}-1}\Consumption{\Class s}{\Action s}s=u_{\nu+1}\rho-\sum_{s=1}^{u_{\nu}}\Consumption{\Class s}{\Action s}s.\label{eq:regret_2}
\end{equation}
By the resource constrain at round $u_{\nu}$,
\begin{align*}
\sum_{k=1}^{K}\HatPolicy{\Class{u_{\nu}}}k{u_{\nu}}\Contilde{\Class{u_{\nu}}}k{u_{\nu}}\le & u_{\nu}\rho-\sum_{s=1}^{u_{\nu}-1}\Consumption{\Class s}{\Action s}s\\
= & u_{\nu}\rho+\Consumption{\Class{u_{\nu}}}{\Action{u_{\nu}}}{u_{\nu}}-\sum_{s=1}^{u_{\nu}}\Consumption{\Class s}{\Action s}s.
\end{align*}
Plugging in \eqref{eq:regret_2},
\begin{align*}
\rho_{u_{\nu+1}}\ge & \left(u_{\nu+1}-u_{\nu}\right)\rho-\Consumption{\Class{u_{\nu}}}{\Action{u_{\nu}}}{u_{\nu}}+\sum_{k=1}^{K}\HatPolicy{\Class{u_{\nu}}}k{u_{\nu}}\Contilde{\Class{u_{\nu}}}k{u_{\nu}}\\
\ge & \left(u_{\nu+1}-u_{\nu}\right)\rho-\Consumption{\Class{u_{\nu}}}{\Action{u_{\nu}}}{u_{\nu}}+\sum_{k=1}^{K}\HatPolicy{\Class{u_{\nu}}}k{u_{\nu}}\Optcon{\Class{u_{\nu}}}k+\sum_{k=1}^{K}\HatPolicy{\Class{u_{\nu}}}k{u_{\nu}}\left(\Contilde{\Class{u_{\nu}}}k{u_{\nu}}-\Optcon{\Class{u_{\nu}}}k\right).
\end{align*}
Taking conditional expectation on both sides gives
\[
\begin{split}\CE{\rho_{u_{\nu+1}}}{\Class{u_{\nu+1}}}\ge & \CE{u_{\nu+1}-u_{\nu}}{\Class{u_{\nu+1}}}\rho+\CE{-\Consumption{\Class{u_{\nu}}}{\Action{u_{\nu}}}{u_{\nu}}+\sum_{k=1}^{K}\HatPolicy{\Class{u_{\nu}}}k{u_{\nu}}\Optcon{\Class{u_{\nu}}}k}{\Class{u_{\nu+1}}}\\
 & +\CE{\sum_{k=1}^{K}\HatPolicy{\Class{u_{\nu}}}k{u_{\nu}}\left(\Contilde{\Class{u_{\nu}}}k{u_{\nu}}-\Optcon{\Class{u_{\nu}}}k\right)}{\Class{u_{\nu+1}}}\\
= & \CE{u_{\nu+1}-u_{\nu}}{\Class{u_{\nu+1}}}\rho+\Expectation\left[\sum_{k=1}^{K}\HatPolicy{\Class{u_{\nu}}}k{u_{\nu}}\left(\Contilde{\Class{u_{\nu}}}k{u_{\nu}}-\Optcon{\Class{u_{\nu}}}k\right)\right]\\
\ge & \CE{u_{\nu+1}-u_{\nu}}{\Class{u_{\nu+1}}}\rho+\Expectation\left[\sum_{k=1}^{K}\HatPolicy{\Class{u_{\nu}}}k{u_{\nu}}\left(\Contilde{\Class{u_{\nu}}}k{u_{\nu}}-\Optcon{\Class{u_{\nu}}}k\right)\Indicator{\mathcal{G}_{u_{\nu}}}\right]-\Probability\left(\mathcal{G}_{u_{\nu}}^{c}\right)\mathbf{1}_{m},
\end{split}
\]
where the equality holds by Assumption~\ref{assum:error} and
\begin{align*}
\Expectation\left[\left\{ -\Consumption{\Class{u_{\nu}}}{\Action{u_{\nu}}}{u_{\nu}}+\sum_{k=1}^{K}\HatPolicy{\Class{u_{\nu}}}k{u_{\nu}}\Optcon{\Class{u_{\nu}}}k\right\} \right]= & \Expectation\left[\left\{ -\Optcon{\Class{u_{\nu}}}{\Action{u_{\nu}}}+\sum_{k=1}^{K}\HatPolicy{\Class{u_{\nu}}}k{u_{\nu}}\Optcon{\Class{u_{\nu}}}k\right\} \right]\\
= & \Expectation\left[\left\{ -\sum_{k=1}^{K}\HatPolicy{\Class{u_{\nu}}}k{u_{\nu}}\Optcon{\Class{u_{\nu}}}k+\sum_{k=1}^{K}\HatPolicy{\Class{u_{\nu}}}k{u_{\nu}}\Optcon{\Class{u_{\nu}}}k\right\} \right]\\
= & 0.
\end{align*}
For the last term, by the bound~\eqref{eq:b_tilde_l2_bound},
\begin{align*}
\Expectation\left[\sum_{k=1}^{K}\HatPolicy{\Class{u_{\nu}}}k{u_{\nu}}\left(\Contilde{\Class{u_{\nu}}}k{u_{\nu}}-\Optcon{\Class{u_{\nu}}}k\right)\Indicator{\mathcal{G}_{u_{\nu}}}\right]\ge&-\Expectation\left[\sum_{k=1}^{K}\HatPolicy{\Class{u_{\nu}}}k{u_{\nu}}\norm{\Contilde{\Class{u_{\nu}}}k{u_{\nu}}-\Optcon{\Class{u_{\nu}}}k}_{\infty}\Indicator{\mathcal{G}_{u_{\nu}}}\right]\mathbf{1}_{m}\\\ge&-\Expectation\left[2\Predbound{u_{\nu}-1}{\sigma_{b}}{\delta}\sqrt{\CE{\Indicator{\Action{u_{\nu}}\in[K]}}{u_{\nu}}}\right]\mathbf{1}_{m}.
\end{align*}
Thus we obtain a lower bound,
\begin{equation}
\begin{split}
\CE{\rho_{u_{\nu+1}}}{\Class{u_{\nu+1}}}\ge&\CE{u_{\nu+1}-u_{\nu}}{\Class{u_{\nu+1}}}\rho-\Probability\left(\mathcal{G}_{u_{\nu}}^{c}\right)\mathbf{1}_{m}\\&-2\Expectation\left[\Predbound{u_{\nu}-1}{\sigma_{b}}{\delta}\sqrt{\CE{\Indicator{\Action{u_{\nu}}\in[K]}}{u_{\nu}}}\right]\mathbf{1}_{m}.
\end{split}
\label{eq:regret_3}
\end{equation}
\[
\]

\paragraph{Step 4. An upper bound for $OPT$} 
In the optimization problem \eqref{eq:oracle_problem},
all constraints are linear with respect to the variable and there
exist a feasible solution. Thus the problem satisfies the Slater's
condition and strong duality \citep{boyd2004convex}. Then,
\[
\frac{OPT}{T}=\max_{\pi_{k}^{(j)}}\min_{\lambda\in\Real_{+}^{m}}\min_{\mu^{(j)}\ge0}\min_{\nu_{k}^{(j)}\ge0}L\left(\pi_{k}^{(j)},\lambda,\mu^{(j)},\nu_{k}^{(j)}\right),
\]
where $L$ is the Lagrangian function:
\begin{align*}
L\left(\pi_{k}^{(j)},\lambda,\mu^{(j)},\nu_{k}^{(j)}\right):= & \sum_{j=1}^{J}\sum_{k=1}^{K}p_{j}\pi_{k}^{(j)}\OptUtility jk+\left(\rho-\sum_{j=1}^{J}\sum_{k=1}^{K}p_{j}\pi_{k}^{(j)}\mathbf{b}_{k}^{\star(j)}\right)^{\top}\lambda\\
 & +\sum_{j=1}^{J}\mu^{(j)}\left(1-\sum_{k=1}^{K}\Policy jk1\right)+\sum_{j=1}^{J}\sum_{k=1}^{K}\nu_{k}^{(j)}\Policy jk1.
\end{align*}
Minimizing over $\mu^{(j)}$ and $\nu_{k}^{(j)}$ gives
\begin{align*}
&\min_{\mu_{t}^{(j)}\ge0}\min_{\nu_{k,t}^{(j)}\ge0}L\left(\pi_{k}^{(j)},\lambda,\mu^{(j)},\nu_{k}^{(j)}\right)\\&=\begin{cases}
\sum_{j=1}^{J}\sum_{k=1}^{K}p_{j}\pi_{k}^{(j)}\OptUtility jk+\left(\rho-\sum_{j=1}^{J}\sum_{k=1}^{K}p_{j}\pi_{k}^{(j)}\mathbf{b}_{k}^{\star(j)}\right)^{\top}\lambda & \sum_{k=1}^{K}\pi_{k}^{(j)}\le1,\pi_{k}^{(j)}\ge0\\
-\infty & o.w.
\end{cases},
\end{align*}
which implies
\[
\begin{split}\frac{OPT}{T}= & \max_{\pi_{k}^{(j)}}\min_{\lambda\in\Real_{+}^{m}}\min_{\mu_{t}^{(j)}\ge0}\min_{\nu_{k,t}^{(j)}\ge0}L\left(\pi_{k}^{(j)},\lambda,\mu^{(j)},\nu_{k}^{(j)}\right)\\
\le & \max_{\sum_{k=1}^{K}\pi_{k}^{(j)}\le1,\pi_{k}^{(j)}\ge0}\min_{\lambda\in\Real_{+}^{m}}\sum_{j=1}^{J}\sum_{k=1}^{K}p_{j}\pi_{k}^{(j)}\OptUtility jk+\left(\rho-\sum_{j=1}^{J}\sum_{k=1}^{K}p_{j}\pi_{k}^{(j)}\mathbf{b}_{k}^{\star(j)}\right)^{\top}\lambda\\
= & \max_{\sum_{k=1}^{K}\pi_{k}^{(j)}\le1,\pi_{k}^{(j)}\ge0}\min_{\lambda\in\Real_{+}^{m}}\sum_{j=1}^{J}p_{j}\left\{ \sum_{k=1}^{K}\pi_{k}^{(j)}\OptUtility jk+\left(\rho-\sum_{k=1}^{K}\pi_{k}^{(j)}\mathbf{b}_{k}^{\star(j)}\right)^{\top}\lambda\right\} \\
\le & \min_{\lambda\in\Real_{+}^{m}}\max_{\sum_{k=1}^{K}\pi_{k}^{(j)}\le1,\pi_{k}^{(j)}\ge0}\sum_{j=1}^{J}p_{j}\left\{ \sum_{k=1}^{K}\pi_{k}^{(j)}\OptUtility jk+\left(\rho-\sum_{k=1}^{K}\pi_{k}^{(j)}\mathbf{b}_{k}^{\star(j)}\right)^{\top}\lambda\right\} \\
\le & \min_{\lambda\in\Real_{+}^{m}}\sum_{j=1}^{J}p_{j}\max_{\sum_{k=1}^{K}\pi_{k}^{(j)}\le1,\pi_{k}^{(j)}\ge0}\left\{ \sum_{k=1}^{K}\pi_{k}^{(j)}\OptUtility jk+\left(\rho-\sum_{k=1}^{K}\pi_{k}^{(j)}\mathbf{b}_{k}^{\star(j)}\right)^{\top}\lambda\right\} .
\end{split}
\]
Let $\{\bar{\pi}_{k}^{(j)}:j\in[J],k\in[K]\}$ be the maximizer. If
$\rho-\sum_{k=1}^{K}\bar{\pi}{}_{k}^{(j)}\mathbf{b}_{k}^{\star(j)}$
is negative for some element and $j\in[J]$, then the optimal value
becomes $-\infty$. Thus
\begin{align*}
\frac{OPT}{T}\le & \min_{\lambda\in\Real_{+}^{m}}\sum_{j=1}^{J}p_{j}\max_{\sum_{k=1}^{K}\pi_{k}^{(j)}\le1,\pi_{k}^{(j)}\ge0}\left\{ \sum_{k=1}^{K}\pi_{k}^{(j)}\OptUtility jk+\left(\rho-\sum_{k=1}^{K}\pi_{k}^{(j)}\mathbf{b}_{k}^{\star(j)}\right)^{\top}\lambda\right\} \\
= & \min_{\lambda\in\Real_{+}^{m}}\sum_{j=1}^{J}p_{j}\max_{\sum_{k=1}^{K}\pi_{k}^{(j)}\le1,\pi_{k}^{(j)}\ge0,\rho-\sum_{k=1}^{K}\pi_{k}^{(j)}\mathbf{b}_{k}^{\star(j)}\ge0}\left\{ \sum_{k=1}^{K}\pi_{k}^{(j)}\OptUtility jk+\left(\rho-\sum_{k=1}^{K}\pi_{k}^{(j)}\mathbf{b}_{k}^{\star(j)}\right)^{\top}\lambda\right\} \\
= & \sum_{j=1}^{J}p_{j}\max_{\sum_{k=1}^{K}\pi_{k}^{(j)}\le1,\pi_{k}^{(j)}\ge0,\rho-\sum_{k=1}^{K}\pi_{k}^{(j)}\mathbf{b}_{k}^{\star(j)}\ge0}\left\{ \sum_{k=1}^{K}\pi_{k}^{(j)}\OptUtility jk\right\} .
\end{align*}
 For each $j\in[J]$ and $\mathbf{v}\in\Real_{+}^{m}$, let $\TildePolicy jk{\mathbf{v}}$
be the solution to the optimization problem:
\begin{equation}
\begin{split}\max_{\Policy jk{\mathbf{v}}} & \sum_{k=1}^{K}\Policy jk{\mathbf{v}}\OptUtility jk\\
\text{s.t.} & \sum_{k=1}^{K}\Policy jk{\mathbf{v}}\mathbf{b}_{k}^{\star(j)}\le\mathbf{v}.
\end{split}
\label{eq:regret_problem}
\end{equation}
Then,
\begin{align*}
\frac{OPT}{T}\le & \sum_{j=1}^{J}p_{j}\max_{\sum_{k=1}^{K}\pi_{k}^{(j)}\le1,\pi_{k}^{(j)}\ge0,\rho-\sum_{k=1}^{K}\pi_{k}^{(j)}\mathbf{b}_{k}^{\star(j)}\ge0}\left\{ \sum_{k=1}^{K}\pi_{k}^{(j)}\OptUtility jk\right\} \\
= & \sum_{j=1}^{J}p_{j}\sum_{k=1}^{K}\TildePolicy jk{\rho}\OptUtility jk.
\end{align*}
For each $\nu\in[|\mathcal{U}|-1]$,
\begin{align*}
\Expectation\left[\left(u_{\nu+1}-u_{\nu}\right)\frac{OPT}{T}\right]\le & \Expectation\left[\left(u_{\nu+1}-u_{\nu}\right)\sum_{j=1}^{J}p_{j}\sum_{k=1}^{K}\TildePolicy jk{\rho}\OptUtility jk\right]\\
= & \Expectation\left[\left(u_{\nu+1}-u_{\nu}\right)\sum_{k=1}^{K}\TildePolicy{\Class{u_{\nu+1}}}k{\rho}\OptUtility{\Class{u_{\nu+1}}}k\right]
\end{align*}
In \eqref{eq:regret_problem}, all
constraints are linear with respect to the variable and there exist
a feasible solution. Thus the problem satisfies the Slater's condition
and strong duality \citep{boyd2004convex}. The dual problem of \eqref{eq:regret_problem}
is
\begin{equation}
\begin{split}\min_{\lambda_{\mathbf{v}}^{(j)}\in\Real_{+}^{m}} & \mathbf{v}^{\top}\lambda_{\mathbf{v}}^{(j)}\\
\text{s.t.} & \left(\mathbf{b}_{k}^{\star(j)}\right)^{\top}\lambda_{\mathbf{v}}^{(j)}\ge\OptUtility jk,\quad\forall k\in[K].
\end{split}
\label{eq:regret_dual}
\end{equation}
Let $\tilde{\lambda}_{\mathbf{v}}^{(j)}$ be the solution to \eqref{eq:regret_dual}.
By strong duality, for each $\nu\in[|\mathcal{U}|-1]$, 
\begin{equation}
\begin{split}
&\Expectation\left[\left(u_{\nu+1}-u_{\nu}\right)\sum_{k=1}^{K}\TildePolicy{\Class{u_{\nu+1}}}k{\rho}\OptUtility{\Class{u_{\nu+1}}}k\right]\\&=\Expectation\left[\left(u_{\nu+1}-u_{\nu}\right)\rho^{\top}\tilde{\lambda}_{\rho}^{(\Class{u_{\nu+1}})}\right]\\&=\Expectation\left[\CE{\left(u_{\nu+1}-u_{\nu}\right)\rho}{\Class{u_{\nu+1}}}^{\top}\tilde{\lambda}_{\rho}^{(\Class{u_{\nu+1}})}\right]\\&=\Expectation\left[\left(\Probability\left(\mathcal{G}_{u_{\nu}}^{c}\right)+2\Expectation\left[\sqrt{\CE{\Indicator{\Action{u_{\nu}}\in[K]}}{u_{\nu}}}\Predbound{u_{\nu}-1}{\lambda}{\sigma_{b}}\right]\right)\mathbf{1}_{m}^{\top}\tilde{\lambda}_{\rho}^{(\Class{u_{\nu+1}})}\right]\\&\quad+\Expectation\left[\left(\CE{\left(u_{\nu+1}-u_{\nu}\right)\rho}{\Class{u_{\nu+1}}}-\Probability\left(\mathcal{G}_{u_{\nu}}^{c}\right)-2\Expectation\left[\sqrt{\CE{\Indicator{\Action{u_{\nu}}\in[K]}}{u_{\nu}}}\Predbound{u_{\nu}-1}{\sigma_{b}}{\delta}\right]\right)\mathbf{1}_{m}^{\top}\tilde{\lambda}_{\rho}^{(\Class{u_{\nu+1}})}\right].
\end{split}
\label{eq:regret4_1}
\end{equation}
For the first term, we observe the dual problem of \eqref{eq:oracle_problem},
\begin{equation}
\begin{split}\min_{\lambda\in\Real_{+}^{m}} & \rho\mathbf{1}_{m}^{\top}\lambda\\
\text{s.t.} & \lambda^{\top}\Optcon jk\ge\OptUtility jk,\forall j\in[J],\forall k\in[K].
\end{split}
\label{eq:opt_dual}
\end{equation}
Comparing to the dual problem \eqref{eq:regret_dual}, when $\mathbf{v}=\rho\mathbf{1}_{m}$
and $j=\Class{u_{\nu+1}}$, \eqref{eq:opt_dual} has more constraints
than \eqref{eq:regret_dual} with same objective function. Denote
$\lambda_{\star}$ be the solution to \eqref{eq:opt_dual}. Then,
\[
\rho\mathbf{1}_{m}^{\top}\tilde{\lambda}_{\rho\mathbf{1}_{m}}^{(\Class{u_{\nu+1}})}\le\rho\mathbf{1}_{m}^{\top}\lambda_{\star}=\frac{OPT}{T},
\]
where the last equality holds by strong duality for the oracle problem~\eqref{eq:oracle_problem}. 
Thus the first term in~\eqref{eq:regret4_1} is bounded as
\begin{align*}
&\Expectation\left[\left(\Probability\left(\mathcal{G}_{u_{\nu}}^{c}\right)+2\Expectation\left[\sqrt{\CE{\Indicator{\Action{u_{\nu}}\in[K]}}{u_{\nu}}}\Predbound{u_{\nu}-1}{\lambda}{\sigma_{b}}\right]\right)\mathbf{1}_{m}^{\top}\tilde{\lambda}_{\rho\mathbf{1}_{m}}^{(\Class{u_{\nu+1}})}\right]\\&\le\left(\Probability\left(\mathcal{G}_{u_{\nu}}^{c}\right)+2\Expectation\left[\sqrt{\CE{\Indicator{\Action{u_{\nu}}\in[K]}}{u_{\nu}}}\Predbound{u_{\nu}-1}{\lambda}{\sigma_{b}}\right]\right)\frac{OPT}{\rho T}.
\end{align*}
For the second term in~\eqref{eq:regret4_1}, we observe that $\tilde{\lambda}_{\CE{\rho_{u_{\nu+1}}}{\Class{u_{\nu+1}}}}^{(\Class{u_{\nu}+1})}$
is a feasible solution to \eqref{eq:regret_dual} when $\mathbf{v}=\rho\mathbf{1}_{m}$
and $j=\Class{u_{\nu+1}}$. 
Thus
\begin{align*}
&\Expectation\left[\left(\CE{\left(u_{\nu+1}-u_{\nu}\right)\rho}{\Class{u_{\nu+1}}}\!-\!2\Expectation\left[\sqrt{\CE{\Indicator{\!\Action{u_{\nu}}\!\in\![K]\!}}{u_{\nu}}}\Predbound{u_{\nu}-1}{\lambda}{\sigma_{b}}\right]\!-\!\Probability\left(\mathcal{G}_{u_{\nu}}^{c}\right)\right)\!\mathbf{1}_{m}^{\top}\tilde{\lambda}_{\rho\mathbf{1}_{m}}^{(\Class{u_{\nu+1}})}\right]\\&\le\!\Expectation\left[\left\{ \left(\CE{\left(u_{\nu+1}-u_{\nu}\right)\rho}{\Class{u_{\nu+1}}}\!-\!2\Expectation\left[\sqrt{\CE{\Indicator{\!\Action{u_{\nu}}\!\in\![K]\!}}{u_{\nu}}}\Predbound{u_{\nu}-1}{\lambda}{\sigma_{b}}\right]\!-\!\Probability\left(\mathcal{G}_{u_{\nu}}^{c}\right)\right)\vee0\right\} \!\mathbf{1}_{m}^{\top}\tilde{\lambda}_{\rho\mathbf{1}_{m}}^{(\Class{u_{\nu+1}})}\right]\\&\le\!\Expectation\left[\left\{ \left(\CE{\left(u_{\nu+1}-u_{\nu}\right)\rho}{\Class{u_{\nu+1}}}\!-\!2\Expectation\left[\sqrt{\CE{\Indicator{\!\Action{u_{\nu}}\!\in\![K]\!}}{u_{\nu}}}\Predbound{u_{\nu}-1}{\lambda}{\sigma_{b}}\right]\!-\!\Probability\left(\mathcal{G}_{u_{\nu}}^{c}\right)\right)\vee0\right\} \!\mathbf{1}_{m}^{\top}\tilde{\lambda}_{\CE{\rho_{u_{\nu+1}}}{\Class{u_{\nu+1}}}}^{(\Class{u_{\nu}+1})}\right]\\&\le\!\Expectation\left[\left\{ \CE{\rho_{u_{\nu+1}}}{\Class{u_{\nu+1}}}\vee\mathbf{0_{m}}\right\} ^{\top}\tilde{\lambda}_{\CE{\rho_{u_{\nu+1}}}{\Class{u_{\nu+1}}}}^{(\Class{u_{\nu}+1})}\right],
\end{align*}
where the last inequality holds by \eqref{eq:regret_3}. Because $u_{\nu+1}\in\mathcal{U}$,
we have $\rho_{u_{\nu+1}}>0$ and
\begin{align*}
\Expectation\left[\left\{ \CE{\rho_{u_{\nu+1}}}{\Class{u_{\nu+1}}}\vee\mathbf{0_{m}}\right\} ^{\top}\tilde{\lambda}_{\CE{\rho_{u_{\nu+1}}}{\Class{u_{\nu+1}}}}^{(\Class{u_{\nu}+1})}\right]\le & \Expectation\left[\CE{\rho_{u_{\nu+1}}\vee\mathbf{0_{m}}}{\Class{u_{\nu+1}}}^{\top}\tilde{\lambda}_{\CE{\rho_{u_{\nu+1}}}{\Class{u_{\nu+1}}}}^{(\Class{u_{\nu}+1})}\right]\\
= & \Expectation\left[\CE{\rho_{u_{\nu+1}}}{\Class{u_{\nu+1}}}^{\top}\tilde{\lambda}_{\CE{\rho_{u_{\nu+1}}}{\Class{u_{\nu+1}}}}^{(\Class{u_{\nu}+1})}\right].
\end{align*}
Collecting the bounds, we have
\begin{align*}
\Expectation\left[\left(u_{\nu+1}-u_{\nu}\right)\frac{OPT}{T}\right]\le&\Expectation\left[\CE{\rho_{u_{\nu+1}}}{\Class{u_{\nu+1}}}^{\top}\tilde{\lambda}_{\CE{\rho_{u_{\nu+1}}}{\Class{u_{\nu+1}}}}^{(\Class{u_{\nu}+1})}\right]\\&+\left(2\Expectation\left[\sqrt{\CE{\Indicator{\!\Action{u_{\nu}}\!\in\![K]\!}}{u_{\nu}}}\Predbound{u_{\nu}-1}{\sigma_{b}}{\delta}\right]+\Probability\left(\mathcal{G}_{u_{\nu}}^{c}\right)\right)\frac{OPT}{\rho T}.
\end{align*}
Similar to Step 4, by strong duality, 
\begin{align*}
 & \CE{\rho_{u_{\nu+1}}}{\Class{u_{\nu+1}}}^{\top}\tilde{\lambda}_{\CE{\rho_{u_{\nu+1}}}{\Class{u_{\nu+1}}}}^{(\Class{u_{\nu}+1})}\\
 & =\max_{\sum_{k=1}^{K}\pi_{k}^{(\Class{u_{\nu}+1})}\le1,\pi_{k}^{(\Class{u_{\nu}+1})}\ge0}\min_{\lambda\in\Real_{+}^{m}}\sum_{k=1}^{K}\pi_{k}^{(\Class{u_{\nu}+1})}\OptUtility{\Class{u_{\nu}+1}}k+\left(\CE{\rho_{u_{\nu+1}}}{\Class{u_{\nu+1}}}-\sum_{k=1}^{K}\pi_{k}^{(\Class{u_{\nu}+1})}\mathbf{b}_{k}^{\star(\Class{u_{\nu}+1})}\right)^{\top}\lambda\\
 & \le\min_{\lambda\in\Real_{+}^{m}}\max_{\sum_{k=1}^{K}\pi_{k}^{(\Class{u_{\nu}+1})}\le1,\pi_{k}^{(\Class{u_{\nu}+1})}\ge0}\sum_{k=1}^{K}\pi_{k}^{(\Class{u_{\nu}+1})}\OptUtility{\Class{u_{\nu}+1}}k+\left(\CE{\rho_{u_{\nu+1}}}{\Class{u_{\nu+1}}}-\sum_{k=1}^{K}\pi_{k}^{(\Class{u_{\nu}+1})}\mathbf{b}_{k}^{\star(\Class{u_{\nu}+1})}\right)^{\top}\lambda\\
 & \le\min_{\lambda\in\Real_{+}^{m}}\CE{\max_{\sum_{k=1}^{K}\pi_{k}^{(\Class{u_{\nu}+1})}\le1,\pi_{k}^{(\Class{u_{\nu}+1})}\ge0}\sum_{k=1}^{K}\pi_{k}^{(\Class{u_{\nu}+1})}\OptUtility{\Class{u_{\nu}+1}}k+\left(\rho_{u_{\nu+1}}-\sum_{k=1}^{K}\pi_{k}^{(\Class{u_{\nu}+1})}\mathbf{b}_{k}^{\star(\Class{u_{\nu}+1})}\right)^{\top}\lambda}{\Class{u_{\nu+1}}}\\
 & \le\CE{\max_{\sum_{k=1}^{K}\pi_{k}^{(\Class{u_{\nu}+1})}\le1,\pi_{k}^{(\Class{u_{\nu}+1})}\ge0,\rho_{u_{\nu+1}}-\sum_{k=1}^{K}\pi_{k}^{(\Class{u_{\nu}+1})}\mathbf{b}_{k}^{\star(\Class{u_{\nu}+1})}\ge0}\sum_{k=1}^{K}\pi_{k}^{(\Class{u_{\nu}+1})}\OptUtility{\Class{u_{\nu}+1}}k}{\Class{u_{\nu+1}}}\\
 & =\sum_{k=1}^{K}\TildePolicy{\Class{u_{\nu+1}}}k{\rho_{u_{\nu+1}}}\OptUtility{\Class{u_{\nu}+1}}k.
\end{align*}
Thus we have
\begin{align*}
\Expectation\left[\left(u_{\nu+1}-u_{\nu}\right)\frac{OPT}{T}\right]\le&\Expectation\left[\sum_{k=1}^{K}\TildePolicy{\Class{u_{\nu+1}}}k{\rho_{u_{\nu+1}}}\OptUtility{\Class{u_{\nu}+1}}k\right]\\&+\left(2\Expectation\left[\sqrt{\CE{\Indicator{\!\Action{u_{\nu}}\!\in\![K]\!}}{u_{\nu}}}\Predbound{u_{\nu}-1}{\sigma_{b}}{\delta}\right]+\Probability\left(\mathcal{G}_{u_{\nu}}^{c}\right)\right)\frac{OPT}{\rho T}.
\end{align*}
Under the event $\mathcal{G}_{u_{\nu+1}}$, the policy $\TildePolicy{\Class{u_{\nu+1}}}k{\rho_{u_{\nu+1}}}$
is a feasible solution to the bandit problem \eqref{eq:bandit_problem},
\begin{align*}
\Expectation\left[\sum_{k=1}^{K}\TildePolicy{\Class{u_{\nu+1}}}k{\rho_{u_{\nu+1}}}\OptUtility{\Class{u_{\nu}+1}}k\right]\le & \Expectation\left[\sum_{k=1}^{K}\TildePolicy{\Class{u_{\nu+1}}}k{\rho_{u_{\nu+1}}}\OptUtility{\Class{u_{\nu}+1}}k\Indicator{\mathcal{G}_{u_{\nu+1}}}\right]+\Probability\left(\mathcal{G}_{u_{\nu+1}}^{c}\right)\\
\le & \Expectation\left[\sum_{k=1}^{K}\TildePolicy{\Class{u_{\nu+1}}}k{\rho_{u_{\nu+1}}}\TildeUtility{\Class{u_{\nu+1}}}k{u_{\nu+1}}\Indicator{\mathcal{G}_{u_{\nu+1}}}\right]+\Probability\left(\mathcal{G}_{u_{\nu+1}}^{c}\right)\\
\le & \Expectation\left[\sum_{k=1}^{K}\HatPolicy{\Class{u_{\nu+1}}}k{u_{\nu+1}}\TildeUtility{\Class{u_{\nu+1}}}k{u_{\nu+1}}\Indicator{\mathcal{G}_{u_{\nu+1}}}\right]+\Probability\left(\mathcal{G}_{u_{\nu+1}}^{c}\right).
\end{align*}
 Thus, for each $\nu\in[|\mathcal{U}|-1]$,
\begin{align*}
\Expectation\left[\left(u_{\nu+1}-u_{\nu}\right)\frac{OPT}{T}\right]\le&\Expectation\left[\sum_{k=1}^{K}\HatPolicy{\Class{u_{\nu+1}}}k{u_{\nu+1}}\TildeUtility{\Class{u_{\nu+1}}}k{u_{\nu+1}}\Indicator{\mathcal{G}_{u_{\nu+1}}}\right]+\Probability\left(\mathcal{G}_{u_{\nu+1}}^{c}\right)\\&+\left(2\Expectation\left[\sqrt{\CE{\Indicator{\!\Action{u_{\nu}}\!\in\![K]\!}}{u_{\nu}}}\Predbound{u_{\nu}-1}{\sigma_{b}}{\delta}\right]+\Probability\left(\mathcal{G}_{u_{\nu}}^{c}\right)\right)\frac{OPT}{\rho T}.
\end{align*}
Summing up over $\nu$,
\begin{align*}
\Expectation\left[\sum_{\nu=1}^{|\mathcal{U}|-1}\left(u_{\nu+1}-u_{\nu}\right)\frac{OPT}{T}\right]\le&\Expectation\left[\sum_{t\in\mathcal{U}}\sum_{k=1}^{K}\HatPolicy{\Class t}kt\TildeUtility{\Class t}kt\Indicator{\mathcal{G}_{t}}\right]+\left(1+\frac{OPT}{\rho T}\right)\sum_{t=1}^{T}\Probability\left(\mathcal{G}_{t}^{c}\right)\\&+\left(\sum_{\nu=1}^{|\mathcal{U}|-1}2\Expectation\left[\sqrt{\CE{\Indicator{\!\Action{u_{\nu}}\!\in\![K]\!}}{u_{\nu}}}\Predbound{u_{\nu}-1}{\sigma_{b}}{\delta}\right]\right)\frac{OPT}{\rho T}\\\le&\Expectation\left[\sum_{t\in\mathcal{U}}\sum_{k=1}^{K}\HatPolicy{\Class t}kt\TildeUtility{\Class t}kt\Indicator{\mathcal{G}_{t}}\right]+\left(1+\frac{OPT}{\rho T}\right)\sum_{t=1}^{T}\Probability\left(\mathcal{G}_{t}^{c}\right)\\&+2\left(\sum_{t=1}^{T}\Expectation\left[\sqrt{\Expectation\left[\Indicator{\!\Action t\!\in\![K]\!}\right]}\Predbound{t-1}{\sigma_{b}}{\delta}\right]\right)\frac{OPT}{\rho T}\\\le&\Expectation\left[\sum_{t\in\mathcal{U}}\sum_{k=1}^{K}\HatPolicy{\Class t}kt\TildeUtility{\Class t}kt\Indicator{\mathcal{G}_{t}}\right]+\left(1+\frac{OPT}{\rho T}\right)\sum_{t=1}^{T}\Probability\left(\mathcal{G}_{t}^{c}\right)\\&+2\sqrt{T\Expectation\left[\sum_{t=1}^{T}\Predbound{t-1}{\sigma_{b}}{\delta}^{2}\Indicator{\Action t\in[K]}\right]}\frac{OPT}{\rho T},
\end{align*}
where the last inequality holds by Cauchy-Schwartz inequality, 
By~\eqref{eq:regret_1},
\begin{align*}
\Expectation\left[\sum_{\nu=1}^{|\mathcal{U}|-1}\left(u_{\nu+1}-u_{\nu}\right)\frac{OPT}{T}\right]\le&\Expectation\left[\sum_{t\in\mathcal{U}}\sum_{k=1}^{K}\HatPolicy{\Class t}kt\TildeUtility{\Class t}kt\Indicator{\mathcal{G}_{t}}\right]+\left(1+\frac{OPT}{\rho T}\right)\sum_{t=1}^{T}\Probability\left(\mathcal{G}_{t}^{c}\right)\\&+2\sqrt{T\Expectation\left[\sum_{t=1}^{T}\Predbound{t-1}{\sigma_{b}}{\delta}^{2}\Indicator{\Action t\in[K]}\right]}\frac{OPT}{\rho T}\\\le&\Expectation\left[\sum_{t=1}^{T}R_{t}^{\widehat{\pi}}\right]+\left(2+\frac{OPT}{\rho T}\right)\sum_{t=1}^{T}\Probability\left(\mathcal{G}_{t}^{c}\right)\\&+2\sqrt{T\Expectation\left[\sum_{t=1}^{T}\Predbound{t-1}{\sigma_{b}}{\delta}^{2}\Indicator{\Action t\in[K]}\right]}\frac{OPT}{\rho T}\\&2\sqrt{T\Expectation\left[\sum_{t=1}^{T}\Predbound{t-1}{\sigma_{r}}{\delta}^{2}\Indicator{\Action t\in[K]}\right]}\\\le&\Expectation\left[\sum_{t=1}^{T}R_{t}^{\widehat{\pi}}\right]+\left(2+\frac{OPT}{\rho T}\right)\sum_{t=1}^{T}\Probability\left(\mathcal{G}_{t}^{c}\right)\\&2\left(1+\frac{OPT}{\rho T}\right)\sqrt{T\Expectation\left[\sum_{t=1}^{T}\Predbound{t-1}{\sigma_{b}\vee\sigma_{r}}{\delta}^{2}\Indicator{\Action t\in[K]}\right]}\end{align*}
Because the last choice of the algorithm happens at round $\tau$, we have $\rho_{\tau}>0$
and $u_{|\mathcal{U}|}=\tau$. 
And by definition, $u_{1}=\xi$. 
Thus
\[
\Expectation\left[\sum_{\nu=1}^{|\mathcal{U}|-1}\left(u_{\nu+1}-u_{\nu}\right)\frac{OPT}{T}\right]=\Expectation\left[\left(u_{\abs{\mathcal{U}}}-u_{1}\right)\frac{OPT}{T}\right]=\frac{OPT}{T}\Expectation\left[\tau-\xi\right].
\]
Rearranging the terms
\begin{align*}
\Expectation\left[\sum_{t=1}^{T}R_{t}^{\widehat{\pi}}\right]&\ge\Expectation\left[\sum_{\nu=1}^{|\mathcal{U}|-1}\left(u_{\nu+1}-u_{\nu}\right)\frac{OPT}{T}\right]-\left(2+\frac{OPT}{\rho T}\right)\sum_{t=1}^{T}\Probability\left(\mathcal{G}_{t}^{c}\right)\\&-2\left(1+\frac{OPT}{\rho T}\right)\sqrt{T\Expectation\left[\sum_{t=1}^{T}\Predbound{t-1}{\sigma_{b}\vee\sigma_{r}}{\delta}^{2}\Indicator{\Action t\in[K]}\right]}\\&\ge\frac{OPT}{T}\Expectation\left[\tau-\xi\right]-\left(2+\frac{OPT}{\rho T}\right)\sum_{t=1}^{T}\Probability\left(\mathcal{G}_{t}^{c}\right)\\&-2\left(1+\frac{OPT}{\rho T}\right)\sqrt{T\Expectation\left[\sum_{t=1}^{T}\Predbound{t-1}{\sigma_{b}\vee\sigma_{r}}{\delta}^{2}\Indicator{\Action t\in[K]}\right]},
\end{align*}
completes the proof.
\end{proof}

\subsection{Proof of Lemma~\ref{lem:exploration_bound}}
\begin{proof}
Let us fix $\delta\in(0,T^{-2})$ throughout the proof.
\paragraph{Step 1. Bounding the minimum eigenvalue of $\{F_{\nu}:\nu\in[n_{T}]\}$:}
 By Lemma \ref{lem:matrix_neg},
with probability at least $1-T\delta$, 
\begin{align*}
\frac{1}{2Kd}F_{\nu}= & \frac{1}{2Kd}\sum_{u=1}^{\nu}\StackedContext k{\AdmitRound u}\StackedContext k{\AdmitRound u}^{\top}+8\frac{K-1}{K}\log\frac{Jd}{\delta}\\
\succeq & \frac{1}{4Kd}\sum_{u=1}^{\nu}\CE{\StackedContext k{\AdmitRound u}\StackedContext k{\AdmitRound u}^{\top}}{\History{\AdmitRound u-1}}+8\frac{K-1}{K}\log\frac{Jd}{\delta}-\log\frac{Jd}{\delta}\\
\succeq & \frac{1}{4Kd}\sum_{u=1}^{\nu}\CE{\StackedContext k{\AdmitRound u}\StackedContext k{\AdmitRound u}^{\top}}{\History{\AdmitRound u-1}},
\end{align*}
for all $\nu\in[n_{T}]$. By Assumption \ref{assum:independent_contexts}
and \ref{assum:positive_definiteness}, 
\begin{align*}
\Mineigen{\CE{\StackedContext k{\AdmitRound u}\StackedContext k{\AdmitRound u}^{\top}}{\History{\AdmitRound u-1}}}= & \lambda_{\min}\!\begin{pmatrix}p_{1}\Expectation_{X_{k}\sim\Classdistribution 1}\left[\sum_{k=1}^{K}X_{k}X_{k}^{\top}\right] & 0 & 0\\
0 & \ddots & 0\\
0 & 0 & p_{J}\Expectation_{\mathbf{x}_{k}\sim\Classdistribution J}\left[\sum_{k=1}^{K}X_{k}X_{k}^{\top}\right]
\end{pmatrix}\\
\ge & p_{\min}\min_{j\in[J]}\Mineigen{\Expectation_{X_{k}\sim\Classdistribution j}\left[\sum_{k=1}^{K}X_{k}X_{k}^{\top}\right]}\\
\ge & p_{\min}K\alpha.
\end{align*}
Thus, with probability at least $1-\delta/T$, 
\begin{align*}
\Mineigen{F_{\nu}}\ge & \frac{1}{2}\Mineigen{\sum_{u=1}^{\nu}\CE{\StackedContext ku\StackedContext ku^{\top}}{\History{u-1}}}\\
\ge & \frac{1}{2}\sum_{u=1}^{\nu}\Mineigen{\CE{\StackedContext ku\StackedContext ku^{\top}}{\History{u-1}}}\\
\ge & \frac{p_{\min}K\alpha\nu}{2},
\end{align*}
for all $\nu\in[n_{T}]$.

\paragraph{Step 2. Bounding the probability of $\mathcal{M}_{t}$:} 
Under the event proved in Step 1, the event $\mathcal{M}_{t}$ is implied
by
\begin{equation}
\frac{p_{\min}K\alpha n_{t}}{2}\ge 4Kd\left\{ \sum_{\nu=1}^{n_{t}}\frac{288\left(K-1\right)\log\left(\frac{Jd}{\delta}\right)}{\alpha Kp_{\min}\nu}+35\log\frac{Jd}{\delta}\right\},
\label{eq:exploration1}
\end{equation}
for all $t\in[T]$. 
The right hand side is bounded as
\begin{align*}
4Kd\left\{ \sum_{\nu=1}^{n_{t}}\frac{288\left(K-1\right)\log\left(\frac{Jd}{\delta}\right)}{\alpha Kp_{\min}\nu}+35\log\frac{Jd}{\delta}\right\} 
\le&\frac{4\cdot288Kd\log\left(\frac{Jd}{\delta}\right)\log n_{t}}{\alpha p_{\min}}+140Kd\log\frac{Jd}{\delta}
\\
\le&\frac{4\cdot288Kd\log\left(\frac{Jd}{\delta}\right)\log T}{\alpha p_{\min}}+140Kd\log\frac{Jd}{\delta}.
\end{align*}
Plugging in \eqref{eq:exploration1} and rearranging the terms, 
\[
n_{t}\ge d\log\left(\frac{Jd}{\delta}\right)\left\{ \frac{8\cdot288\log T}{\alpha^{2}p_{\min}^{2}}+\frac{280}{\alpha p_{\min}}\right\} ,
\]
implies the event $\mathcal{M}_{t}$ for all $t\in[T]$ with probability
at least $1-T\delta$. 
In other words, 
\[
\Probability\left(\mathcal{M}_{t}^{c}\right)\le\Probability\left(n_{t}<dM_{\alpha,p,T}\log\left(\frac{Jd}{\delta}\right)\right)+T\delta,
\]
for all $t\in[T]$, where $M_{\alpha,p,T}:=\left\{ \frac{8\cdot288\log T}{\alpha^{2}p_{\min}^{2}}+\frac{280}{\alpha p_{\min}}\right\}$.

\paragraph{Step 3. Bounding $\xi$:} 
Let $\tilde{t}=\inf_{t\in[T]}\{\mathcal{M}_{t}\text{ happens}\}$ be the first round that $\mathcal{M}_{t}$ happens. 
After round $\tilde{t}$, the algorithm skips the rounds until $\rho_{t}>0$ holds and then
pulls an action according to the policy. Thus, for the round $\xi-1$,
\begin{align*}
\left(\xi-1\right)\rho-\sum_{s=1}^{\xi-2}\Consumption{\Class s}{\Action s}s= & \left(\xi-1\right)\rho-\sum_{s=1}^{\tilde{t}}\Consumption{\Class s}{\Action s}s\le0.
\end{align*}
rearranging the terms, and taking expectation,
\begin{equation}
\Expectation\left[\xi\right]\le1+\rho^{-1}\Expectation\left[\sum_{s=1}^{\tilde{t}}\Consumption{\Class s}{\Action s}s\right]\le1+\rho^{-1}\Expectation\left[\tilde{t}\right].
\label{eq:exploration2}
\end{equation}
Now we need an upper bound for $\tilde{t}$. 
For $t\in[\tilde{t}-1]$,
the event $\mathcal{M}_{t}$ does not happen and the algorithm admits
the arrival for $t\in[\tilde{t}]$. Thus, $n_{t}=t$ for all $t\in[\tilde{t}]$.
For $t=\tilde{t}-1$, the event $\mathcal{M}_{\tilde{t}-1}$ does
not happen and
\[
\Mineigen{F_{n_{\tilde{t}-1}}}\le4Kd\left\{ \sum_{\nu=1}^{n_{\tilde{t}-1}}\frac{144\left(K-1\right)\log\left(\frac{Jd}{\delta}\right)}{\Mineigen{F_{\nu}}}+35\log\frac{Jd}{\delta}\right\} .
\]
By the fact proved in Step 1, with probability at least $1-\delta/T$,
\[
\frac{p_{\min}K\alpha n_{\tilde{t}-1}}{2}\le4Kd\left\{ \sum_{\nu=1}^{n_{\tilde{t}-1}}\frac{288\left(K-1\right)\log\left(\frac{Jd}{\delta}\right)}{p_{\min}K\alpha\nu}+35\log\frac{Jd}{\delta}\right\} .
\]
Plugging in $n_{\tilde{t}-1}=\tilde{t}-1$ and rearranging the terms,
\begin{align*}
\tilde{t}-1\le&\frac{4d}{p_{\min}\alpha}\left\{ \sum_{\nu=1}^{\tilde{t}-1}\frac{288\left(K-1\right)\log\left(\frac{Jd}{\delta}\right)}{p_{\min}K\alpha\nu}+35\log\frac{Jd}{\delta}\right\} 
\\
\le&\frac{4d}{p_{\min}\alpha}\left\{ \frac{288\log T}{p_{\min}\alpha}+2\right\} \log\frac{Jd}{\delta}.
\end{align*}
Then with probability at least $1-\delta/T$,
\begin{align*}
\tilde{t}\le&1+d\log\left(\frac{Jd}{\delta}\right)\left\{ \frac{8\cdot288\log T}{\alpha^{2}p_{\min}^{2}}+\frac{140}{\alpha p_{\min}}\right\} 
\\
:=&1+M_{\alpha,p,T}d\log\left(\frac{Jd}{\delta}\right).
\end{align*}
Thus,
\begin{align*}
\Expectation\left[\tilde{t}\right]= & \Expectation\left[\tilde{t}\Indicator{\tilde{t}<1+M_{\alpha,p,T}d\log\left(\frac{Jd}{\delta}\right)}\right]+\Expectation\left[\tilde{t}\Indicator{\tilde{t}\ge1+M_{\alpha,p,T}d\log\left(\frac{Jd}{\delta}\right)}\right]\\
\le & 1+dM_{\alpha,p,T}\log\left(\frac{Jd}{\delta}\right)+T\Probability\left(\tilde{t}\ge1+M_{\alpha,p,T}d\log\left(\frac{Jd}{\delta}\right)\right)\\
\le & 1+dM_{\alpha,p,T}\log\left(\frac{Jd}{\delta}\right)+T^{2}\delta
\end{align*}
Plugging in \eqref{eq:exploration2},
\begin{align*}
\Expectation\left[\xi\right]\le & 1+\rho^{-1}\Expectation\left[\tilde{t}\right]\\
\le & 1+\frac{1+dM_{\alpha,p,T}\log\left(\frac{Jd}{\delta}\right)+T^{2}\delta}{\rho}.
\end{align*}

\paragraph{Step 4. Proving the lower bound for $\tau$:} Let $\tau$ be the
stopping time of the algorithm. Because the algorithm admits arrival
at round $\tau$, we have $\rho_{\tau}>0$. From the resource constraint in the bandit problem \eqref{eq:bandit_problem},
\[
\sum_{k=1}^{K}\HatPolicy{\Class{\tau}}k{\tau}\left(\widehat{\mathbf{b}}_{k,\tau}^{(\Class{\tau})}-\frac{\Predbound{\tau-1}{\sigma_{b}}{\delta}}{\sqrt{p_{\Class{\tau}}}}\mathbf{1}_{m}\right):=\sum_{k=1}^{K}\HatPolicy{\Class{\tau}}k{\tau}\tilde{\mathbf{b}}_{k,\tau}^{(\Class{\tau})}:\le\tau\rho-\sum_{s=1}^{\tau-1}\Consumption{\Class s}{\Action s}s
\]
Because algorithm stops at round $\tau$, there exists an $r\in[m]$
such that $\sum_{s=1}^{\tau}\Conentry{\Class s}{\Action s}sr\ge T\rho(r)$.
Rearranging the terms,
\[
\begin{split}
\tau\rho\ge&\sum_{s=1}^{\tau-1}\Conentry{\Class s}{\Action s}sr+\sum_{k=1}^{K}\HatPolicy{\Class{\tau}}k{\tau}\tilde{b}_{k,\tau}^{(\Class{\tau})}(r)\\\ge&T\rho-\Conentry{\Class{\tau}}{\Action{\tau}}{\tau}r+\sum_{k=1}^{K}\HatPolicy{\Class{\tau}}k{\tau}\tilde{b}_{k,\tau}^{(\Class{\tau})}(r)\\=&T\rho-\Conentry{\Class{\tau}}{\Action{\tau}}{\tau}r+\sum_{k=1}^{K}\HatPolicy{\Class{\tau}}k{\tau}b_{k,\tau}^{\star(\Class{\tau})}(r)+\sum_{k=1}^{K}\HatPolicy{\Class{\tau}}k{\tau}\left(\tilde{b}_{k,\tau}^{(\Class{\tau})}(r)-b_{k,\tau}^{\star(\Class{\tau})}(r)\right)\\\ge&T\rho-\Conentry{\Class{\tau}}{\Action{\tau}}{\tau}r+\sum_{k=1}^{K}\HatPolicy{\Class{\tau}}k{\tau}b_{k}^{\star(\Class{\tau})}(r)-\sum_{k=1}^{K}\HatPolicy{\Class{\tau}}k{\tau}\norm{\tilde{\mathbf{b}}_{k,\tau}^{(\Class{\tau})}-\Optcon{\Class{\tau}}k}_{\infty}.
\end{split}
\]
Taking expectation on both side,
\begin{equation}
\begin{split}
\Expectation\left[\tau\rho\right]\ge&T\rho+\Expectation\left[-\Conentry{\Class{\tau}}{\Action{\tau}}{\tau}r+\sum_{k=1}^{K}\HatPolicy{\Class{\tau}}k{\tau}b_{k,\tau}^{\star(\Class{\tau})}(r)\right]-\Expectation\left[\sum_{k=1}^{K}\HatPolicy{\Class{\tau}}k{\tau}\norm{\tilde{\mathbf{b}}_{k,\tau}^{(\Class{\tau})}-\Optcon{\Class{\tau}}k}_{\infty}\right]\\=&T\rho-\Expectation\left[\sum_{k=1}^{K}\HatPolicy{\Class{\tau}}k{\tau}\norm{\tilde{\mathbf{b}}_{k,\tau}^{(\Class{\tau})}-\Optcon{\Class{\tau}}k}_{\infty}\right]\\=&T\rho-\Expectation\left[\sum_{k=1}^{K}\HatPolicy{\Class{\tau}}k{\tau}\norm{\tilde{\mathbf{b}}_{k,\tau}^{(\Class{\tau})}-\Optcon{\Class{\tau}}k}_{\infty}\Indicator{\mathcal{E}_{\tau}\cap\mathcal{M}_{\tau-1}}\right]-\Expectation\left[\sum_{k=1}^{K}\HatPolicy{\Class{\tau}}k{\tau}\norm{\tilde{\mathbf{b}}_{k,\tau}^{(\Class{\tau})}-\Optcon{\Class{\tau}}k}_{\infty}\Indicator{\mathcal{E}_{\tau}^{c}\cup\mathcal{M}_{\tau-1}^{c}}\right]\\\ge&T\rho-2\Expectation\left[\Predbound{\tau-1}{\sigma_{b}}{\delta}\sqrt{\CE{\Indicator{\Action t\in[K]}}{\tau}}\right]-\Expectation\left[\sum_{k=1}^{K}\HatPolicy{\Class{\tau}}k{\tau}\norm{\tilde{\mathbf{b}}_{k,\tau}^{(\Class{\tau})}-\Optcon{\Class{\tau}}k}_{\infty}\Indicator{\mathcal{E}_{\tau}^{c}\cup\mathcal{M}_{\tau-1}^{c}}\right],
\end{split}
\label{eq:resource1}
\end{equation}
where the last inequality holds by~\eqref{eq:b_tilde_l2_bound}.
Because $\tilde{\mathbf{b}}_{k,\tau}^{(\Class{\tau})}(r)\le T\rho$ almost surely,
\begin{align*}
\Expectation\left[\sum_{k=1}^{K}\HatPolicy{\Class{\tau}}k{\tau}\norm{\tilde{\mathbf{b}}_{k,\tau}^{(\Class{\tau})}-\Optcon{\Class{\tau}}k}_{\infty}\Indicator{\mathcal{E}_{\tau}^{c}\cup\mathcal{M}_{\tau-1}^{c}}\right]\le&T\rho\Probability\left(\mathcal{E}_{\tau}^{c}\cup\mathcal{M}_{\tau-1}^{c}\right)\\
=&T\rho\Probability\left(\mathcal{E}_{\tau}^{c}\right)\\
\le&T\rho\left(4(m+1)\delta+7T^{-1}\right)\\=&7\rho+4(m+1)T\delta,
\end{align*}
where the equality holds because the algorithm takes action according to the policy at round $\tau$ and the last inequality holds by Theorem
\ref{thm:u_convergence}. 
from \eqref{eq:resource1},
\begin{align*}
\Expectation\left[\tau\rho\right]\ge&T\rho-7\rho+4(m+1)T\rho\delta-2\Expectation\left[\Predbound{\tau-1}{\sigma_{b}}{\delta}\sqrt{\CE{\Indicator{\Action t\in[K]}}{\tau}}\right]\\\ge&T\rho-7\rho+4(m+1)T\rho\delta-2\Predbound 1{\sigma_{b}}{\delta}
\end{align*}
Rearranging the terms,
\begin{align*}
\Expectation\left[T-\tau\right]\le&\frac{4(m+1)T\delta+7+2\Predbound{\tau-1}{\sigma_{b}}{\delta}}{\rho}.
\end{align*}

\paragraph{Step 5. Proving a bound for the sum of probabilities} 
Because the algorithm admits the arrival when $\mathcal{M}_{t-1}$ does not happen, 
\[
\mathcal{M}_{t-1}^{c}=\mathcal{M}_{t-1}^{c}\cap\left\{ \Action t\in[K]\right\} .
\]
 Then
\begin{align*}
\Probability\left(\mathcal{M}_{t-1}^{c}\right)=&\Probability\left(\mathcal{M}_{t-1}^{c}\cap\left\{ \Action t\in[K]\right\} \right)\\=&\Probability\left(\mathcal{M}_{t-1}^{c}\cap\left\{ \Action t\in[K]\right\} \cap\left\{ n_{t-1}\ge M_{\alpha,p,T}d\log\left(\frac{Jd}{\delta}\right)\right\} \right)\\&+\Probability\left(\mathcal{M}_{t-1}^{c}\cap\left\{ \Action t\in[K]\right\} \cap\left\{ n_{t-1}<M_{\alpha,p,T}d\log\left(\frac{Jd}{\delta}\right)\right\} \right)\\\le&\Probability\left(\mathcal{M}_{t-1}^{c}\cap\left\{ n_{t-1}\ge M_{\alpha,p,T}d\log\left(\frac{Jd}{\delta}\right)\right\} \right)\\&+\Probability\left(\left\{ \Action t\in[K]\right\} \cap\left\{ n_{t-1}<M_{\alpha,p,T}d\log\left(\frac{Jd}{\delta}\right)\right\} \right)\\\le&T\delta+\Probability\left(\left\{ \Action t\in[K]\right\} \cap\left\{ n_{t-1}<M_{\alpha,p,T}d\log\left(\frac{Jd}{\delta}\right)\right\} \right),
\end{align*}
where the last inequality holds by the fact proved in Step 2. Summing
over $t\in[T]$, 
\begin{align*}
\sum_{t=1}^{T}\Probability\left(\mathcal{M}_{t-1}^{c}\right)\le&T^{2}\delta+\sum_{t=1}^{T}\Probability\left(\left\{ \Action t\in[K]\right\} \cap\left\{ n_{t-1}<M_{\alpha,p,T}d\log\left(\frac{Jd}{\delta}\right)\right\} \right)\\=&T^{2}\delta+\Expectation\left[\sum_{t=1}^{T}\Indicator{\Action t\in[K]}\Indicator{n_{t-1}<M_{\alpha,p,T}d\log\left(\frac{Jd}{\delta}\right)}\right].
\end{align*}
Set $\mu:=M_{\alpha,p,T}d\log\left(\frac{Jd}{\delta}\right)$ and
suppose 
\begin{equation}
\sum_{t=1}^{T}\Indicator{\Action t\in[K]}\Indicator{n_{t-1}<\mu}>\mu.\label{eq:exploration3}
\end{equation}
Let $\AdmitRound 1<\AdmitRound 2<\cdots<\AdmitRound{|\mathcal{A}|}$
be the ordered admitted round in $\mathcal{A}:=\{t\in[T]:\Action t\in[K]\}$.
By definition, $n_{\AdmitRound{\nu}}=\nu$ for $\nu\in[|\mathcal{A}|]$.
By \eqref{eq:exploration3}, the event$\left\{ \Action t\in[K]\right\} $
happens at least $\mu+1$ times over the horizon $[T]$ and $|\mathcal{A}|>\mu$.
For any $\nu\in(\mu,|\mathcal{A}|]$,the number of admitted round
is $n_{\AdmitRound{\nu}}>\mu$ and
\begin{align*}
\sum_{t=\varepsilon}^{T-1}\Indicator{n_{t-1}<\mu}\Indicator{\Action t\in[K]}= & \sum_{\nu=1}^{|\mathcal{A}|}\Indicator{n_{\AdmitRound{\nu}-1}<\mu}\Indicator{\Action{\AdmitRound{\nu}}\in[K]}\\
\le & \sum_{\nu=1}^{|\mathcal{A}|}\Indicator{n_{\AdmitRound{\nu}-1}<\mu}\Indicator{n_{\AdmitRound{\nu}}=n_{\AdmitRound{\nu}-1}+1}\\
= & \sum_{\nu=1}^{|\mathcal{A}|}\Indicator{n_{\AdmitRound{\nu}-1}<\mu}\Indicator{\nu=n_{\AdmitRound{\nu}-1}+1}\\
\le & \sum_{\nu=1}^{|\mathcal{A}|}\Indicator{\nu-1<\mu},\\
= & \sum_{\nu=1}^{|\mathcal{A}|}\Indicator{\nu<\mu+1}\\
= & \mu,
\end{align*}
which contradicts with \eqref{eq:exploration3}. Thus
\[
\Expectation\left[\sum_{t=1}^{T}\Indicator{\Action t\in[K]}\Indicator{n_{t-1}<M_{\alpha,p,T}d\log\left(\frac{Jd}{\delta}\right)}\right]\le\mu:=M_{\alpha,p,T}d\log\left(\frac{Jd}{\delta}\right),
\]
which proves,

\[
\sum_{t=1}^{T}\Probability\left(\mathcal{M}_{t-1}^{c}\right)\le T^{2}\delta+M_{\alpha,p,T}d\log\left(\frac{Jd}{\delta}\right).
\]
\end{proof}

\subsection{Proof of Theorem \ref{thm:regret_bound}}
\begin{proof}
From Lemma \ref{lem:reward_lower_bound}, rearranging the terms,
\begin{align*}
\mathcal{R}_{T}^{\widehat{\pi}}:= & OPT-\Expectation\left[\sum_{t=1}^{T}R_{t}^{\widehat{\pi}}\right]\\
\le & \frac{OPT}{T}\left\{ T-\Expectation\left[\tau-\xi\right]\right\} \\
 & +\left(2+\frac{OPT}{\rho T}\right)\sum_{t=1}^{T}\Probability\left(\mathcal{M}_{t-1}^{c}\cup\mathcal{E}_{t}^{c}\right)\\
 & +2\sqrt{T\Expectation\left[\sum_{t=1}^{T}\Predbound{t-1}{\sigma_{r}}{\delta}^{2}\Indicator{\Action t\in[K]}\right]}\\
 & +2\sqrt{T\Expectation\left[\sum_{t=1}^{T}\Predbound{t-1}{\sigma_{b}}{\delta}^{2}\Indicator{\Action t\in[K]}\right]}\frac{OPT}{\rho T}.
\end{align*}
By Lemma \ref{lem:exploration_bound},
\begin{align*}
\Expectation\left[\xi\right] & \le1+\frac{1\!+\!dM_{\alpha,p,T}\log\left(\frac{Jd}{\delta}\right)\!+\!T^{2}\delta}{\rho},\\
\Expectation\left[T-\tau\right]&\le\frac{4(m+1)T\delta+7+2\Predbound{\tau-1}{\sigma_{b}}{\delta}}{\rho}.
\end{align*}
By definition of $\Predbound{t}{\sigma}{\delta}$,
\begin{align*}
\Expectation\left[T-\tau\right]\le&\frac{4(m+1)T\delta+7+32\sqrt{J\log JKT}+12\Selfbound{\sigma_{b}}{\delta}}{\rho}
\\
=&\frac{4(m+1)T\delta+7+32\sqrt{J\log JKT}+C_{\sigma}(\delta)\sqrt{Jd}}{\rho}.
\end{align*}
This implies
\begin{align*}
&\frac{OPT}{T}\left\{ T-\Expectation\left[\tau-\xi\right]\right\} \\&\le\frac{OPT}{T\rho}\left(\rho+4(m+1)T\delta+8+32\sqrt{J\log\left(\frac{JK}{\delta}\right)}\!+\!C_{\sigma_{b}}(\delta)\sqrt{Jd}+dM_{\alpha,p,T}\log\left(\frac{Jd}{\delta}\right)\!+\!T^{2}\delta\right)\\&\le\frac{OPT}{T\rho}\left(\rho+8+\left(5mT+T^{2}\right)\delta+32\sqrt{J\log\left(\frac{JK}{\delta}\right)}\!+\!C_{\sigma_{b}}(\delta)\sqrt{Jd}+dM_{\alpha,p,T}\log\left(\frac{Jd}{\delta}\right)\right).
\end{align*}

{[}Step 3. Bounding the sum of probability{]} Because $T\ge8d\alpha^{-1}p_{\min}^{-1}\log JdT$, by Theorem~\ref{thm:u_convergence} and Lemma~\ref{lem:exploration_bound},
\begin{align*}
\sum_{t=1}^{T}\Probability\left(\mathcal{M}_{t-1}^{c}\cup\mathcal{E}_{t}^{c}\right)=&\sum_{t=1}^{T}\left\{ \Probability\left(\mathcal{M}_{t-1}^{c}\right)+\Probability\left(\mathcal{M}_{t-1}\cap\mathcal{E}_{t}^{c}\right)\right\} \\
\le&T^{3}\delta+dM_{\alpha,p,T}\log\left(\frac{Jd}{\delta}\right)+\sum_{t=1}^{T}\Probability\left(\mathcal{M}_{t-1}\cap\mathcal{E}_{t}^{c}\right)\\
\le&T^{3}\delta+dM_{\alpha,p,T}\log\left(\frac{Jd}{\delta}\right)+8d\alpha^{-1}p_{\min}^{-1}\log JdT\\&+\sum_{t=8d\alpha^{-1}p_{\min}^{-1}\log JdT}^{T}\Probability\left(\mathcal{M}_{t-1}\cap\mathcal{E}_{t}^{c}\right)\\
\le&T^{3}\delta+dM_{\alpha,p,T}\log\left(\frac{Jd}{\delta}\right)+8d\alpha^{-1}p_{\min}^{-1}\log JdT+4(m+1)T\delta+7.
\end{align*}
By definition of $\Predbound t{\sigma}{\delta}$ and $\Selfbound{\sigma}{\delta}$,
\begin{align*}
\Expectation\left[\sum_{t=1}^{T}\Predbound{t-1}{\sigma_{r}}{\delta}^{2}\Indicator{\Action t\in[K]}\right]=&\Expectation\left[\sum_{t=1}^{T}\left(\frac{16\sqrt{J\log JKT}}{\sqrt{t-1}}+\frac{4\sqrt{2}\Selfbound{\sigma_{r}}{\delta}}{\sqrt{n_{t-1}}}\right)^{2}\Indicator{\Action t\in[K]}\right]\\\le&\Expectation\left[\sum_{t=1}^{T}\frac{\left(16\sqrt{J\log JKT}+4\sqrt{2}\Selfbound{\sigma_{r}}{\delta}\right)^{2}}{n_{t-1}}\Indicator{\Action t\in[K]}\right]\\\le&\Expectation\left[\sum_{t=1}^{T}\frac{\left(16\sqrt{J\log JKT}+4\sqrt{2}\Selfbound{\sigma_{r}}{\delta}\right)^{2}}{n_{t-1}}\Indicator{n_{t}=n_{t-1}+1}\right]\\\le&\left(16\sqrt{J\log JKT}+4\sqrt{2}\Selfbound{\sigma_{r}}{\delta}\right)^{2}\log T,
\end{align*}
where the first inequality holds by $n_{t}\le t$ almost surely. Thus
by definition of $\Selfbound{\sigma}{\delta}:=8\sqrt{Jd}+96\sigma\sqrt{Jd\log\frac{4}{\delta}}$,
\begin{align*}
2\sqrt{T\Expectation\left[\sum_{t=1}^{T}\Predbound{t-1}{\sigma_{r}}{\delta}^{2}\Indicator{\Action t\in[K]}\right]}\le&\left(32\sqrt{J\log JKT}+4\sqrt{6}\Selfbound{\sigma_{r}}{\delta}\right)\sqrt{T\log T}\\\le&\left(32\sqrt{J\log JKT}+C_{\sigma_{r}}(\delta)\sqrt{Jd}\right)\sqrt{T\log T},
\end{align*}
where $C_{\sigma}(\delta):=8\sqrt{2}\cdot\left(8+96\sigma\sqrt{\log\frac{4}{\delta}}\right)$.
Similarly,
\[
2\sqrt{T\Expectation\left[\sum_{t=1}^{T}\Predbound{t-1}{\sigma_{b}}{\delta}^{2}\Indicator{\Action t\in[K]}\right]}\frac{OPT}{\rho T}\le\left(32\sqrt{J\log JKT}+C_{\sigma_{b}}(\delta)\sqrt{Jd}\right)\sqrt{T\log T}\frac{OPT}{\rho T}
\]
Collecting the bounds,
\begin{align*}
\mathcal{R}_{T}^{\widehat{\pi}}&\le\frac{OPT}{T\rho}\left(\rho+8+\left(5mT+T^{2}\right)\delta+32\sqrt{J\log JKT}\!+\!C_{\sigma_{b}}(\delta)\sqrt{Jd}+dM_{\alpha,p,T}\log\left(\frac{Jd}{\delta}\right)\right)\\&+\left(2+\frac{OPT}{\rho T}\right)\left\{ T^{3}\delta+dM_{\alpha,p,T}\log\left(\frac{Jd}{\delta}\right)+4d\alpha^{-1}p_{\min}^{-1}\log JdT+4(m+1)T\delta+7\right\} \\&+2\left(1+\frac{OPT}{\rho T}\right)\left\{ 32\sqrt{J\log JKT}+C_{\sigma_{b}\vee\sigma_{r}}(\delta)\sqrt{Jd}\right\} \sqrt{T\log T}\\&\le\left(2+\frac{OPT}{\rho T}\right)\Bigg\{\left(96\sqrt{J\log JKT}+3C_{\sigma_{r}\vee\sigma_{r}}(\delta)\sqrt{Jd}\right)\sqrt{T\log T}+2dM_{\alpha,p,T}\log\left(\frac{Jd}{\delta}\right)\\&\hspace{2.5cm}+4d\alpha^{-1}p_{\min}^{-1}\log JdT+15+10mT^{3}\delta\Bigg\},
\end{align*}
Plugging in  $\delta=m^{-1}T^{-3}$ proves~\eqref{eq:regret_bound}.
\end{proof}

\subsection{Deriving Regret Bound for Unknown Class Arrival Probabilities}
\label{subsec:unknown_p}
In this section, we provide a direct solution to derive the same theoretical results even if the class prior probabilities are unknown.
For each $j\in[J]$, let $\widehat{p}^{(j)}_{t}:=\frac{1}{t}\sum_{s=1}^{t} I(j_s=j)$ denote the empirical estimate for class prior $p_j$.  
Then, by Lemma~\ref{lem:matrix_neg}, we have
\[
\frac{1}{2} p_j - \frac{\log(2JT^2)}{t} \le \widehat{p}^{(j)}_{t} \le \frac{3}{2} p_j + \frac{\log(2JT^2)}{t},
\]
with probability at least $1-T^{-1}$ for all $j\in[J]$ and $t\in[T]$.
When $t \ge 4p_{\min}^{-1} \log(2JT^2)$, the empirical estimate
$\widehat{p}^{(j)}_{t}$ satisfies $(1/4) p_j \le \widehat{p}^{(j)}_{t} \le
(7/4) p_j$.  
In addition, replacing $p_j$ with $\widehat{p}^{(j)}_{t}$ only affects the
proof of Step 1 in Lemma 5.2, as detailed in Section B.4.  
Therefore, with minor adjustments to the constants in
$\gamma_{t-1,\sigma}(\delta)$, the regret bound remains the same even when the class prior probabilities are unknown.

\section{Technical lemmas}

\begin{lemma}
\label{lem:Azuma_Hoeffding_ineqaulity} (Azuma-Hoeffding's inequality)
\citet{azuma1967weighted} If a super-martingale $(Y_{t};t\ge0)$
corresponding to filtration $\Filtration t$, satisfies $\abs{Y_{t}-Y_{t-1}}\le c_{t}$
for some constant $c_{t}$, for all $t=1,\ldots,T$, then for any
$a\ge0$, 
\[
\Probability\left(Y_{T}-Y_{0}\ge a\right)\le e^{-\frac{a^{2}}{2\sum_{t=1}^{T}c_{t}^{2}}}.
\]
Thus with probability at least $1-\delta$,
\[
Y_{T}-Y_{0}\le\sqrt{2\log\frac{1}{\delta}\sum_{t=1}^{T}c_{t}^{2}}.
\]
\end{lemma}

\begin{lemma}
\label{lem:ordering} For a sequence $u_{1}\ge u_{2}\ge\cdots\ge u_{n}\ge0$
and nonnegative real sequences $\{p_{i}\}_{i\in[n]}$ and $\{q_{i}\}_{i\in[n]}$
such that $\sum_{i=1}^{n}p_{i}=\sum_{i=1}^{n}q_{i}$, if $p_{1}>q_{1}$
then 
\[
\sum_{i=1}^{n}p_{i}u_{i}\ge\sum_{i=1}^{n}q_{i}u_{i}.
\]
\end{lemma}

\begin{proof}
When $n=1$, $p_{1}u_{1}\ge q_{1}u_{1}$, for any $u_{1}\ge0$. Suppose
for any sequence $u_{1}\ge u_{2}\ge\cdots\ge u_{n-1}\ge0$ and nonnegative
real sequences $\{p_{i}\}_{i\in[n-1]}$ and $\{q_{i}\}_{i\in[n-1]}$
such that $\sum_{i=1}^{n-1}p_{i}=\sum_{i=1}^{n-1}q_{i}$, 
\[
p_{1}>q_{1}\Longrightarrow\sum_{i=1}^{n-1}p_{i}u_{i}\ge\sum_{i=1}^{n-1}q_{i}u_{i}.
\]
For a sequence $u_{1}\ge u_{2}\ge\cdots\ge u_{n}\ge0$ and nonnegative
real sequences $\{p_{i}\}_{i\in[n]}$ and $\{q_{i}\}_{i\in[n]}$ such
that $\sum_{i=1}^{n}p_{i}=\sum_{i=1}^{n}q_{i}$, and $p_{1}>q_{1}$,
there exist $k\in[n]\backslash\{1\}$ such that $p_{k}<q_{k}$. In
case of $k=n$, define a sequence
\begin{align*}
\tilde{q}_{i} & =q_{i},\quad\forall i\in[n-2]\\
\tilde{q}_{n-1} & =q_{n-1}-p_{n}+q_{n}\ge0.
\end{align*}
Then $\sum_{i=1}^{n-1}\tilde{q}_{i}=\sum_{i=1}^{n-1}p_{i}$ and
\begin{align*}
\sum_{i=1}^{n}p_{i}u_{i}= & \sum_{i=1}^{n-1}p_{i}u_{i}+p_{n}u_{n}\\
\ge & \sum_{i=1}^{n-1}\tilde{q}_{i}u_{i}+p_{n}u_{n}\\
= & \sum_{i=1}^{n-1}q_{i}u_{i}+\left(-p_{n}+q_{n}\right)u_{n-1}+p_{n}u_{n}\\
\ge & \sum_{i=1}^{n-1}q_{i}u_{i}+\left(-p_{n}+q_{n}\right)u_{n}+p_{n}u_{n}\\
= & \sum_{i=1}^{n}q_{i}u_{i}.
\end{align*}
In case of $k\neq n$, denote a sequence
\begin{align*}
\tilde{q}_{i} & =q_{i},\quad\forall i\in[n-1]\backslash\{k\}\\
\tilde{q}_{k} & =q_{k}-p_{k}+q_{n}.
\end{align*}
Then $\sum_{i=1}^{n-1}\tilde{q}_{i}=\sum_{j\neq k}p_{i}$ and
\begin{align*}
\sum_{i=1}^{n}p_{i}u_{i}= & \sum_{i\neq k}p_{i}u_{i}+p_{k}u_{k}\\
\ge & \sum_{i=1}^{n-1}\tilde{q}_{i}u_{i}+p_{k}u_{k}\\
\ge & \sum_{i=1}^{n-1}q_{i}u_{i}-p_{k}u_{k}+q_{n}u_{k}+p_{k}u_{k}\\
= & \sum_{i=1}^{n-1}q_{k}u_{k}+q_{n}u_{k}\\
\ge & \sum_{i=1}^{n}q_{k}u_{k}.
\end{align*}
By induction, the proof is complete.

\[
\]
\end{proof}
\begin{lemma}
\label{lem:matrix_neg} Let $\{\XX{\tau}:\tau\in[t]\}$ be a $\Real^{d\times d}$-valued
stochastic process adapted to the filtration $\{\Filtration{\tau}:\tau\in[t]\}$,
i.e., $\XX{\tau}$ is $\Filtration{\tau}$-measurable for $\tau\in[t]$.
Suppose $\XX{\tau}$ is a positive definite symmetric matrices such
that$\Maxeigen{\XX{\tau}}\le\frac{1}{2}$.Then with probability at
least $1-\delta,$
\[
\sum_{\tau=1}^{t}\XX{\tau}\succeq\frac{1}{2}\sum_{\tau=1}^{t}\CE{\XX{\tau}}{\Filtration{\tau-1}}-\log\frac{d}{\delta}I_{d}.
\]
In addition, with probability at least $1-\delta$,
\[
\sum_{\tau=1}^{t}\XX{\tau}\preceq\frac{3}{2}\sum_{\tau=1}^{t}\CE{\XX{\tau}}{\Filtration{\tau-1}}+\log\frac{d}{\delta}I_{d}.
\]
 
\end{lemma}

\begin{proof}
This proof is an adapted version of Lemma 12.2 in \citet{lattimore2020bandit}
for matrix stochastic process using the argument of \citet{tropp2012user}.
For the lower bound, It is sufficient to prove that
\[
\Maxeigen{-\sum_{\tau=1}^{t}\XX{\tau}+\frac{1}{2}\sum_{\tau=1}^{t}\CE{\XX{\tau}}{\Filtration{\tau-1}}}\le\log\frac{d}{\delta},
\]
with probability at least $1-\delta$. By the spectral mapping theorem,
\begin{align*}
\exp\left(\Maxeigen{-\sum_{\tau=1}^{t}\XX{\tau}+\frac{1}{2}\sum_{\tau=1}^{t}\CE{\XX{\tau}}{\Filtration{\tau-1}}}\right)\le & \Maxeigen{\exp\left(-\sum_{\tau=1}^{t}\XX{\tau}+\frac{1}{2}\sum_{\tau=1}^{t}\CE{\XX{\tau}}{\Filtration{\tau-1}}\right)}\\
\le & \Trace{\exp\left(-\sum_{\tau=1}^{t}\XX{\tau}+\frac{1}{2}\sum_{\tau=1}^{t}\CE{\XX{\tau}}{\Filtration{\tau-1}}\right)}.
\end{align*}
Taking expectation on both side gives,
\begin{align*}
\Expectation & \exp\left(\Maxeigen{-\sum_{\tau=1}^{t}\XX{\tau}+\frac{1}{2}\sum_{\tau=1}^{t}\CE{\XX{\tau}}{\Filtration{\tau-1}}}\right)\\
\le & \Expectation\Trace{\exp\left(-\sum_{\tau=1}^{t}\XX{\tau}+\frac{1}{2}\sum_{\tau=1}^{t}\CE{\XX{\tau}}{\Filtration{\tau-1}}\right)}\\
= & \Expectation\Trace{\CE{\exp\left(-\sum_{\tau=1}^{t-1}\XX{\tau}+\frac{1}{2}\sum_{\tau=1}^{t}\CE{\XX{\tau}}{\Filtration{\tau-1}}+\log\exp\left(-\XX t\right)\right)}{\Filtration{t-1}}}\\
\le & \Expectation\Trace{\exp\left(-\sum_{\tau=1}^{t-1}\XX{\tau}+\frac{1}{2}\sum_{\tau=1}^{t}\CE{\XX{\tau}}{\Filtration{\tau-1}}+\log\CE{\exp\left(-\XX t\right)}{\Filtration{t-1}}\right)}.
\end{align*}
The last inequality holds due to Lieb's theorem \citet{tropp2015introduction}.
Because $e^{x}\le1+\frac{1}{2}x$for all $x\in[-1/2,0]$, and the
eigenvalue of $-\XX t$ lies in $[-1/2,0]$, we have
\[
\CE{\exp\left(-\XX t\right)}{\Filtration{t-1}}\preceq I-\frac{1}{2}\CE{\XX t}{\Filtration{t-1}}\preceq\exp\left(-\frac{1}{2}\CE{\XX t}{\Filtration{t-1}}\right),
\]
by the spectral mapping theorem. Thus we have
\begin{align*}
 & \Expectation\exp\left(\Maxeigen{-\sum_{\tau=1}^{t}\XX{\tau}+\frac{1}{2}\sum_{\tau=1}^{t}\CE{\XX{\tau}}{\Filtration{\tau-1}}}\right)\\
 & \le\Expectation\Trace{\exp\left(-\sum_{\tau=1}^{t-1}\XX{\tau}+\frac{1}{2}\sum_{\tau=1}^{t}\CE{\XX{\tau}}{\Filtration{\tau-1}}+\log\exp\left(-\frac{1}{2}\CE{\XX t}{\Filtration{t-1}}\right)\right)}\\
 & =\Expectation\Trace{\exp\left(-\sum_{\tau=1}^{t-1}\XX{\tau}+\frac{1}{2}\sum_{\tau=1}^{t}\CE{\XX{\tau}}{\Filtration{\tau-1}}-\frac{1}{2}\CE{\XX t}{\Filtration{t-1}}\right)}\\
 & =\Expectation\Trace{\exp\left(-\sum_{\tau=1}^{t-1}\XX{\tau}+\frac{1}{2}\sum_{\tau=1}^{t-1}\CE{\XX{\tau}}{\Filtration{\tau-1}}\right)}\\
 & \le\vdots\\
 & \le\Expectation\Trace{\exp\left(O\right)}=d
\end{align*}
Now my Markov's inequality,
\begin{align*}
 & \Probability\left(\Maxeigen{-\sum_{\tau=1}^{t}\XX{\tau}+\frac{1}{2}\sum_{\tau=1}^{t}\CE{\XX{\tau}}{\Filtration{\tau-1}}}>\log\frac{d}{\delta}\right)\\
 & \le\Expectation\exp\left(\Maxeigen{-\sum_{\tau=1}^{t}\XX{\tau}+\frac{1}{2}\sum_{\tau=1}^{t}\CE{\XX{\tau}}{\Filtration{\tau-1}}}\right)\frac{\delta}{d}\\
 & \le\delta.
\end{align*}
For the upper bound, we prove
\[
\Maxeigen{\sum_{\tau=1}^{t}\XX{\tau}-\frac{3}{2}\sum_{\tau=1}^{t}\CE{\XX{\tau}}{\Filtration{\tau-1}}}\le\log\frac{d}{\delta},
\]
in a similar way using the fact that $e^{x}\le1+(3/2)x$ on $x\in[0,1/2]$. 
\end{proof}
\begin{lemma}
\label{lem:sub_Gaussian}Suppose a random variable $X$ satisfies
$\Expectation[X]=0$, and let $Y$ be an $\sigma$-sub-Gaussian random
variable. If $\abs X\le\abs Y$ almost surely, then $X$ is $6\sigma$-sub-Gaussian.
\end{lemma}

\begin{proof}
Because $\abs X\le\abs Y$,
\begin{align*}
\Expectation\left[\exp\left(\frac{X^{2}}{6\sigma^{2}}\right)\right]\le & \Expectation\left[\exp\left(\frac{Y^{2}}{6\sigma^{2}}\right)\right]\\
= & 1+\Expectation\left[\int_{0}^{\infty}\Indicator{\abs Y\ge x}\frac{x}{3\sigma^{2}}e^{\frac{x^{2}}{6\sigma^{2}}}dx\right]\\
\le & 1+\int_{0}^{\infty}\Probability\left(\abs Y\ge x\right)\frac{x}{3\sigma^{2}}e^{\frac{x^{2}}{6\sigma^{2}}}dx.
\end{align*}
Because 
\begin{align*}
\Probability\left(\abs Y\ge x\right)= & \Probability\left(Y\ge x\right)+\Probability\left(-Y\le x\right)\\
\le & 2e^{-\frac{x^{2}}{2\sigma^{2}}},
\end{align*}
we have
\begin{align*}
\Expectation\left[\exp\left(\frac{X^{2}}{6\sigma^{2}}\right)\right]\le & 1+\int_{0}^{\infty}\frac{2x}{3\sigma^{2}}e^{-\frac{x^{2}}{3\sigma^{2}}}dx\\
\le & 2.
\end{align*}
Now for any $\lambda\in\Real$,
\begin{align*}
\Expectation\left[\exp\left(\lambda X\right)\right]= & \Expectation\left[\sum_{n=0}^{\infty}\frac{\left(\lambda X\right)^{n}}{n!}\right]\\
= & 1+\Expectation\left[\sum_{n=2}^{\infty}\frac{\left(\lambda X\right)^{n}}{n!}\right]\\
\le & 1+\Expectation\left[\frac{\lambda^{2}X^{2}}{2}\sum_{n=2}^{\infty}\frac{\abs{\lambda X}^{n-2}}{(n-2)!}\right]\\
\le & 1+\frac{\lambda^{2}}{2}\Expectation\left[X^{2}\exp\left(\abs{\lambda X}\right)\right].
\end{align*}
Because $6\sigma^{2}\lambda^{2}+\frac{X^{2}}{12\sigma^{2}}\ge\abs{\lambda X}$
, 
\begin{align*}
\Expectation\left[\exp\left(\lambda X\right)\right]\le & 1+\frac{\lambda^{2}}{2}\exp\left(6\sigma^{2}\lambda^{2}\right)\Expectation\left[X^{2}\exp\left(\frac{X^{2}}{12\sigma^{2}}\right)\right]\\
= & 1+6\sigma^{2}\lambda^{2}\exp\left(6\sigma^{2}\lambda^{2}\right)\Expectation\left[\frac{X^{2}}{12\sigma^{2}}\exp\left(\frac{X^{2}}{12\sigma^{2}}\right)\right]\\
\le & 1+6\sigma^{2}\lambda^{2}\exp\left(6\sigma^{2}\lambda^{2}\right)\Expectation\left[\exp\left(\frac{X^{2}}{6\sigma^{2}}\right)\right]\\
\le & 1+12\sigma^{2}\lambda^{2}\exp\left(6\sigma^{2}\lambda^{2}\right)\\
\le & \left(1+12\sigma^{2}\lambda^{2}\right)\exp\left(6\sigma^{2}\lambda^{2}\right)\\
\le & \exp\left(\frac{36}{2}\sigma^{2}\lambda^{2}\right).
\end{align*}
Thus $X$ is $6\sigma$-sub-Gaussian. 
\end{proof}
\begin{lemma}
\label{lem:dim_reduction}\citep[Lemma 2.3]{lee2016} Let $\left\{ N_{t}\right\} $
be a martingale on a Hilbert space $(\mathcal{H},\norm{\cdot}_{\mathcal{H}})$.
Then there exists a $\Real^{2}$-valued martingale $\left\{ P_{t}\right\} $
such that for any time $t\ge0$, $\norm{P_{t}}_{2}=\norm{N_{t}}_{\mathcal{H}}$
and $\norm{P_{t+1}-P_{t}}_{2}=\norm{N_{t+1}-N_{t}}_{\mathcal{H}}$. 
\end{lemma}

\begin{lemma}
\label{lem:eta_x_bound} (A dimension-free bound for vector-valued
martingales.) Let $\{\Filtration s\}_{s=0}^{t}$ be a filtration and
$\{\eta_{s}\}_{s=1}^{t}$ be a real-valued stochastic process such
that $\eta_{s}$ is $\Filtration{\tau}$-measurable. Let $\left\{ X_{s}\right\} _{s=1}^{t}$
be an $\Real^{d}$-valued stochastic process where $X_{s}$ is $\Filtration 0$-measurable.
Assume that $\{\eta_{s}\}_{s=1}^{t}$ are $\sigma$-sub-Gaussian as
in Assumption \ref{assum:error}. Then with probability at least $1-\delta$,
\begin{equation}
\norm{\sum_{s=1}^{t}\eta_{s}X_{s}}_{2}\le12\sigma\sqrt{\sum_{s=1}^{t}\norm{X_{s}}_{2}^{2}}\sqrt{\log\frac{4t^{2}}{\delta}}.\label{eq:etaX_bound}
\end{equation}
\end{lemma}

\begin{proof}
Fix a $t\ge1$. For each $s=1,\ldots,t$, we have $\CE{\eta_{s}}{\Filtration{s-1}}=0$
and $X_{s}$ is $\Filtration 0$-measurable. Thus the stochastic process,
\begin{equation}
\left\{ \sum_{s=1}^{u}\eta_{s}X_{s}\right\} _{u=1}^{t}
\end{equation}
is a $(\Real^{d},\norm{\cdot}_{2})$-martingale. Since $(\Real^{d},\norm{\cdot}_{2})$
is a Hilbert space, by Lemma \ref{lem:dim_reduction}, there exists
an $\Real^{2}$-martingale $\{M_{u}\}_{u=1}^{t}$ such that 
\begin{equation}
\norm{\sum_{s=1}^{u}\eta_{s}X_{s}}_{2}=\norm{M_{u}}_{2},\;\norm{\eta_{u}X_{u}}_{2}=\norm{M_{u}-M_{u-1}}_{2},
\end{equation}
and $M_{0}=0$. Set $M_{u}=(M_{1}(u),M_{2}(u))^{\top}$. Then for
each $i=1,2$, and $u\ge2$, 
\begin{align*}
\abs{M_{i}(u)-M_{i}(u-1)}\le & \norm{M_{u}-M_{u-1}}_{2}\\
= & \norm{\eta_{u}X_{u}}_{2}\\
= & \abs{\eta_{u}}\norm{X_{u}}_{2},
\end{align*}
almost surely. By Lemma \ref{lem:sub_Gaussian}, $M_{i}(u)-M_{i}(u-1)$
is $6\sigma$-sub-Gaussian. By Lemma \ref{lem:Azuma_Hoeffding_ineqaulity},
for $x>0$, 
\begin{align*}
\Probability\left(\abs{M_{i}(t)}>x\right)= & \Probability\left(\abs{\sum_{u=1}^{t}M_{i}(u)-M_{i}(u-1)}>x\right)\\
\le & 2\exp\left(-\frac{x^{2}}{72t\sigma^{2}\sum_{s=1}^{t}\norm{X_{s}}_{2}^{2}}\right),
\end{align*}
for each $i=1,2$. Thus, with probability $1-\delta/2$, 
\[
M_{i}(t)^{2}\le72\left(\sum_{s=1}^{t}\norm{X_{s}}_{2}^{2}\right)\sigma^{2}\log\frac{4}{\delta}.
\]
In summary, with probability at least $1-\delta/2$, 
\[
\norm{\sum_{\tau=1}^{t}\eta_{s}X_{s}}_{2}=\sqrt{M_{1}(t)^{2}+M_{2}(t)^{2}}\le6\sigma\sqrt{\sum_{s=1}^{t}\norm{X_{s}}_{2}^{2}}\sqrt{2\log\frac{4t^{2}}{\delta}}.
\]
\end{proof}

\end{document}